\pdfoutput=1

\documentclass[11pt]{article}

\usepackage[]{acl}
\usepackage{times}

\usepackage{amsmath,amsfonts,bm}

\def\eqref#1{equation~\ref{#1}}

\def\1{\bm{1}}

\DeclareMathAlphabet{\mathsfit}{\encodingdefault}{\sfdefault}{m}{sl}
\SetMathAlphabet{\mathsfit}{bold}{\encodingdefault}{\sfdefault}{bx}{n}

\usepackage{url}
\usepackage{breakurl}
\usepackage{enumitem}
\usepackage{hyperref}
\usepackage{url}
\usepackage{comment}
\usepackage{graphicx}
\usepackage{wrapfig}
\usepackage{booktabs}

\usepackage{subcaption}
\usepackage{multirow}
\usepackage{array}
\newif\ifshowcomment
    \showcommenttrue

\ifshowcomment
    \newcommand{\ganqu}[1]{\textcolor{purple}{[{ganqu: #1}]}}
    \newcommand{\yang}[1]{\textcolor{blue}{[yang: #1]}}

\else
    \newcommand{\todo}[1]{}
    \newcommand{\ganqu}[1]{}
    \newcommand{\yang}[1]{}

    \newcommand{\focus}[1]{}
\fi

\title{A Close Look into the Calibration of Pre-trained Language Models}

\author{Yangyi Chen$^{*}$\\
UIUC \\
\And
Lifan Yuan\thanks{~~Equal contribution} \\
HUST \\
\And
Ganqu Cui, Zhiyuan Liu \\
Tsinghua University 
\And
Heng Ji \\
UIUC 
\AND
{\tt yangyic3@illinois.edu} 
\And
{\tt lievanyuan173@gmail.com}
}

\definecolor{mediumelectricblue}{rgb}{0.01, 0.31, 0.59}

\newcommand{\ours}{(\textcolor{mediumelectricblue}{\small ours})}

\begin{document}

\maketitle

\begin{abstract}
Pre-trained language models (PLMs) may fail in giving reliable estimates of their predictive uncertainty. 
We take a close look into this problem, aiming to answer two questions: (1) Do PLMs learn to become calibrated in the training process? (2) How effective are existing calibration methods? For the first question, we conduct fine-grained control experiments to study the dynamic change in PLMs' calibration performance in training. We consider six factors as control variables, including dataset difficulty, available training samples, training steps, the number of tunable parameters, model scale, and pretraining. We observe a consistent change in calibration performance across six factors. We find that PLMs don't learn to become calibrated in training, evidenced by the continual increase in confidence, no matter whether the predictions are correct or not. We highlight that our finding somewhat contradicts two established conclusions: (a) Larger PLMs are more calibrated; (b) Pretraining improves model calibration. Next, we study the effectiveness of existing calibration methods in mitigating the overconfidence issue. Besides unlearnable calibration methods (e.g., label smoothing), we adapt and extend two recently proposed learnable methods that directly collect data to train models to have reasonable confidence estimations. Experimental results show that learnable methods significantly reduce PLMs' confidence in wrong predictions. The code is available at \url{https://github.com/lifan-yuan/PLMCalibration}.




\end{abstract}

\section{Introduction}
Pre-trained language models (PLMs) are successful in many downstream tasks regarding performance~\citep{DBLP:conf/iclr/WangSMHLB19}.
In high-stake applications, it's equally essential for PLMs to possess a sense of calibration~\citep{DBLP:conf/aistats/VaicenaviciusWA19}. 
%
%
However, the confidence scores (a.k.a, predictive probability) of existing deep neural networks cannot serve as reliable estimates of their uncertainty~\citep{DBLP:conf/icml/GuoPSW17}, and a deep understanding of PLMs calibration is lacking.


\begin{figure}[]
    \vspace{-12pt}
    \centering
    \includegraphics[width=0.4\textwidth]{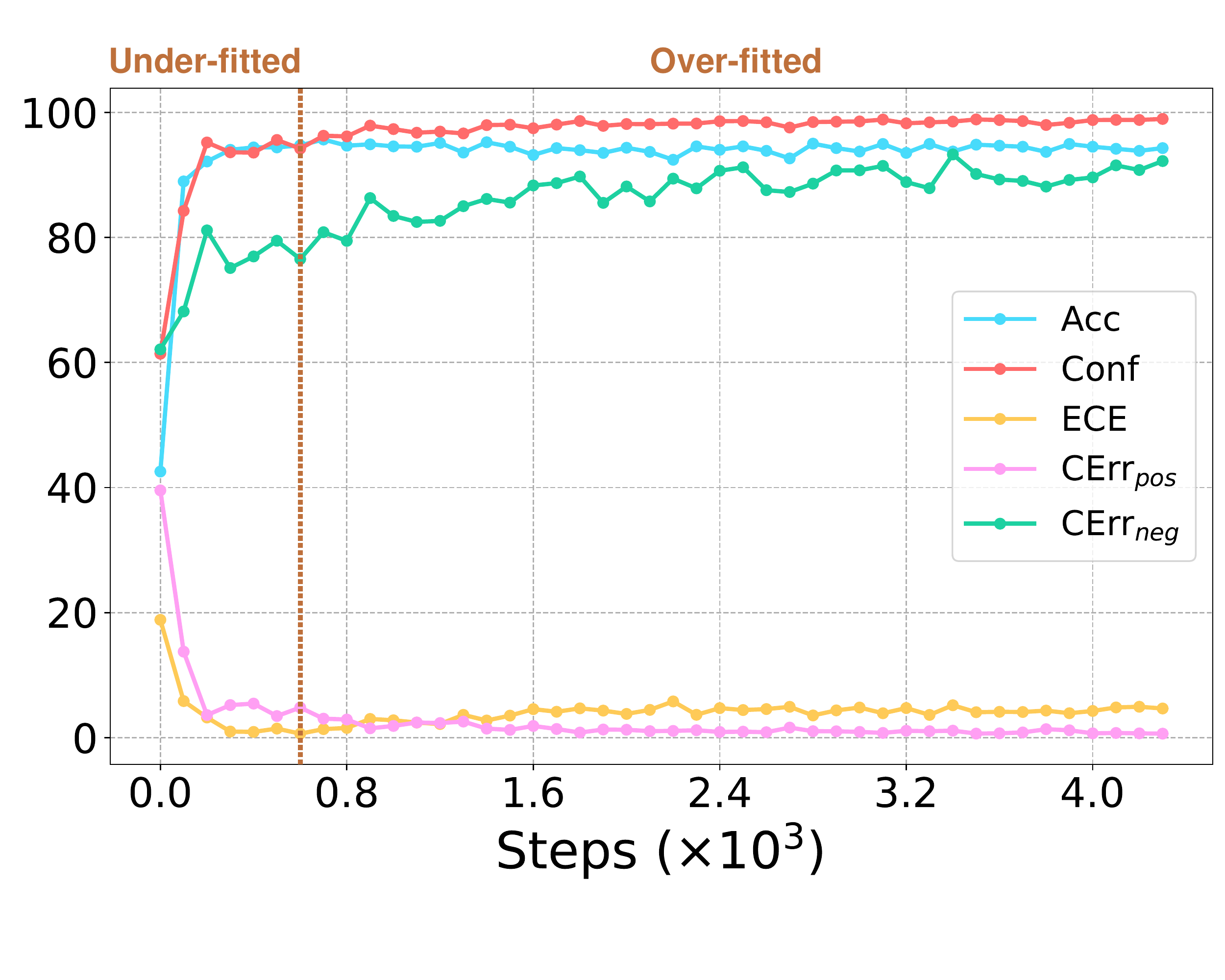}
    \vspace{-10pt}
    \caption{The demonstration of the under-fitted and over-fitted states in the training process with RoBERTa on SST-2.}
    \label{fig:hello}
    \vspace{-15pt}
\end{figure}

In this paper, we give a systematical analysis of PLMs calibration. 
We consider two questions about PLMs calibration:
(1) Do PLMs learn to become calibrated in the training process?
(2) How effective are existing calibration methods?
We first introduce the metrics we adopt for calibration performance evaluation.
The most widely used calibration metric ECE (Expected Calibration Error~\cite {DBLP:conf/aaai/NaeiniCH15}) is considered. 
It measures the difference between confidence and accuracy by portioning samples into various confidence zones.
To give a more comprehensive and practical calibration evaluation, we provide an application-driven perspective, describing two undesirable situations in practice:
(1) Correct predictions (positive) are rejected due to low confidence;
(2) Wrong predictions (negative) are accepted due to high confidence.
We propose to measure the average confidence scores on correct and wrong predictions respectively to characterize undesirable situations.
Two kinds of calibration errors are measured, denoted as CErr$_{pos}$ and CErr$_{neg}$.

%
For the first question, we consider the influence of six factors on PLMs' calibration performance, including dataset difficulty, available training samples, training steps, the number of tunable parameters, model scale, and pretraining.
Some of them are overlooked in previous empirical studies~\citep{DBLP:conf/nips/SnoekOFLNSDRN19, DBLP:conf/cvpr/NixonDZJT19, DBLP:conf/nips/MindererDRHZHTL21}.
We motivate to conduct fine-grained control experiments to study the dynamic change in PLMs' calibration performance in training through manipulating control variables.

\looseness=-1
We empirically observe an overall consistent change in calibration performance across six factors.
All six factors influence PLMs' fitness on the training distribution. 
This results in two states of PLMs considering calibration performance, namely under-fitted and over-fitted states (see Fig.\ref{fig:hello}). 
In the under-fitted state, PLMs' performance and confidence increase at different speeds when more fitted on the training distribution. 
In the over-fitted state, PLMs' confidence continues to increase steadily with little change in performance.
\textbf{We find evidence that PLMs don't learn to become calibrated in training:
PLMs' confidence in their predictions continues to increase when more fitted on the distribution (e.g., more tunable parameters, training longer).}
This results in two miscalibration behaviors:
(1) Increasing ECE in the latter over-fitted state, and (2) Continually increasing confidence in wrong predictions, indicating that PLMs mostly don't know ``what they don't know''.

We highlight our finding presents contradictory views with the two established conclusions:
(a) Larger PLMs show better calibration~\citep{srivastava2022beyond};
(b) Pretraining improves model calibration~\citep{hendrycks2019using}.
We identify that the inconsistency lies in:
(1) The difficulty of evaluation datasets: the performance doesn't saturate in the considered datasets (e.g., BIG-bench~\citep{srivastava2022beyond}).
Thus, the evaluation is on the under-fitted state, leaving the miscalibration behavior in the over-fitted state unobserved;
(2) Evaluation metrics: previous work doesn't measure the confidence in wrong predictions, overlooking the fact that models are becoming more confident in wrong predictions when scaling larger and employing pretraining.

Thus, we find that the main issue of PLMs calibration lies in their overconfidence in wrong predictions, which cannot be trivially solved by increasing the model scale. 
So we consider the effectiveness of existing calibration methods in mitigating the overconfidence issue. 
%
We partition existing calibration methods into unlearnable and learnable groups.
Unlearnable methods heuristically manipulate the original confidence in predictions (e.g., label smoothing).
Learnable methods directly collect data and train models to give reasonable confidence scores in their predictions.  
Namely, an extra calibration task is introduced, which aims to extract features from samples and models' preceding performance to predict whether models' predictions are correct or not.

\looseness=-1
\textbf{In our experiments, we identify the superiority of learnable methods compared to unlearnable ones, considering both in-distribution (ID) and out-of-distribution (OOD) settings.} 
This is characterized by a sharp decrease in their confidence in wrong predictions when using learnable methods, indicating that they significantly mitigate the overconfidence issue. 
Moreover, learnable methods can maintain a reasonable increase in CErr$_{pos}$, holding consistent correlations between the drop in confidence and performance under distribution shifts.
This shows the difference from unlearnable methods, which  take effect by roughly imposing confidence regularization on models' predictions (e.g., label smoothing), resulting in almost the same amount of increase in CErr$_{pos}$ with the decrease in CErr$_{neg}$.

%

%
%
%


\looseness=-1
To further understand learnable calibration methods, we consider the influence of more data and larger model scales for the calibration task, the adopted model for the calibration task, and the data distribution, on PLMs' calibration performance. 
We highlight three findings:
(1) More data and larger model scales for the calibration task both play significant positive roles in PLMs' calibration performance;
(2) PLMs can be trained to give their uncertainty. This finding is consistent with the concurrent work~\citep{lin2022teaching}.
Further, we provide an extension to this conclusion. 
We find that using an extrinsic predictive model can achieve comparable results, given the same calibration training data. 
Thus, we identify that the success of this paradigm essentially lies in the learnable attribute of the calibration task, instead of the PLMs' self-checking process;
(3) PLMs' calibration performance under distribution shifts depends on the evaluation datasets chosen. 
Previous work shows that PLMs exhibit degraded calibration performance under distribution shifts~\citep{DBLP:conf/emnlp/DesaiD20}. 
We find that this conclusion is reversed when the ID datasets are harder and PLMs achieve better performance on OOD datasets.
The concrete arguments and explanations are detailed in Appendix \ref{sec:further_dis}.

\section{Background}
\label{sec:background}
\paragraph{Calibration measure.}
\looseness=-1
We can visualize model calibration through reliability diagram~\citep{degroot1983comparison}.
Based on the diagram, we can measure the ECE~\citep{DBLP:conf/aaai/NaeiniCH15} by partitioning samples into different confidence zones.
The central idea is to measure the absolute difference between models' predictive confidence and accuracy.
Although alternative theoretic-motivated metrics have been proposed~\citep{DBLP:conf/aistats/VaicenaviciusWA19, DBLP:conf/iclr/GuptaRAMS021}, we still employ ECE in our experiments due to its simplicity and popularity.

\paragraph{Benchmark \& Analysis.}
Given appropriate evaluation metrics, large-scale benchmarks have been conducted to analyze model calibration under different settings, spanning model architectures~\citep{DBLP:conf/icml/GuoPSW17, DBLP:conf/nips/MindererDRHZHTL21}, model scales~\citep{DBLP:conf/emnlp/DanR21}, modalities~\citep{DBLP:conf/emnlp/DesaiD20,DBLP:conf/nips/MindererDRHZHTL21, kadavath2022language}, calibration methods~\citep{DBLP:conf/icml/GuoPSW17, DBLP:conf/emnlp/DesaiD20}, and distribution shifts~\citep{DBLP:conf/cvpr/NixonDZJT19, DBLP:conf/emnlp/KongJZLZZ20}. 
Our work is closely related to~\citet{xiao2022uncertainty} that quantifies the uncertainty of PLMs. 
However, previous benchmarks follow the fixed training and evaluation paradigms.
In this paper, we instead conduct a fine-grained and more comprehensive empirical evaluation to take a close look into PLMs calibration from multiple dimensions that have often been overlooked. 
Also, we consider and conduct a detailed analysis of the recently proposed learnable calibration methods~\citep{lin2022teaching, kadavath2022language}.
%


\paragraph{Method.}
Calibration is essential for out-of-distribution detection~\citep{hendrycks2019scaling}, selective prediction~\citep{varshney2022investigating}, robustness~\citep{kumar2022calibrated}, and pseudo-labeling~\citep{rizve2021defense}. 
Existing calibration methods can be partitioned into unlearnable and learnable groups. 
For unlearnable methods, there are mainly four categories.
Post-hoc calibration intends to readjust the output logits referring to the performance on a held-out validation set~\citep{platt1999probabilistic, DBLP:conf/icml/GuoPSW17}. 
Regularization methods aim to prevent models from being over-confident on predictions~\citep{DBLP:conf/cvpr/SzegedyVISW16, DBLP:conf/iclr/PereyraTCKH17}. 
Data augmentation~\citep{DBLP:conf/iclr/HendrycksMCZGL20, DBLP:conf/nips/WangXKYAW21} and model ensemble~\citep{DBLP:conf/icml/GalG16, DBLP:conf/nips/Lakshminarayanan17} have also been empirically proven to improve model calibration.
For learnable methods, the typical way is to first collect data for the calibration task, and then train a model to predict whether the given answer is correct. 
The model can be a multi-layer perceptron, and the features can be hand-engineered~\citep{DBLP:conf/acl/YeD22, DBLP:conf/acl/ZhangGC21, si2022revisiting} or the last hidden states of PLMs~\citep{kadavath2022language}. 
PLMs can also be directly trained to output their uncertainty by words~\citep{lin2022teaching}.


\section{Evaluation Metrics} 
\label{sec:exp_setting}

For basic evaluation, we report accuracy (Acc) and average confidence score (Conf) on the testing set.
For calibration evaluation, we report ECE using equal-mass binning and 100 bins following \citet{DBLP:conf/nips/MindererDRHZHTL21}.
Besides, we provide an application-driven perspective to evaluate model calibration, aiming to quantify two unsatisfied scenarios due to miscalibration in practice:
(1) Correct predictions (positive) are rejected due to low confidence; 
(2) Wrong predictions (negative) are accepted due to high confidence. 
Specifically, we consider the average confidence in correct predictions $\textbf{Conf}_{pos}$ and wrong predictions $\textbf{Conf}_{neg}$ respectively.
For unified comparison, we report two calibration error (CErr) cases, $\text{CErr}_{pos}=1-\text{Conf}_{pos}$ and $\text{CErr}_{neg}=\text{Conf}_{neg}$. 
In principle, we expect calibrated models to have both low CErr$_{pos}$ and CErr$_{neg}$, indicating that they reasonably assign high confidence in correction predictions and low confidence in wrong predictions.

\begin{figure*}
     \centering
     \begin{subfigure}[b]{0.32\textwidth}
         \centering
         \includegraphics[trim=0 10 0 0, clip, width=\textwidth]{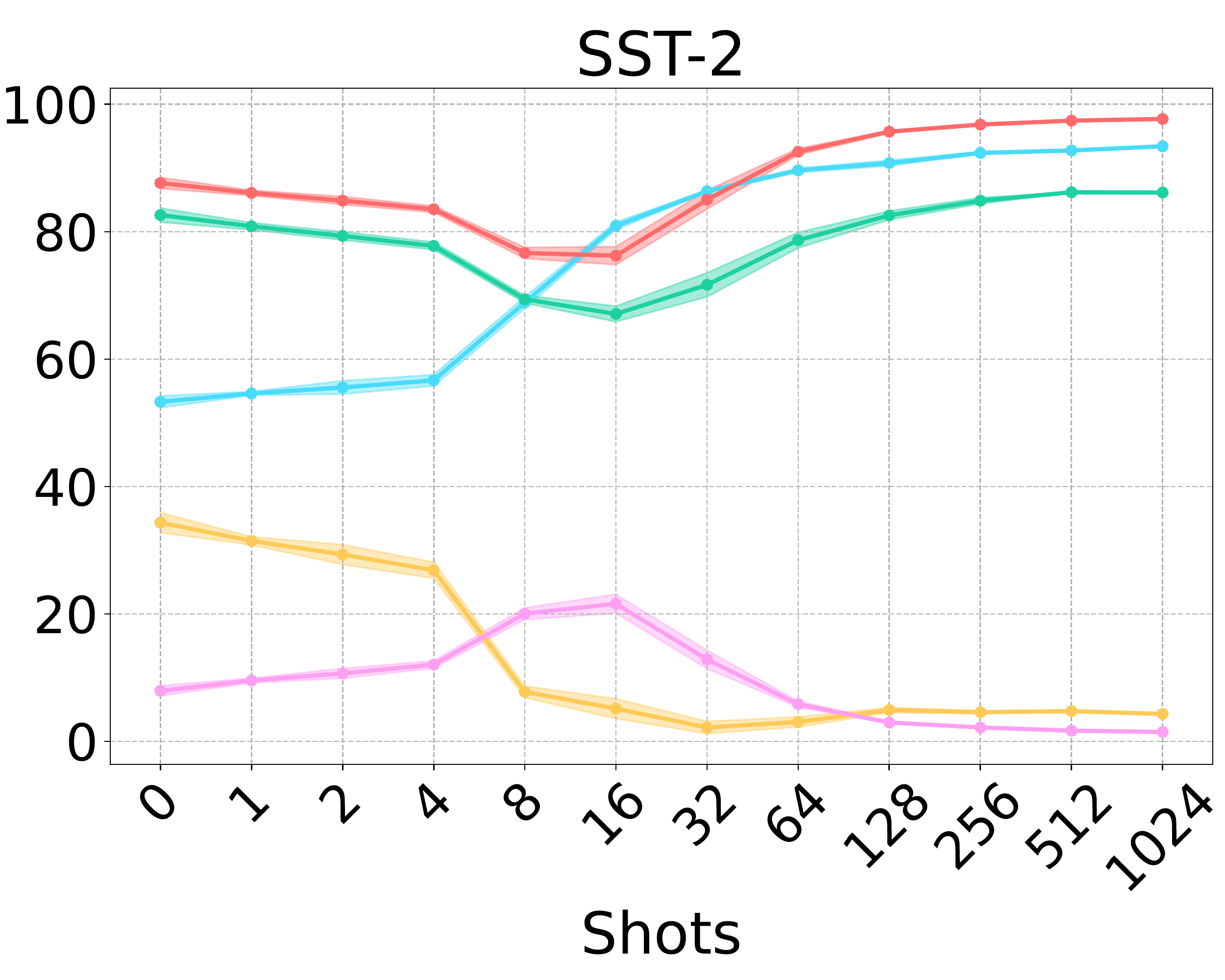}
         \label{fig:kshots-t5-sst2}
     \end{subfigure}  
    \centering
     \begin{subfigure}[b]{0.32\textwidth}
         \centering
         \includegraphics[trim=0 10 0 0, clip,width=\textwidth]{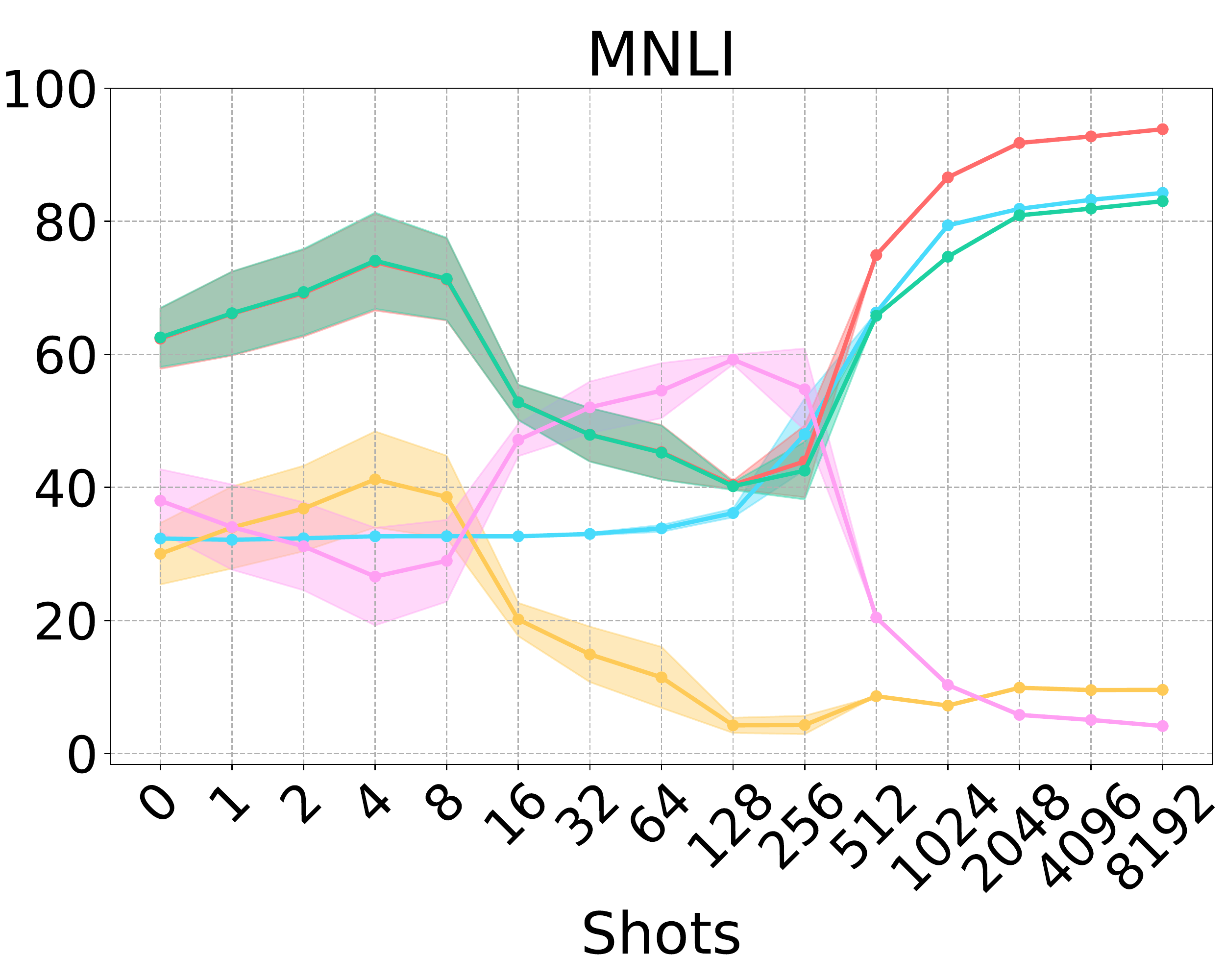}
         \label{fig:kshots-t5-mnli}
     \end{subfigure}
     \begin{subfigure}[b]{0.32\textwidth}
         \centering
         \includegraphics[trim=0 10 0 0, clip, width=\textwidth]{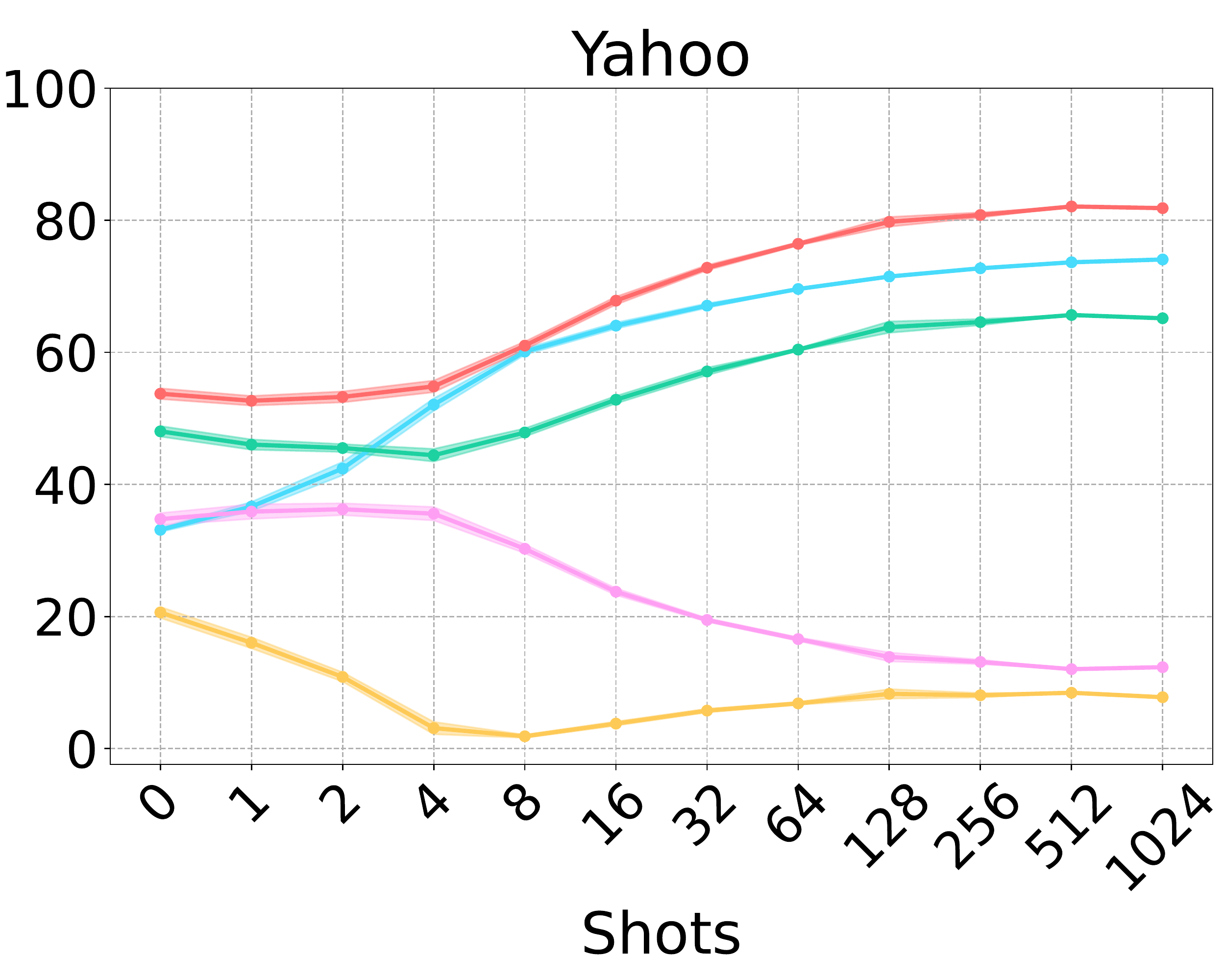}
         \label{fig:kshots-t5-yahoo}
     \end{subfigure}
     
     \vspace{-11pt}
     \begin{subfigure}[b]{0.6\textwidth}
         \centering
         \includegraphics[width=\textwidth]{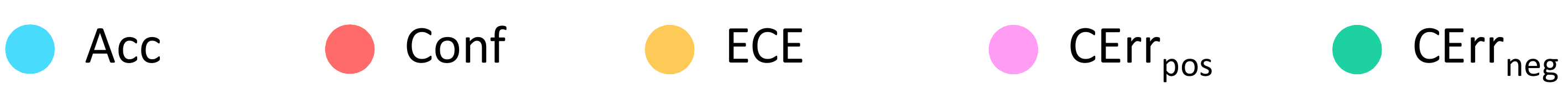}
     \end{subfigure}
         \vspace{-5pt}
        \caption{Results of available training samples with T5.}
        \label{fig:kshots-t5}
\end{figure*}

\begin{figure*}
     \centering
     \begin{subfigure}[b]{0.32\textwidth}
         \centering
         \includegraphics[trim=0 0 0 0, clip,width=\textwidth]{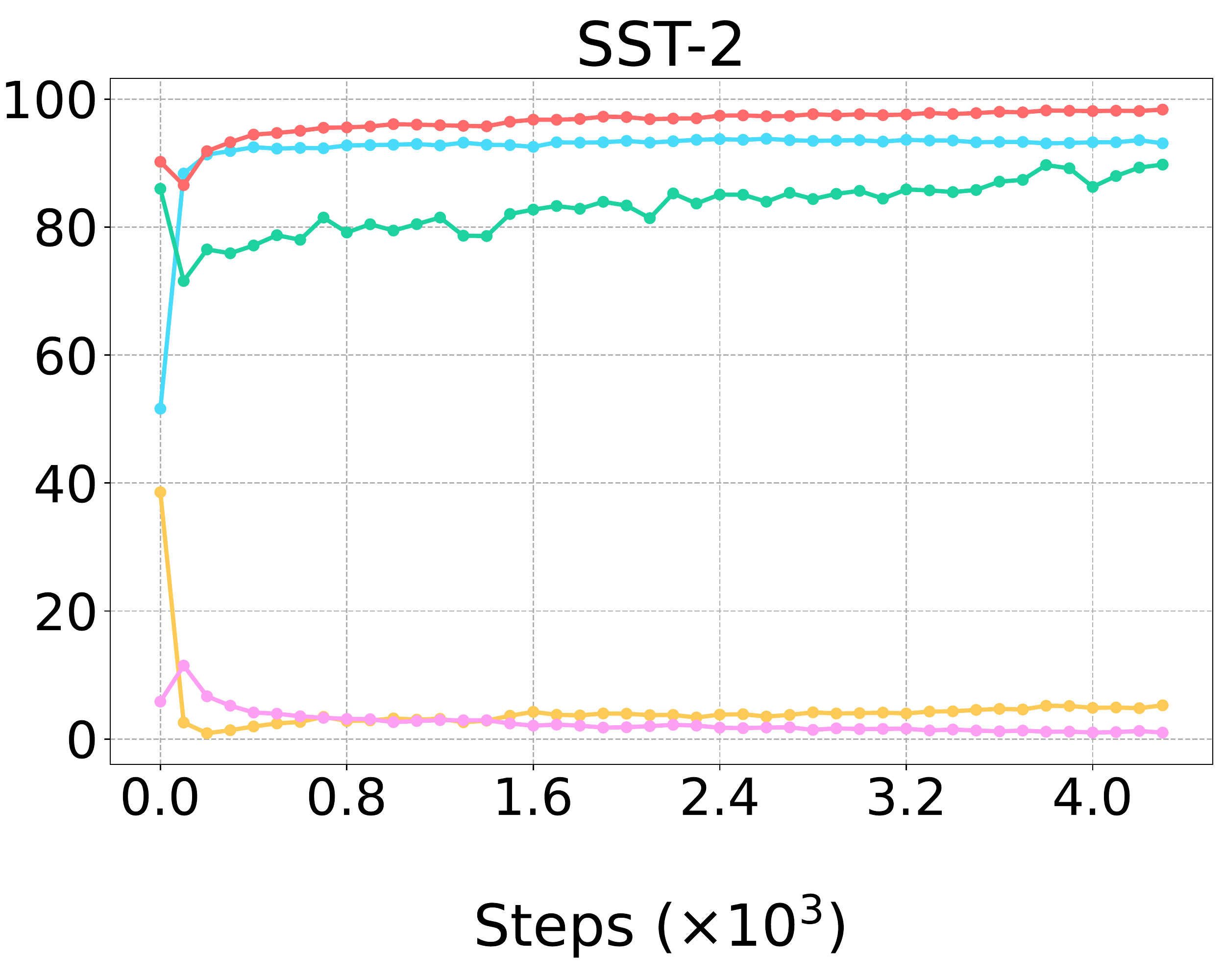}
         \label{fig:dynamics-t5-sst2}
     \end{subfigure}
     \begin{subfigure}[b]{0.32\textwidth}
         \centering
         \includegraphics[trim=0 0 0 0, clip,width=\textwidth]{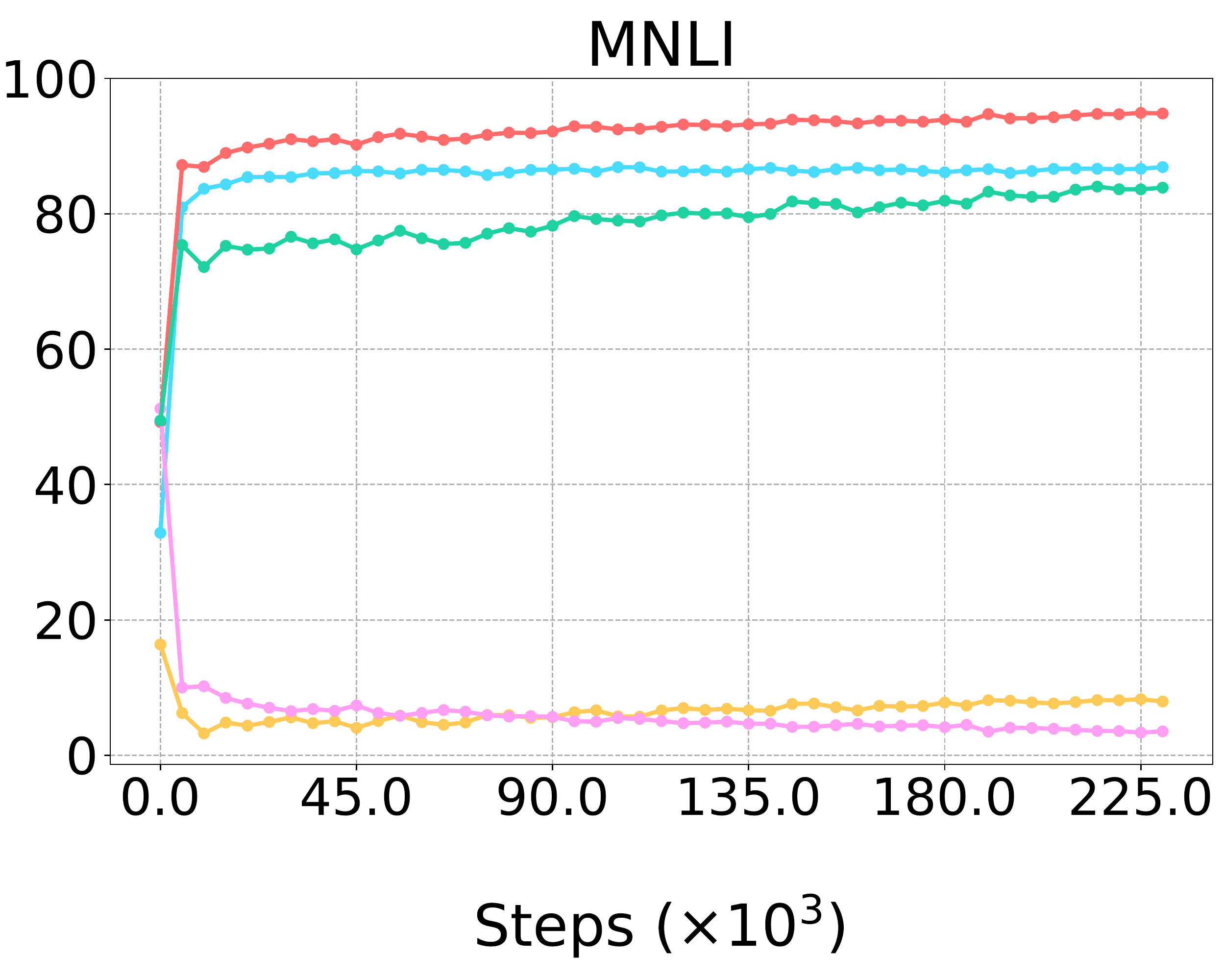}
         \label{fig:dynamics-t5-mnli}
     \end{subfigure}
     \begin{subfigure}[b]{0.32\textwidth}
         \centering
         \includegraphics[trim=0 0 0 10, clip,width=\textwidth]{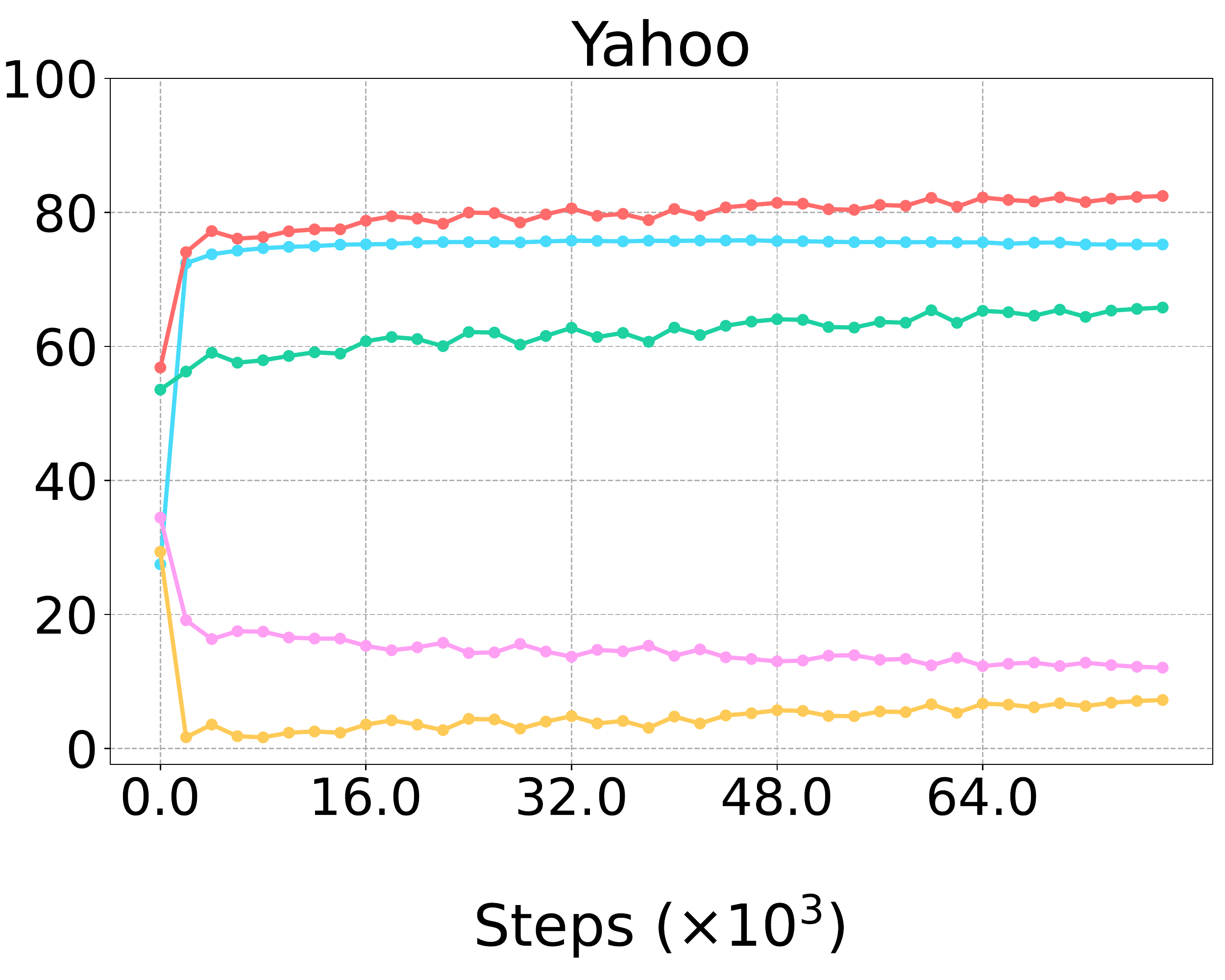}
         \label{fig:dynamics-t5-yahoo}
     \end{subfigure}
        
    \vspace{-13pt}
        
     \begin{subfigure}[b]{0.6\textwidth}
         \centering
         \includegraphics[width=\textwidth]{figures/legend.pdf}
     \end{subfigure}
        \vspace{-5pt}
        \caption{Results of training steps with T5.}
        \label{fig:dynamics-t5}
\end{figure*}

\section{\looseness=-1 Do PLMs Learn to Become Calibrated?}
\label{sec:empirical_study}

\subsection{Experimental Setting}
\looseness=-1
For model architectures, we choose RoBERTa-base~\citep{liu2019roberta} and T5-base~\citep{DBLP:journals/jmlr/RaffelSRLNMZLL20}, since they represent two classic types of PLMs, namely encoder-only and encoder-decoder models.
We experiment with four representative tasks in NLP, including sentiment analysis, natural language inference, news classification, and topic classification.
For datasets, we choose SST-2~\citep{DBLP:conf/emnlp/SocherPWCMNP13}, MNLI~\citep{DBLP:conf/naacl/WilliamsNB18}, AG-News~\citep{DBLP:conf/nips/ZhangZL15}, and Yahoo~\citep{DBLP:conf/nips/ZhangZL15} respectively. 
We employ the prompt-based learning paradigm~\citep{liu2021pre} since its superior performance compared to traditional fine-tuning, especially in the few-shot setting.
Specifically, we inherit the masked language modeling task in the pre-training stage and use templates to wrap samples into prompts.
We fine-tune the whole PLMs to fill in the [mask] position in the prompt. 
The manual template and verbalizer for each dataset are listed in Appendix~\ref{sec:appendix_dataset}.

\begin{figure*}
     \centering
     \begin{subfigure}[b]{0.32\textwidth}
         \centering
         \includegraphics[trim=0 0 0 0, clip, width=\textwidth]{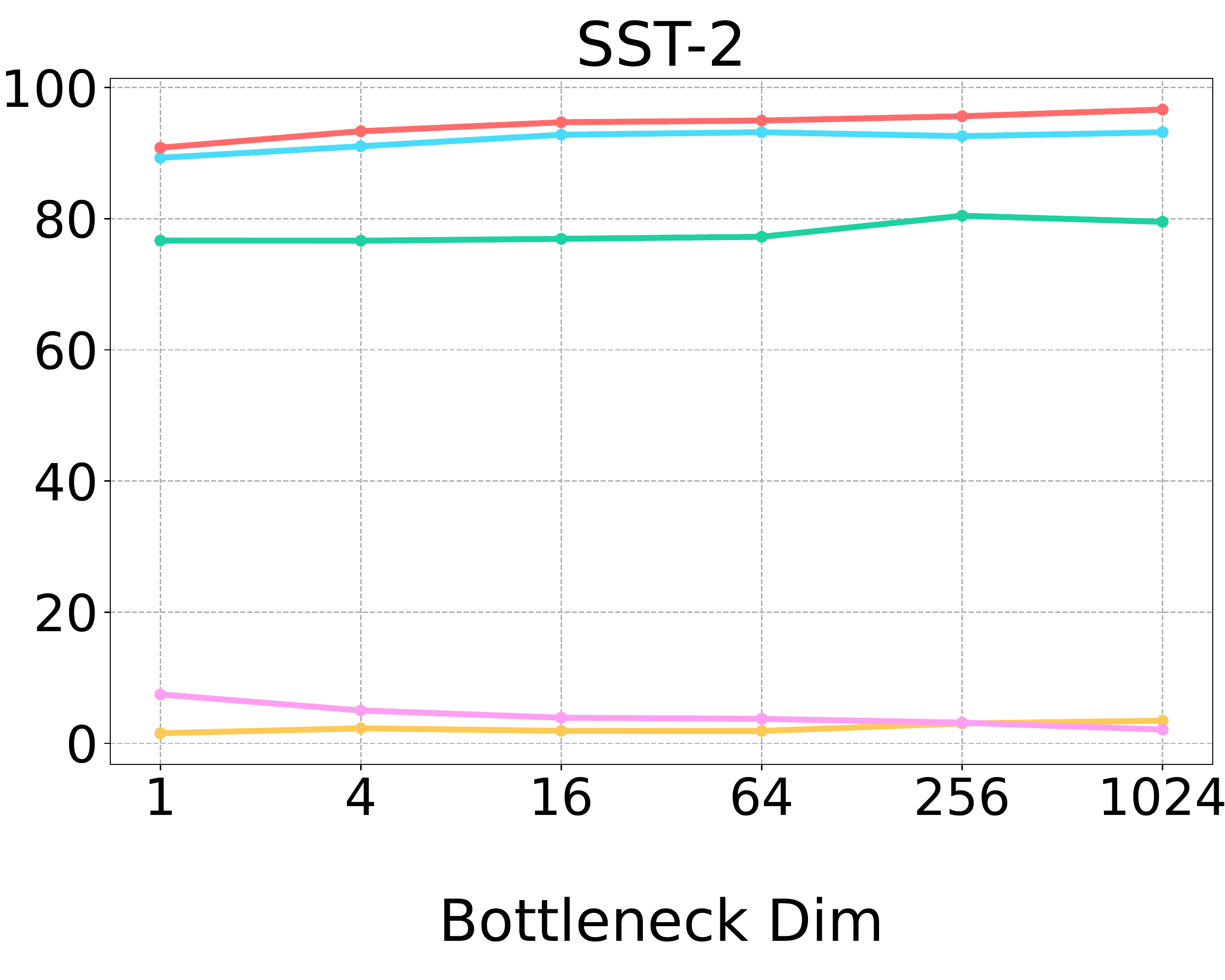}
         \label{fig:adapter-t5-sst2}
     \end{subfigure}
     \begin{subfigure}[b]{0.32\textwidth}
         \centering
         \includegraphics[trim=0 0 0 0, clip,width=\textwidth]{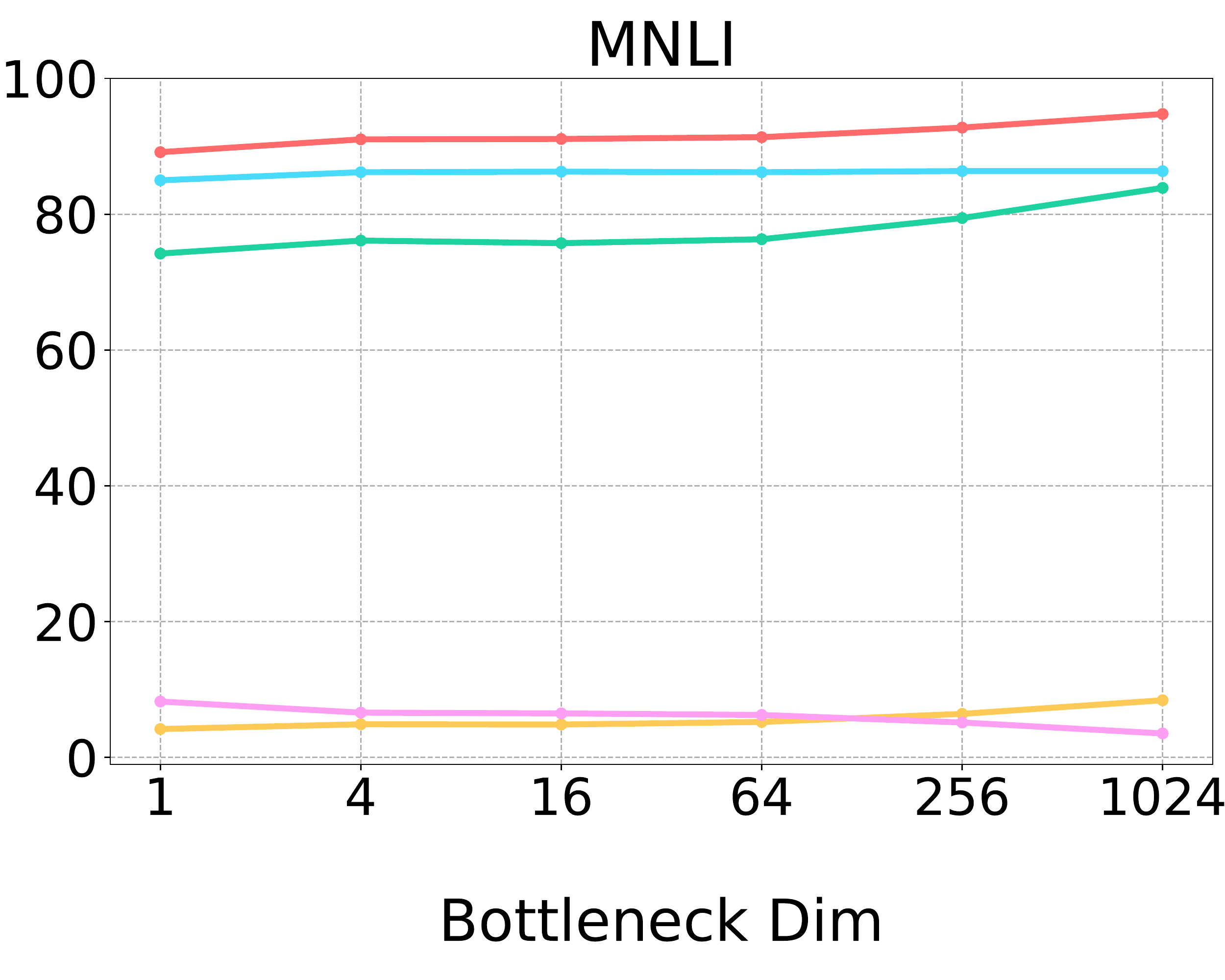}
         \label{fig:adapter-t5-mnli}
     \end{subfigure}
     \begin{subfigure}[b]{0.32\textwidth}
         \centering
         \includegraphics[trim=0 0 0 10, clip,width=\textwidth]{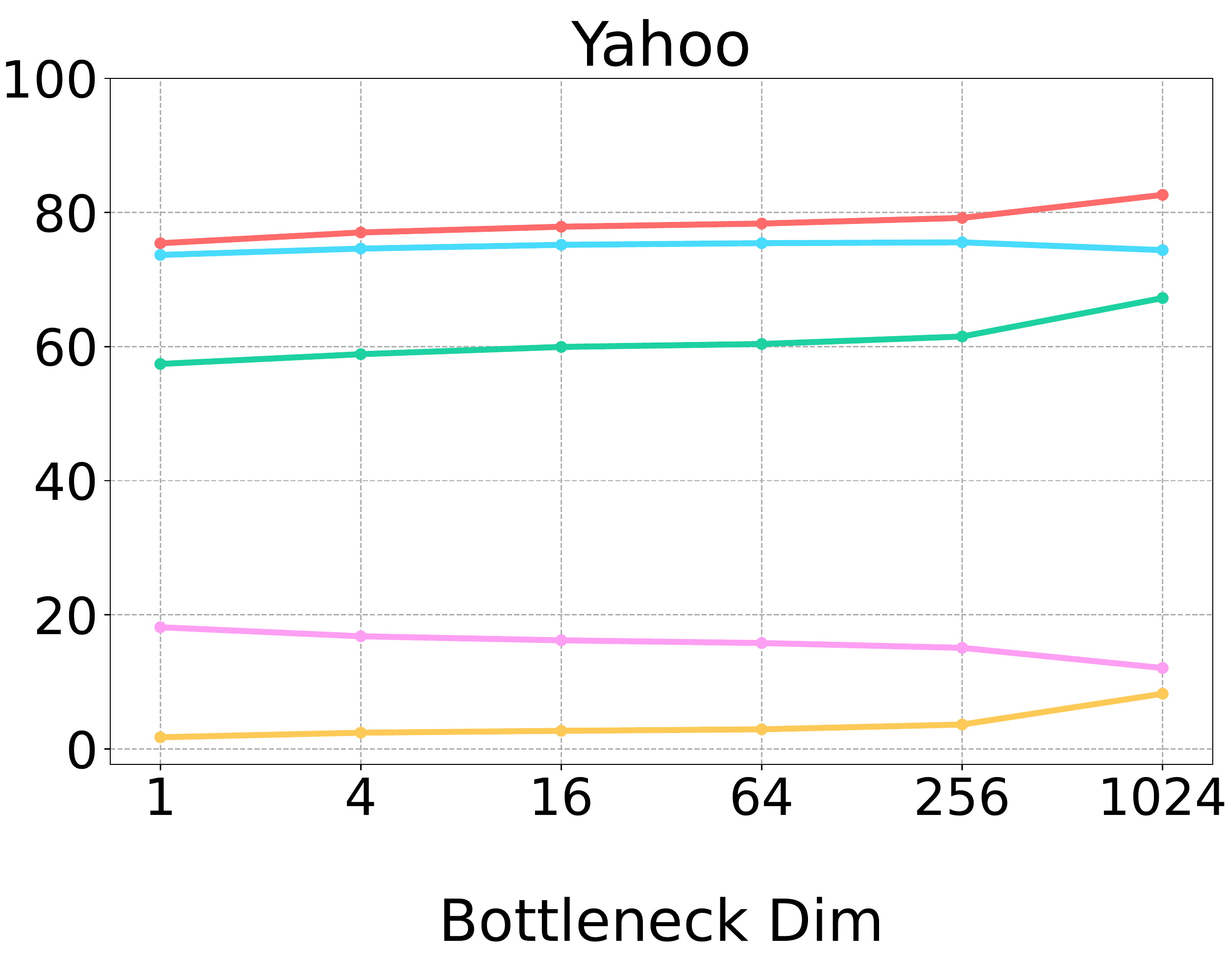}
         \label{fig:adapter-t5-yahoo}
     \end{subfigure}
     
    \vspace{-10pt}
     \begin{subfigure}[b]{0.6\textwidth}
         \centering
         \includegraphics[width=0.95\textwidth]{figures/legend.pdf}
     \end{subfigure}
        \vspace{-5pt}
        \caption{Results of tunable parameters with T5 (Adapter).}
        \label{fig:adapter-t5}
\end{figure*}

\begin{figure*}[!h]
     \centering
     \begin{subfigure}[b]{0.32\textwidth}
         \centering
         \includegraphics[trim=0 0 0 0, clip,width=\textwidth]{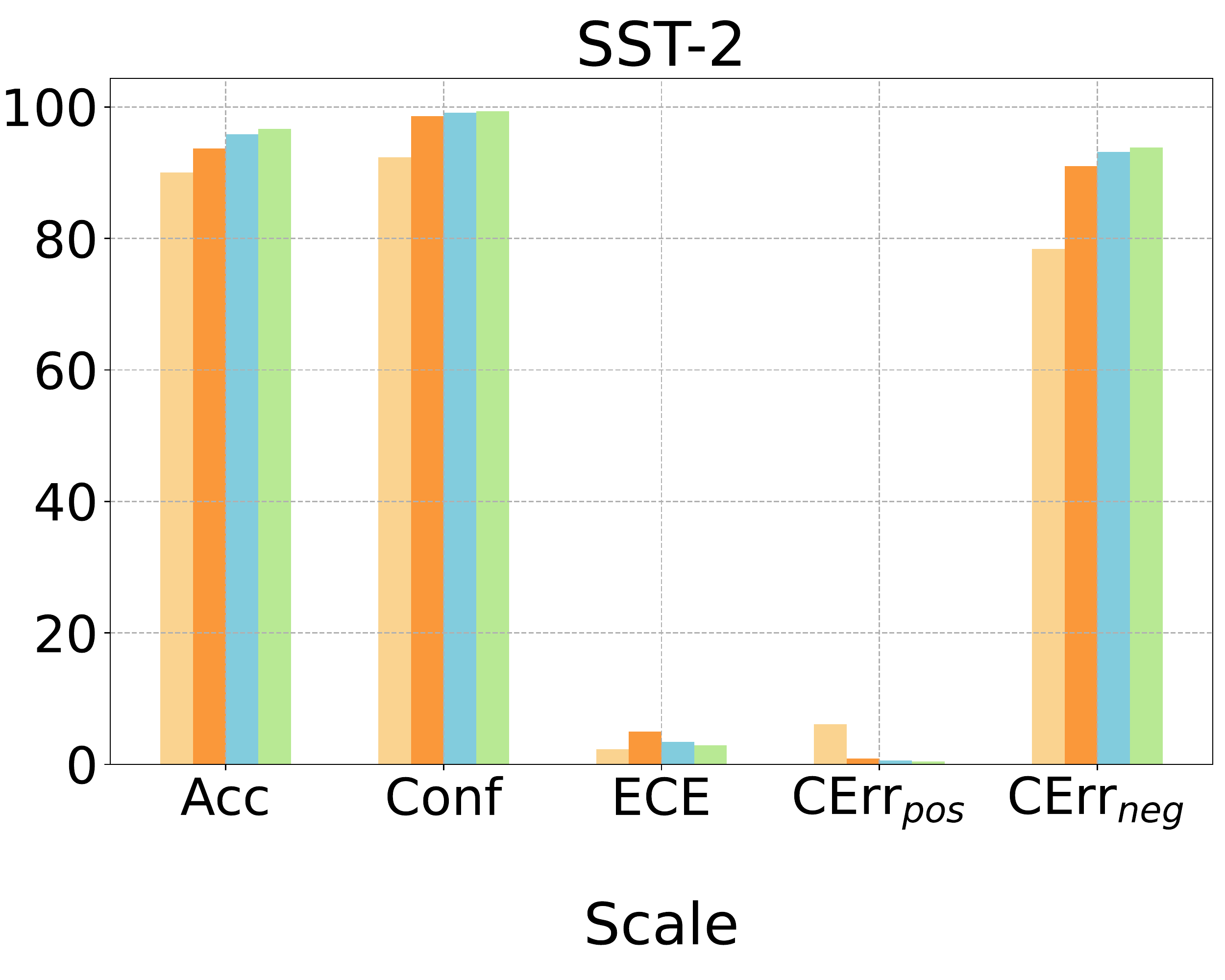}
         \label{fig:scale-t5-sst2}
     \end{subfigure}
     \begin{subfigure}[b]{0.32\textwidth}
         \centering
         \includegraphics[trim=0 0 0 0, clip,width=\textwidth]{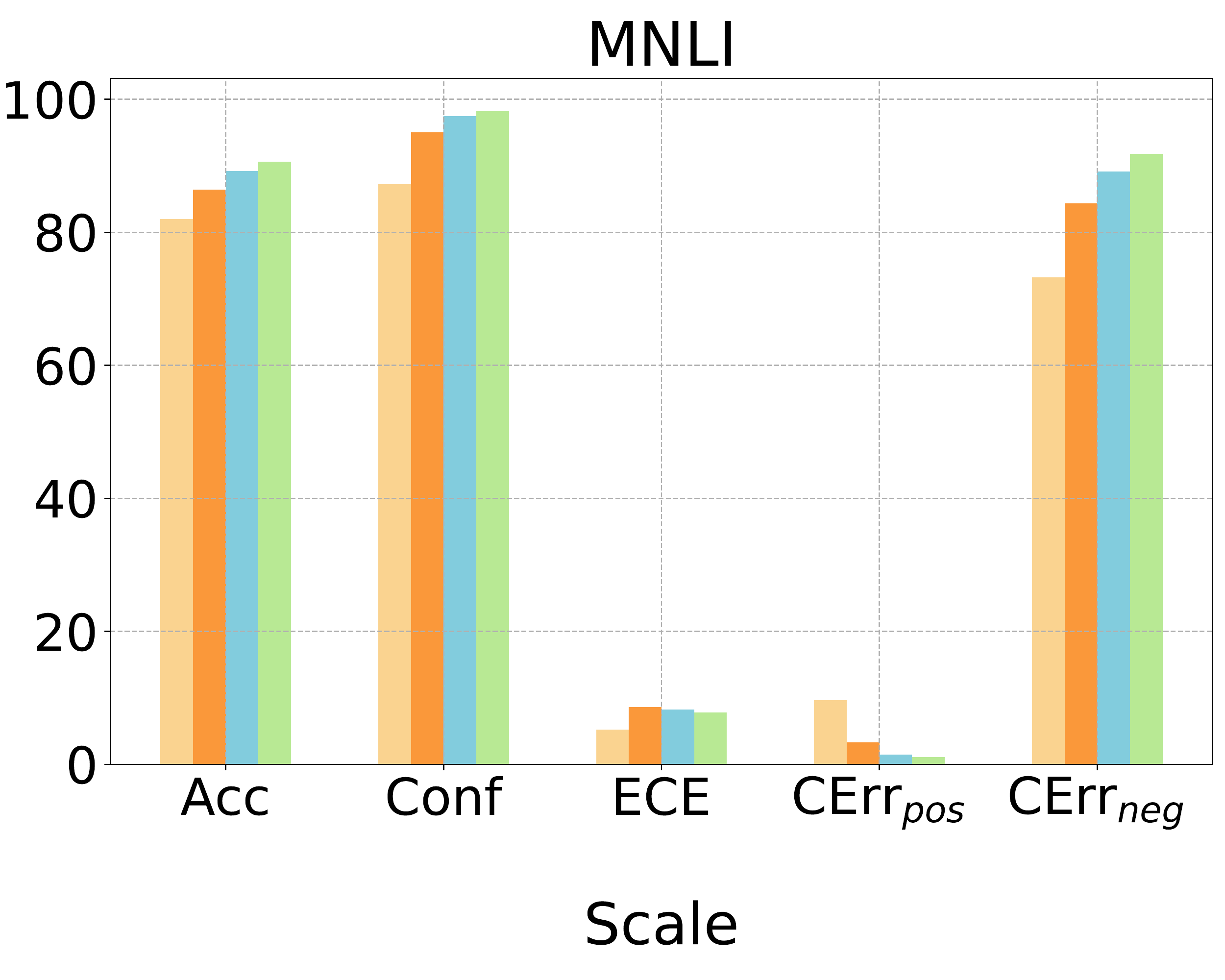}
         \label{fig:scale-t5-mnli}
     \end{subfigure}
     \begin{subfigure}[b]{0.32\textwidth}
         \centering
         \includegraphics[trim=0 0 0 0, clip,width=\textwidth]{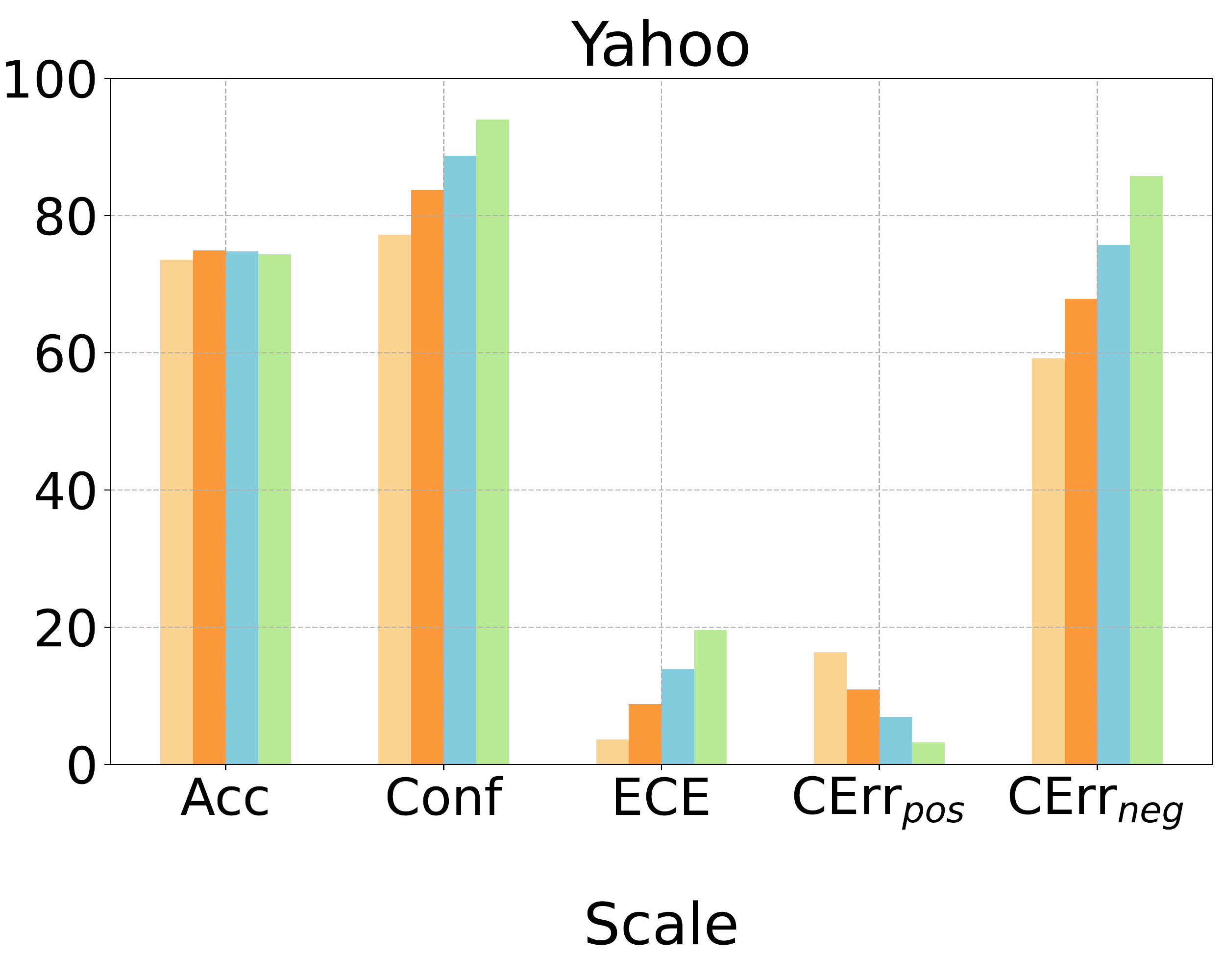}
         \label{fig:scale-t5-yahoo}
     \end{subfigure}
     
    \vspace{-12pt}
     \begin{subfigure}[b]{0.6\textwidth}
         \centering
         \includegraphics[trim=0 390 0 170, clip, width=\textwidth]{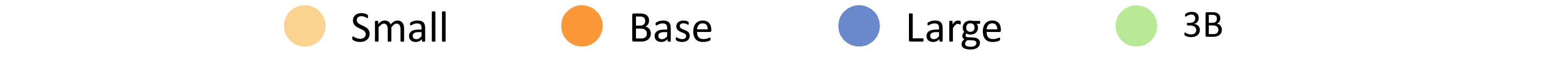}

     \end{subfigure}
        \vspace{-3pt}
        \caption{Results of increasing PLMs scales with T5.}
        \label{fig:scale-t5}
\end{figure*}
\begin{figure*}[htbp]
     \centering
     \begin{subfigure}[b]{0.32\textwidth}
         \centering
         \includegraphics[trim=10 20 0 0, clip, width=\textwidth]{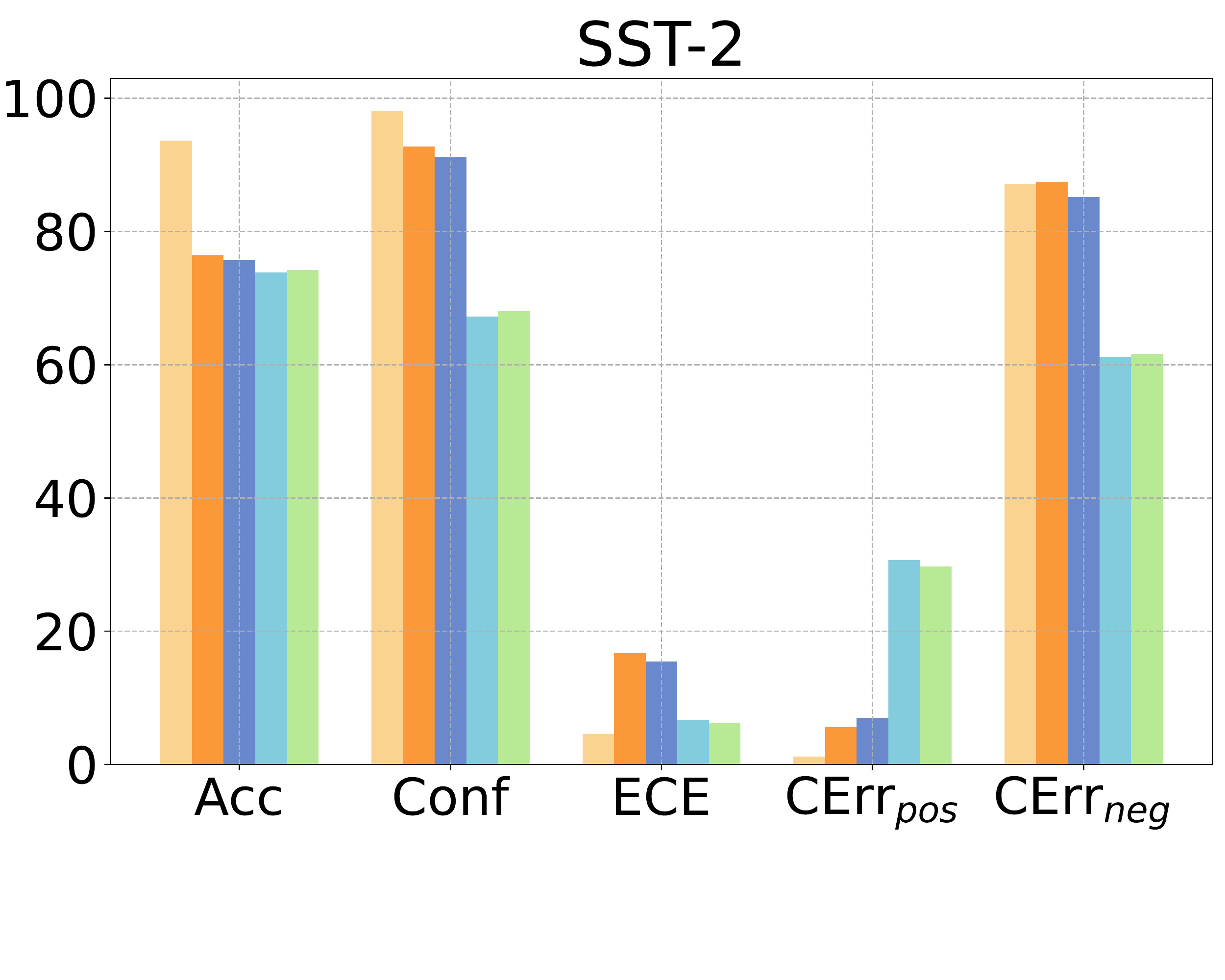}
         \label{fig:pretrain-t5-sst2}
     \end{subfigure}
     \begin{subfigure}[b]{0.32\textwidth}
         \centering
         \includegraphics[trim=10 20 0 0, clip, width=\textwidth]{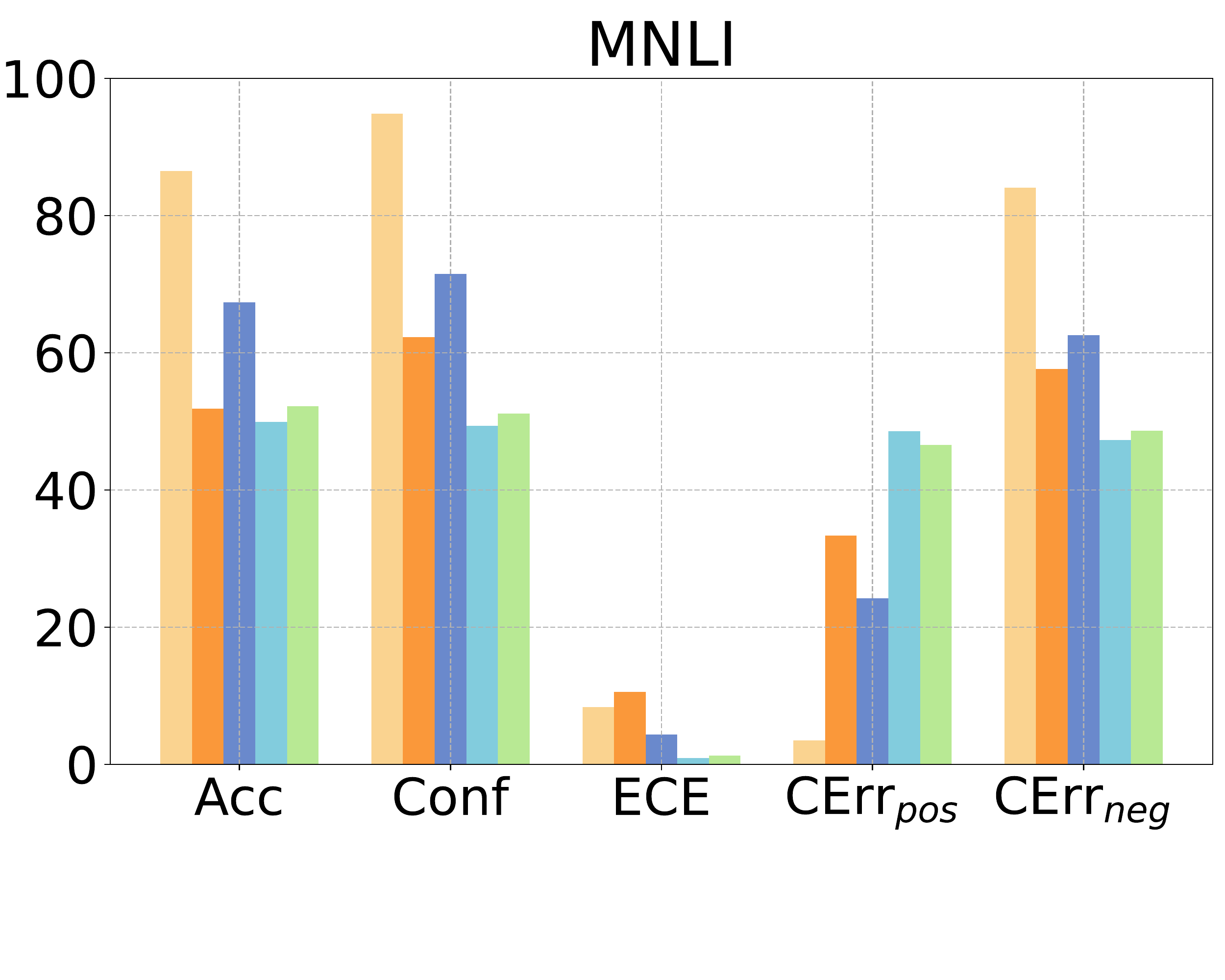}
         \label{fig:pretrain-t5-mnli}
     \end{subfigure}
     \begin{subfigure}[b]{0.32\textwidth}
         \centering
         \includegraphics[trim=10 20 0 0, clip, width=\textwidth]{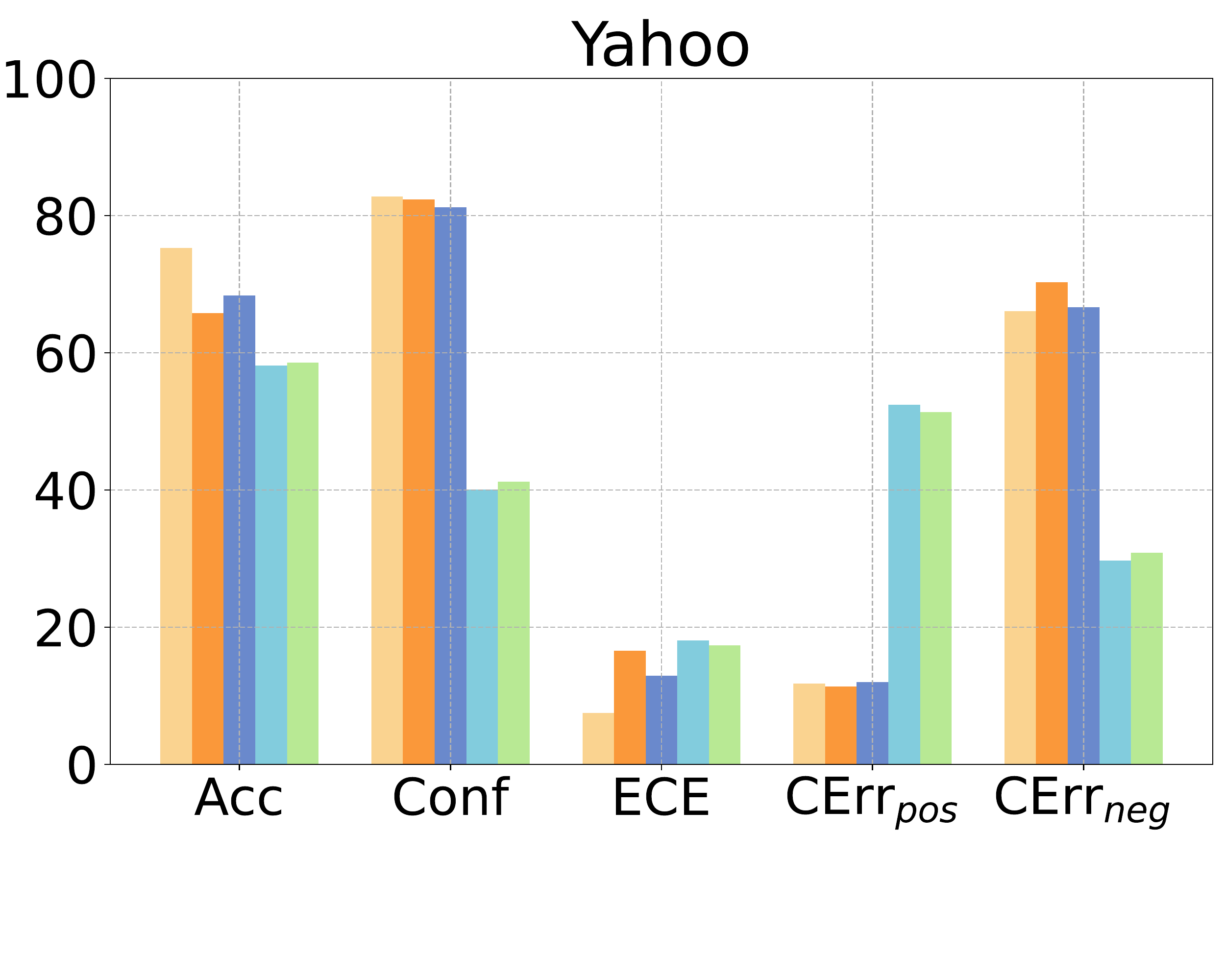}
         \label{fig:pretrain-t5-yahoo}
     \end{subfigure}
     
    \vspace{-17pt}
     \begin{subfigure}[b]{0.64\textwidth}
         \centering
         \includegraphics[width=\textwidth]{ 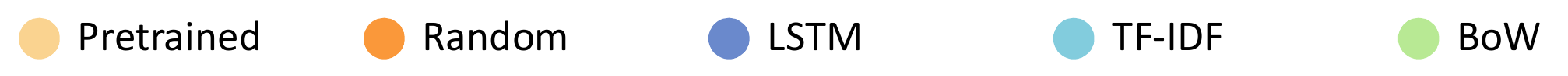}

     \end{subfigure}
        \caption{Results of the pretraining influence with T5.}
        \label{fig:pretrain}
\end{figure*}

\subsection{Experimental Results}
\looseness=-1
We conduct a fine-grained control study to explore the influence of six factors, 
including dataset difficulty, available training samples (Fig.\ref{fig:kshots-t5}), training steps (Fig.\ref{fig:dynamics-t5}), number of tunable parameters (Fig.\ref{fig:adapter-t5} and Fig.\ref{fig:soft_prompt-t5}), pretraining (Fig.\ref{fig:pretrain}), and model scale (Fig.\ref{fig:scale-t5}). 
Due to space limits, we show the corresponding results of RoBERTa and results of T5 on AG-News in Appendix~\ref{sec:appendix_additional_first}.
We summarize the overall conclusions and leave the detailed experimental settings and findings in Appendix~\ref{sec:appendix_additional_first}.

\looseness=-1
We note that all six factors dynamically influence PLMs' fitness on the training distribution, which we identify as the decisive factor of PLMs' calibration performance. 
We observe an overall consistent change in calibration performance across six factors, resulting in two PLMs' states (see Fig.\ref{fig:hello}) in training:

\looseness=-1
\textbf{Under-fitted state.}
In this state, PLMs' performance and confidence increase at different speeds when more fitted on the training distribution.
The ECE score fluctuates during this process.
In principle, miscalibration is due to the mismatch between performance and confidence.
However, we look closely into some critical points where ECE changes sharply (e.g., Fig.\ref{fig:kshots-t5}), and empirically find that the increase or decrease in ECE can be estimated by comparing the increasing rates of PLMs' performance and confidence.
We observe that a larger (smaller) increasing rate in performance reduces (increases) ECE.
Thus, high ECE can be partially attributed to PLMs' relatively rapid growth in confidence with performance lagging behind. 

\textbf{Over-fitted state.}
In this state, PLMs' performance doesn't have a substantial difference due to their generalization ability~\citep{zhang2021understanding}. 
However, PLMs' confidence continues to increase in this state, resulting in increasing ECE.
This is especially obvious when more training steps and tunable parameters are introduced (see Fig.\ref{fig:dynamics-t5} and Fig.\ref{fig:adapter-t5}). 
Thus, being more fitted on the training distribution may bring a negative effect on PLMs calibration. 
In addition, due to the increase of ECE in this state, the evaluation of calibration performance may be sensitive to the training paradigm.
This indicates that previous conclusions drawn from empirical studies should be carefully examined since the training paradigms may be different in model architectures and calibration methods.

\looseness=-1
Given the two states observed, we conclude that \textbf{PLMs don't learn to become calibrated in training, evidenced by the continually increasing confidence in predictions, no matter correct or not, in the fitting process.} 
Specifically, this results in two miscalibration behaviors:
(1) Increasing ECE in the over-fitted state;
(2) The consistent increase in CErr$_{neg}$ throughout the whole training process. This is an undesirable property in practice since users may accept wrong predictions due to their high confidence, and indicates that PLMs mostly don't know ``what they don't know''.

\looseness=-1  
We highlight two of the considered factors, namely pretraining and model scales (Fig.\ref{fig:scale-t5} and Fig.\ref{fig:pretrain}), which are examined in previous work.
Our findings present some contradictory views with the established conclusions:
(1) Larger PLMs show better calibration~\citep{srivastava2022beyond};
(2) Pretraining improves model calibration~\citep{hendrycks2019using}.
Actually, scaling larger and employing pretraining are both strategies to increase PLMs capacity, making them more fitted on the training distribution.
Our general conclusion can also be applied. 
We highlight two observations: 
(1) Essentially, the influence of scaling larger and pretraining on PLMs calibration is dynamically determined by the relative increase in performance and confidence, which is highly relevant to the chosen evaluation datasets.
For example, the original scaling experiments are conducted on BIG-bench~\citep{srivastava2022beyond}, in which the performance is far from saturation and increasing the model scale brings substantial improvement to PLMs performance.
This shows consistency with the identified under-fitted state.
However, when the performance score saturates on evaluation datasets given the certain scale of PLM, scaling larger will only bring up confidence. 
This results in increasing ECE due to the mismatch between two trends (e.g., T5 and RoBERTa on Yahoo);
(2) Scaling larger and employing pretraining consistently bring CErr$_{neg}$ higher.
This indicates that these two strategies don't enable PLMs to learn to become calibrated in the training process.





%



\begin{table*}[thb]
\centering
\resizebox{0.95\textwidth}{!}{
\begin{tabular}{@{}l|ll|ccccc|ccccc|ccccc@{}}
\toprule
\multirow{12}{*}{MNLI}   & \multicolumn{2}{l|}{Dataset}                                  & \multicolumn{5}{c|}{MNLI}                                               & \multicolumn{5}{c|}{HANS}                                                & \multicolumn{5}{c}{ANLI}                                                  \\ \cmidrule(l){2-18} 
                         & \multicolumn{2}{l|}{Method}                                   & Acc            & Conf  & ECE           & CErr$_{pos}$  & CErr$_{neg}$   & Acc            & Conf  & ECE            & CErr$_{pos}$  & CErr$_{neg}$   & Acc            & Conf  & ECE            & CErr$_{pos}$   & CErr$_{neg}$   \\ \cmidrule(l){2-18} 
                         & \multicolumn{1}{l|}{\multirow{5}{*}{Unlearnable}} & Vanilla   & 86.50          & 94.85 & 8.35          & 3.47          & 84.12          & 55.06          & 92.36 & 37.30          & 5.96          & 90.30          & 31.31          & 85.58 & 54.27          & 16.22          & 86.41          \\
                         & \multicolumn{1}{l|}{}                             & TS        & 86.50          & 89.22 & \textbf{2.75} & 8.44          & 74.22          & 55.06          & 83.99 & 28.93          & 14.36         & 81.97          & 31.31          & 75.48 & 44.17          & 26.87          & 76.56          \\
                         & \multicolumn{1}{l|}{}                             & LS        & 86.19          & 85.53 & 3.41          & 13.06         & 76.74          & 56.94          & 83.74 & \textbf{26.80} & 16.19         & 83.64          & 30.50          & 77.71 & 47.21          & 23.77          & 78.36          \\
                         & \multicolumn{1}{l|}{}                             & EDA       & 86.29          & 95.44 & 9.15          & \textbf{3.06} & 86.01          & 52.73          & 92.24 & 39.50          & \textbf{4.61} & 88.72          & 30.34          & 87.45 & 57.11          & \textbf{13.86} & 88.03          \\
                         & \multicolumn{1}{l|}{}                             & Ensemble  & \textbf{86.54} & 94.82 & 8.28          & 3.53          & 84.22          & 56.52          & 91.90 & 35.38          & 6.72          & 90.15          & \textbf{31.41} & 85.49 & 54.09          & 16.49          & 86.40          \\ \cmidrule(l){2-18} 
                         & \multicolumn{1}{l|}{\multirow{5}{*}{Learnable}}   & E-MLP     & 86.50          & 89.28 & 5.52          & 10.69         & 89.10          & 55.06          & 87.38 & 32.34          & 12.59         & 87.34          & 31.31          & 81.65 & 50.74          & 18.39          & 81.66          \\
                         & \multicolumn{1}{l|}{}                             & E-T5 \ours     & 86.50          & 79.43 & 12.24         & 15.35         & 45.84          & 55.06          & 78.74 & 35.30          & 19.11         & 75.97          & 31.31          & 41.67 & 38.68          & 65.84          & 45.11          \\
                         & \multicolumn{1}{l|}{}                             & I-Vanilla & 85.58          & 78.40 & 12.45         & 15.69         & 43.33          & 53.55          & 68.34 & 33.38          & 27.48         & \textbf{63.53} & \textbf{31.41} & 40.92 & 38.30          & 65.43          & 43.82          \\
                         & \multicolumn{1}{l|}{}                             & I-Iter \ours   & 86.30          & 70.86 & 15.49         & 24.07         & 38.95          & 57.12          & 74.92 & 28.39          & 22.16         & 71.02          & 30.69          & 37.02 & \textbf{28.37} & 68.84          & \textbf{39.62} \\
                         & \multicolumn{1}{l|}{}                             & I-Simul \ours   & 86.53          & 76.50 & 17.65         & 17.15         & \textbf{35.64} & \textbf{57.15} & 80.26 & 38.64          & 15.85         & 75.08          & 30.66          & 38.65 & 46.06          & 68.40          & 41.76          \\ \midrule
\multirow{12}{*}{Amazon} & \multicolumn{2}{l|}{Dataset}                                  & \multicolumn{5}{c|}{Amazon}                                             & \multicolumn{5}{c|}{SST-5}                                               & \multicolumn{5}{c}{SemEval}                                               \\ \cmidrule(l){2-18} 
                         & \multicolumn{2}{l|}{Method}                                   & Acc            & Conf  & ECE           & CErr$_{pos}$  & CErr$_{neg}$   & Acc            & Conf  & ECE            & CErr$_{pos}$  & CErr$_{neg}$   & Acc            & Conf  & ECE            & CErr$_{pos}$   & CErr$_{neg}$   \\ \cmidrule(l){2-18} 
                         & \multicolumn{1}{l|}{\multirow{5}{*}{Unlearnable}} & Vanilla   & 91.00          & 95.65 & 4.86          & 2.97          & 82.05          & \textbf{69.73} & 82.78 & 13.52          & 12.30         & 71.72          & 55.03          & 76.83 & 21.75          & 17.54          & 69.94          \\
                         & \multicolumn{1}{l|}{}                             & TS        & 91.00          & 90.50 & \textbf{1.39} & 7.74          & 73.20          & \textbf{69.73} & 71.98 & \textbf{4.94}  & 23.01         & 60.69          & 55.03          & 65.45 & \textbf{10.37} & 29.14          & 58.83          \\
                         & \multicolumn{1}{l|}{}                             & LS        & 91.25          & 85.75 & 6.78          & 13.14         & 74.09          & 70.67          & 73.50 & 5.55           & 22.53         & 63.95          & 53.57          & 69.79 & 16.23          & 25.65          & 64.53          \\
                         & \multicolumn{1}{l|}{}                             & EDA       & 92.00          & 96.29 & 4.29          & \textbf{2.51} & 82.46          & 67.67          & 87.58 & 20.20          & \textbf{7.97} & 78.27          & \textbf{57.27} & 83.11 & 25.96          & \textbf{11.87} & 76.40          \\
                         & \multicolumn{1}{l|}{}                             & Ensemble  & 91.57          & 95.78 & 4.21          & 2.88          & 81.14          & 69.35          & 83.00 & 13.66          & 12.13         & 72.00          & 56.34          & 77.81 & 21.47          & 16.52          & 70.49          \\ \cmidrule(l){2-18} 
                         & \multicolumn{1}{l|}{\multirow{5}{*}{Learnable}}   & E-MLP     & 91.00          & 91.34 & 5.13          & 8.66          & 91.31          & \textbf{69.73} & 84.06 & 14.73          & 16.04         & 84.28          & 55.03          & 75.87 & 20.83          & 24.17          & 75.91          \\
                         & \multicolumn{1}{l|}{}                             & E-T5 \ours     & 91.00          & 70.36 & 20.65         & 23.02         & 3.40           & \textbf{69.73} & 35.23 & 38.72          & 57.70         & 18.95          & 55.03          & 27.61 & 28.30          & 58.42          & 10.50          \\
                         & \multicolumn{1}{l|}{}                             & I-Vanilla & 89.14          & 70.03 & 19.11         & 21.79         & \textbf{2.91}  & 68.23          & 32.70 & 38.85          & 58.35         & \textbf{13.49} & 42.52          & 21.53 & 21.80          & 55.84          & \textbf{4.79}  \\
                         & \multicolumn{1}{l|}{}                             & I-Iter \ours    & \textbf{92.20} & 72.66 & 19.54         & 21.66         & 5.58           & 70.67          & 33.17 & 38.49          & 60.59         & 18.13          & 55.38          & 26.91 & 28.86          & 59.90          & 10.52          \\
                         & \multicolumn{1}{l|}{}                             & I-Simul \ours   & 91.87          & 71.72 & 20.15         & 22.38         & 5.09           & 69.54          & 31.45 & 38.26          & 61.73         & 15.88          & 55.28          & 26.35 & 29.37          & 60.57          & 10.17          \\ \midrule
\multirow{12}{*}{Civil}  & \multicolumn{2}{l|}{Dataset}                                  & \multicolumn{5}{c|}{Civil}                                              & \multicolumn{5}{c|}{Hate Speech}                                         & \multicolumn{5}{c}{Implicit Hate}                                         \\ \cmidrule(l){2-18} 
                         & \multicolumn{2}{l|}{Method}                                   & Acc            & Conf  & ECE           & CErr$_{pos}$  & CErr$_{neg}$   & Acc            & Conf  & ECE            & CErr$_{pos}$  & CErr$_{neg}$   & Acc            & Conf  & ECE            & CErr$_{pos}$   & CErr$_{neg}$   \\ \cmidrule(l){2-18} 
                         & \multicolumn{1}{l|}{\multirow{5}{*}{Unlearnable}} & Vanilla   & 86.08          & 94.23 & 7.74          & 3.88          & 82.12          & 75.52          & 92.54 & 17.23          & 5.88          & 87.72          & 60.64          & 89.68 & 28.83          & 8.62           & 87.04          \\
                         & \multicolumn{1}{l|}{}                             & TS        & 86.08          & 89.65 & \textbf{3.16} & 7.79          & 73.27          & 75.52          & 86.29 & 11.13 & 11.60         & 79.84          & 60.64          & 82.24 & 21.38          & 15.49          & 78.71          \\
                         & \multicolumn{1}{l|}{}                             & LS        & 86.30          & 84.93 & 5.29          & 13.62         & 75.78          & 74.48          & 83.51 & \textbf{9.03}  & 14.65         & 78.15          & 60.64          & 81.19 & \textbf{20.55} & 17.36          & 78.95          \\
                         & \multicolumn{1}{l|}{}                             & EDA       & 86.87          & 95.46 & 8.59          & \textbf{3.09} & 85.83          & 73.64          & 95.20 & 21.56          & \textbf{3.57} & 91.75          & \textbf{61.95} & 92.92 & 30.97          & \textbf{5.78}  & 90.80          \\
                         & \multicolumn{1}{l|}{}                             & Ensemble  & 86.04          & 94.51 & 8.46          & 3.65          & 83.10          & 75.36          & 93.57 & 18.80          & 5.04          & 89.35          & 60.83          & 90.98 & 30.14          & 7.50           & 88.62          \\ \cmidrule(l){2-18} 
                         & \multicolumn{1}{l|}{\multirow{5}{*}{Learnable}}   & E-MLP     & 86.08          & 90.61 & 4.52          & 9.40          & 90.62          & 75.52          & 88.93 & 13.41          & 11.13         & 89.10          & 60.64          & 87.41 & 26.78          & 12.59          & 87.42          \\
                         & \multicolumn{1}{l|}{}                             & E-T5 \ours     & 86.08          & 66.22 & 19.87         & 23.24         & 0.99           & 75.52          & 41.80 & 46.42          & 55.51         & 33.51          & 60.64          & 25.28 & 40.27          & 64.82          & 10.02          \\
                         & \multicolumn{1}{l|}{}                             & I-Vanilla & 75.31          & 63.39 & 11.92         & 15.95         & \textbf{0.35}  & \textbf{75.73} & 39.32 & 48.19          & 57.19         & \textbf{28.43} & 56.39          & 22.68 & 38.30          & 65.48          & \textbf{7.38}  \\
                         & \multicolumn{1}{l|}{}                             & I-Iter \ours    & 86.58          & 69.04 & 17.53         & 20.50         & 1.61           & 74.06          & 45.69 & 44.92          & 52.14         & 39.52          & 61.29          & 29.05 & 38.67          & 60.89          & 13.11          \\
                         & \multicolumn{1}{l|}{}                             & I-Simul \ours   & \textbf{87.06} & 70.69 & 16.55         & 19.04         & 1.62           & 73.01          & 46.63 & 46.34          & 50.30         & 38.31          & 61.14          & 30.50 & 40.17          & 58.65          & 13.44          \\ \bottomrule
\end{tabular}
}
\caption{Results of T5's calibration performance under standard distribution shifts. We observe that learnable methods can significantly mitigate the overconfidence issue.}
\vspace{-10pt}
\label{tab:t5-shift}
\end{table*}

\section{How Effective are Existing Methods?}
\looseness=-1


\subsection{Calibration Methods}
\label{sec:cal_method}
We choose representative calibration methods from each category summarized in Sec.~\ref{sec:background}.
For unlearnable methods, we consider vanilla fine-tuning (Vanilla), temperature scaling (TS)~\citep{DBLP:conf/icml/GuoPSW17}, label smoothing (LS)~\citep{DBLP:conf/cvpr/SzegedyVISW16}, easy data augmentation (EDA)~\citep{DBLP:conf/emnlp/WeiZ19}, and deep-ensemble (Ensemble)~\citep{DBLP:conf/nips/Lakshminarayanan17}.
For learnable methods, an extra calibration task is introduced, aiming to train a model to predict whether the original predictions are correct or not.
Each sample in the dataset of the calibration task consists of the original input, the model's original prediction, and the label indicating whether the original prediction is correct or not.
We adopt the validation set to generate the training set for the calibration task.
We describe the specially designed training paradigms of different methods in the following paragraph and leave the detailed construction process of the calibration training dataset in Appendix~\ref{sec:appendix_construction_of_cal}.

\looseness=-1
For better clarification, we use the main task to denote the original task.
The predictive model for the calibration task can be a separate extrinsic model that we use ``E-'' for denotation.
Specifically, we adapt the method proposed in \citet{kadavath2022language} that uses MLP as the extrinsic model (E-MLP) and the inputs are the hidden states of the main task model.
Based on a similar intuition, we extend this method by using an extra T5 as the extrinsic model (E-T5). 
An example of the template to wrap the sample into an input prompt is: ``\textless original input\textgreater, the model's prediction is \textless prediction\textgreater, is the prediction True or False? It's \textless mask\textgreater.''
The probability of the ``True'' class in the calibration task is deemed as PLMs' confidence in their predictions. 
The concrete manual template and verbalizer of the calibration task for each dataset are listed in Table~\ref{tab:calibration_template}.

Besides, the main task model can also be directly employed to perform the calibration task. We deem this paradigm as the intrinsic one, denoted as ``I-''.
\citet{lin2022teaching} show that GPT-3~\citep{brown2020language} can be trained to output the uncertainty by words.
We adapt this method by first training the model using the main task data, and then continuing the training by using the calibration task data (I-Vanilla).
However, this continual learning paradigm may result in degraded performance in the main task according to our results. 
To tackle this, we propose two more practical intrinsic calibration methods through modifying the training paradigm. 
Specifically, we train PLMs iteratively (I-Iter) or simultaneously (I-Simul) on the original task and the calibration task. 
The latter can be achieved due to the unified text-to-text training paradigm. 
The input is the same as E-T5.

\subsection{Experimental Setting}
PLMs are expected to tackle out-of-distribution (OOD) samples in practice, particularly in the presence of adversarial attacks~\citep{DBLP:conf/emnlp/ChenGCQH0S22}.
Thus, we experiment with both in-distribution (ID) and OOD settings. 
We consider natural language inference, sentiment analysis, and hate-speech detection tasks due to their well-established OOD datasets in NLP. 
Specifically, we choose MNLI (HANS, ANLI), Amazon (SST-5, SemEval), and Civil (Hate Speech, Implicit Hate) as the ID (OOD) datasets.
The references and detailed descriptions of chosen datasets for ID and OOD evaluation are in Appendix~\ref{sec:appendix_dataset}.


\subsection{Experimental Results}
\label{sec:standard}
The results are listed in Table~\ref{tab:t5-shift} (T5) and Table~\ref{tab:roberta-shift} (RoBERTa). 
We summarize the overall conclusions as follows:
All calibration methods have negligible influence on PLMs' performance in the ID and OOD settings except I-Vanilla.
However, PLMs are significantly less calibrated under considered distribution shifts, especially on challenging datasets due to the severe mismatch between performance and confidence. 
For example, the vanilla T5 achieves only 30.53\% accuracy on ANLI, but its average confidence is up to 93.77\%. 
For ID evaluation, we observe lower ECE, consistent with \citet{DBLP:conf/emnlp/DesaiD20}.
However, the conclusion that PLMs are calibrated on ID data~\citep{DBLP:conf/emnlp/DesaiD20} is questionable given our answer to the first question (see Sec.~\ref{sec:empirical_study}).
The low ECE can be attributed to their high performance on ID datasets and consistently assigning high confidence scores to their predictions. 
We further show the conclusion that PLMs calibration degrades under distribution shifts is one-sided and heavily depends on the evaluation datasets chosen in Appendix \ref{sec:further_dis}.



\textbf{Unlearnable methods.} 
We summarize the findings as follows:
(1) Data augmentation and model ensemble don't bring substantial benefits to PLMs calibration, considering the three calibration metrics spanning all evaluation datasets and two PLMs.
The reason lies in their inability to relieve the overconfident issue, resulting in the same Cerr$_{neg}$ with the vanilla fine-tuning;
(2) TS achieves overall better ECE, maintaining a strong baseline method, with LS being the second effective method for the unlearnable category.
This is consistent with previous empirical studies~\citep{DBLP:conf/cvpr/NixonDZJT19}.
However, we can observe almost the same amount of increase in CErr$_{pos}$ with the decrease in CErr$_{neg}$. 
The reason is that these two methods directly impose confidence regularization on predictions, which don't actually make PLMs have clear confidence estimations.


\begin{table*}[t]

\resizebox{\textwidth}{!}{

\begin{tabular}{@{}l|l|ccccc|ccccc|ccccc@{}}
\toprule
Dataset Size            & Dataset   & \multicolumn{5}{c|}{Amazon}                                             & \multicolumn{5}{c|}{SST-5}                                                & \multicolumn{5}{c}{SemEval}                                                        \\ \midrule
\multirow{6}{*}{Small}  & Method    & Acc            & Conf  & ECE           & CErr$_{pos}$  & CErr$_{neg}$   & Acc            & Conf  & ECE            & CErr$_{pos}$   & CErr$_{neg}$   & Acc            & Conf           & ECE            & CErr$_{pos}$   & CErr$_{neg}$   \\ \cmidrule(l){2-17} 
                        & E-MLP     & 91.00          & 90.41 & \textbf{1.71} & \textbf{9.59} & 90.39          & \textbf{69.73} & 87.81 & 18.08          & \textbf{12.16} & 87.73          & \textbf{55.03} & 86.86          & 31.83          & \textbf{13.11} & 86.83          \\
                        & E-T5 \ours     & 91.00          & 68.92 & 22.08         & 28.16         & 39.44          & \textbf{69.73} & 55.95 & 15.12          & 41.71          & 50.58          & \textbf{55.03} & 50.99          & \textbf{8.54}  & 43.17          & 43.84          \\
                        & I-Vanilla & 89.06          & 68.45 & 20.61         & 28.01         & 39.62          & 63.92          & 56.49 & \textbf{10.66} & 39.82          & 49.96          & 51.48          & 49.47          & 9.12           & 44.10          & 42.64          \\
                        & I-Iter \ours    & 90.58          & 68.96 & 21.62         & 28.08         & 40.47          & 69.63          & 56.69 & 12.95          & 41.27          & 52.00          & 53.72          & 53.89          & 10.24          & 43.31          & 50.64          \\
                        & I-Simul \ours   & \textbf{91.37} & 80.44 & 15.44         & 15.05         & \textbf{32.78} & 71.13          & 66.28 & 26.97          & 25.58          & \textbf{46.23} & 54.08          & \textbf{37.51} & 34.94          & 53.82          & \textbf{27.30} \\ \midrule
\multirow{6}{*}{Middle} & Method    & Acc            & Conf  & ECE           & CErr$_{pos}$  & CErr$_{neg}$   & Acc            & Conf  & ECE            & CErr$_{pos}$   & CErr$_{neg}$   & Acc            & Conf           & ECE            & CErr$_{pos}$   & CErr$_{neg}$   \\ \cmidrule(l){2-17} 
                        & E-MLP     & 91.00          & 90.44 & \textbf{4.35} & \textbf{9.56} & 90.41          & \textbf{69.73} & 85.18 & \textbf{15.45} & \textbf{14.69} & 84.87          & 55.03          & 78.39          & 23.36          & \textbf{21.63} & 78.42          \\
                        & E-T5 \ours     & 91.00          & 71.03 & 19.97         & 22.40         & 4.63           & \textbf{69.73} & 31.73 & 38.80          & 61.80          & 16.83          & 55.03          & 29.72          & 26.28          & 56.23          & 12.54          \\
                        & I-Vanilla & 88.25          & 70.91 & 17.34         & 20.16         & \textbf{3.86}  & 63.07          & 29.81 & 34.08          & 59.42          & \textbf{11.42} & 48.08          & 25.32          & 23.69          & 55.53          & \textbf{7.59}  \\
                        & I-Iter \ours    & \textbf{91.69} & 71.76 & 19.93         & 22.23         & 5.43           & 68.23          & 33.46 & 36.87          & 59.79          & 18.96          & \textbf{56.23} & 35.21          & \textbf{21.42} & 50.98          & 17.48          \\
                        & I-Simul \ours   & 91.38          & 70.92 & 20.47         & 22.80         & 4.30           & 70.29          & 32.03 & 42.12          & 60.65          & 14.72          & 54.75          & 26.18          & 30.70          & 59.34          & 8.67           \\ \midrule
\multirow{6}{*}{Large}  & Method    & Acc            & Conf  & ECE           & CErr$_{pos}$  & CErr$_{neg}$   & Acc            & Conf  & ECE            & CErr$_{pos}$   & CErr$_{neg}$   & Acc            & Conf           & ECE            & CErr$_{pos}$   & CErr$_{neg}$   \\ \cmidrule(l){2-17} 
                        & E-MLP     & 91.00          & 91.34 & \textbf{5.13} & \textbf{8.66} & 91.31          & 69.73          & 84.06 & \textbf{14.73} & \textbf{16.04} & 84.28          & 55.03          & 75.87          & \textbf{20.83} & \textbf{24.17} & 75.91          \\
                        & E-T5 \ours     & 91.00          & 70.36 & 20.65         & 23.02         & 3.40           & 69.73          & 35.23 & 38.72          & 57.70          & 18.95          & 55.03          & 27.61          & 28.30          & 58.42          & 10.50          \\
                        & I-Vanilla & 89.14          & 70.03 & 19.11         & 21.79         & \textbf{2.91}  & 68.23          & 32.70 & 38.85          & 58.35          & \textbf{13.49} & 42.52          & 21.53          & 21.80          & 55.84          & \textbf{4.79}  \\
                        & I-Iter \ours    & \textbf{92.20} & 72.66 & 19.54         & 21.66         & 5.58           & \textbf{70.67} & 33.17 & 38.49          & 60.59          & 18.13          & \textbf{55.38} & 26.91          & 28.86          & 59.90          & 10.52          \\
                        & I-Simul \ours   & 91.87          & 71.72 & 20.15         & 22.38         & 5.09           & 69.54          & 31.45 & 38.26          & 61.73          & 15.88          & 55.28          & 26.35          & 29.37          & 60.57          & 10.17          \\ \bottomrule
\end{tabular}

}
\caption{Results of T5's calibration performance with increasing dataset sizes. 
We observe a significant improvement in calibration performance when increasing the dataset size from small to middle.}

\label{tab:emergent_dev_size}
\end{table*}

\looseness=-1
\textbf{Learnable methods.}
Compared to unlearnable methods, learnable ones significantly mitigate the overconfidence issue, reflected in the sharp decrease in CErr$_{neg}$, indicating that learnable methods output very low confidence in wrong predictions.
But we also observe that learnable methods lower the confidence in correct predictions, resulting in increasing CErr$_{pos}$ and ECE. 
However, we highlight two observations indicating that learnable methods essentially teach models to have clearer confidence estimations, instead of roughly reducing the confidence like LS:  
(1) Compared to the vanilla version, the increase in CErr$_{pos}$ is significantly lower than the decrease in CErr$_{neg}$, especially on ID samples; 
(2) Learnable methods give obviously lower confidence in OOD samples, and the average confidence drop is highly relevant to the performance drop under distribution shifts. 
Thus, the low confidence and relatively higher CErr$_{pos}$ and ECE on OOD samples may be reasonable.




Further, we give a detailed analysis of extrinsic and intrinsic learnable methods and also compare our extended calibration methods with previous methods:
(1) For extrinsic methods, the extended E-T5 exhibits significantly better calibration performance compared to the adapted E-MLP considering the mitigation of the overconfidence issue.
The essential difference mainly lies in the extrinsic model for the calibration task.
We find that using the larger capacity model as the extrinsic calibrator shows the same trend with shifting from the vanilla fine-tuning to learnable methods. 
We further study this scaling effect in Sec.~\ref{sec:emergent};
(2) For intrinsic methods, the three different training paradigms don't show substantial differences considering the calibration performance, and none of them consistently achieves the best performance on all datasets.
As a comparison, our methods (I-Iter and I-Simul) address the degraded performance issue of I-Vanilla and make the main task performance match with the vanilla fine-tuning;
(3) Interestingly, there doesn't exist a substantial difference between the extrinsic E-T5 method and other intrinsic methods, given the same base architecture (e.g., T5). 
This finding leads us to reconsider the conclusion in \citet{lin2022teaching} that PLMs can be trained to give their uncertainty by words. 
Given the comparable performance between intrinsic and extrinsic methods, we provide an extension to this conclusion.
We identify that the success of this paradigm essentially lies in the learnable attribute of the calibration task, instead of the self-checking process of PLMs.
Namely, the findings in previous work may not only be attributed to the capability of PLMs but also the ”learnable” property of the calibration task.



\subsection{Emergent Calibration}
\label{sec:emergent}
In Sec.~\ref{sec:standard}, we identify the potential in learnable methods. 
However, a detailed exploration of learnable calibration methods is lacking. 
We conduct experiments to study the influence of two important factors, namely the dataset size and the model scale for the calibration task, on PLMs calibration.
Note that the model scale in this section considers the model adopted for the calibration task, instead of the main task.
  
\textbf{Dataset size.}
Table~\ref{tab:emergent_dev_size} shows the results of different sizes of the calibration dataset. 
Two basic findings are:
(1) The five learnable methods show a consistent trend when increasing the dataset size, indicating that the essence of these methods is the same;
(2) The size of datasets for training the calibration task doesn't have a substantial influence on PLMs performance on the main task.

Beyond these, we observe that there is a sharp difference in calibration performance when increasing the dataset size from small to middle. The trend is overall consistent with the one observed when shifting from vanilla fine-tuning to learnable calibration methods. 
The trend can be summarized as:
(1) For ID samples, we can observe a sharp decrease in CErr$_{neg}$ with relatively less negative influence on ECE and CErr$_{pos}$;
(2) For OOD samples, the CErr$_{pos}$ and ECE increase significantly along with increasing the dataset size. 
However, given the arguments in Sec.~\ref{sec:standard}, we identify that PLMs' calibration performance improves when trained on larger calibration datasets. 
Besides, we don't observe further improvement in calibration performance when increasing the dataset size from middle to large. 
This is consistent with normal task training, where increasing the dataset size doesn't increase performance after a critical point.


\textbf{Model scale.}
\looseness=-1
Table~\ref{tab:emergent_scale} shows the results of various model scales. 
Two basic findings are:
(1) The five learnable methods still show a consistent trend when scaling larger; 
(2) We observe a consistent confidence increase when scaling larger, which is similar to the trend observed in Sec.~\ref{sec:empirical_study}, where increasing capacity makes PLMs more confident. 

\looseness=-1
Surprisingly, although the confidence continues to increase, for ID samples, we observe a consistent decrease in CErr$_{pos}$ with neglectable influence on ECE and CErr$_{neg}$ when scaling larger.
Note that the dataset for the calibration task is collected from ID.
Thus, if provided enough ID samples for the calibration task training, scaling larger enables models to better learn the calibration task, ensuring better calibration performance on ID samples.
For OOD samples, we don't observe a consistent trend due to the influence of various factors.
Specifically, when using out-of-the-box to tackle OOD samples, the  problem of distribution shifts appears in the introduced calibration task.
Whether scaling the calibration-task model larger improves calibration performance under distribution shifts is determined by many factors (e.g., the dataset difficulty, the overconfidence issue in the calibration task).
We leave it for future exploration.

\section{Conclusion}
We take a close look into PLMs calibration, motivating to answer two central questions:
(1) Do PLMs learn to become calibrated in the training process? 
(2) How effective are existing calibration methods?
We present a comprehensive empirical study, including the analysis of various decisive factors and concrete calibration methods.
Besides the findings that support existing conclusions, we also provide extensions or contradictory arguments to some established conclusions. 
%

\section*{Limitations and Future Work}
We identify two limitations in our work that necessitate further investigation and improvement.
First, only empirical results are presented in our work. 
A theoretical understanding of PLMs calibration is still lacking.
Going forward, we are motivated to investigate this problem from the standpoint of feature learning.
We see great potential in unifying several problems in AI safety~\citep{DBLP:journals/corr/abs-2104-14235} from a feature-learning perspective, including spurious correlations~\citep{DBLP:conf/acl/GuWCL19, DBLP:conf/naacl/WangSY022}, robustness~\citep{DBLP:journals/corr/abs-2110-15317, DBLP:conf/acl/0002PTK22}, backdoor learning~\citep{DBLP:conf/qrs/ShengHLC22, DBLP:conf/nips/CuiYHCLS22}, and calibration~\citep{DBLP:conf/emnlp/UlmerFH22}. 
Second, we propose three simple extended calibration methods based on existing ones. 
In our experiments, we evaluate the calibration performance of existing and our calibration methods.
We make an assumption that we have a large held-out validation set that can be employed as the training dataset for the calibration task.
We demonstrate the effectiveness of learnable calibration methods in this ideal situation.
However, in practice, we need to make the decision about how to allocate the data for the main task and the calibration task given limited training samples.

\bibliography{custom}

\begin{thebibliography}{66}
\expandafter\ifx\csname natexlab\endcsname\relax\def\natexlab#1{#1}\fi

\bibitem[{Brown et~al.(2020)Brown, Mann, Ryder, Subbiah, Kaplan, Dhariwal,
  Neelakantan, Shyam, Sastry, Askell, Agarwal, Herbert{-}Voss, Krueger,
  Henighan, Child, Ramesh, Ziegler, Wu, Winter, Hesse, Chen, Sigler, Litwin,
  Gray, Chess, Clark, Berner, McCandlish, Radford, Sutskever, and
  Amodei}]{brown2020language}
Tom~B. Brown, Benjamin Mann, Nick Ryder, Melanie Subbiah, Jared Kaplan,
  Prafulla Dhariwal, Arvind Neelakantan, Pranav Shyam, Girish Sastry, Amanda
  Askell, Sandhini Agarwal, Ariel Herbert{-}Voss, Gretchen Krueger, Tom
  Henighan, Rewon Child, Aditya Ramesh, Daniel~M. Ziegler, Jeffrey Wu, Clemens
  Winter, Christopher Hesse, Mark Chen, Eric Sigler, Mateusz Litwin, Scott
  Gray, Benjamin Chess, Jack Clark, Christopher Berner, Sam McCandlish, Alec
  Radford, Ilya Sutskever, and Dario Amodei. 2020.
\newblock \href
  {https://proceedings.neurips.cc/paper/2020/hash/1457c0d6bfcb4967418bfb8ac142f64a-Abstract.html}
  {Language models are few-shot learners}.
\newblock In \emph{Advances in Neural Information Processing Systems 33: Annual
  Conference on Neural Information Processing Systems 2020, NeurIPS 2020,
  December 6-12, 2020, virtual}.

\bibitem[{Chen et~al.(2022)Chen, Gao, Cui, Qi, Huang, Liu, and
  Sun}]{DBLP:conf/emnlp/ChenGCQH0S22}
Yangyi Chen, Hongcheng Gao, Ganqu Cui, Fanchao Qi, Longtao Huang, Zhiyuan Liu,
  and Maosong Sun. 2022.
\newblock \href {https://aclanthology.org/2022.emnlp-main.771} {Why should
  adversarial perturbations be imperceptible? rethink the research paradigm in
  adversarial {NLP}}.
\newblock In \emph{Proceedings of the 2022 Conference on Empirical Methods in
  Natural Language Processing, {EMNLP} 2022, Abu Dhabi, United Arab Emirates,
  December 7-11, 2022}, pages 11222--11237. Association for Computational
  Linguistics.

\bibitem[{Cui et~al.(2022)Cui, Yuan, He, Chen, Liu, and
  Sun}]{DBLP:conf/nips/CuiYHCLS22}
Ganqu Cui, Lifan Yuan, Bingxiang He, Yangyi Chen, Zhiyuan Liu, and Maosong Sun.
  2022.
\newblock \href
  {http://papers.nips.cc/paper\_files/paper/2022/hash/2052b3e0617ecb2ce9474a6feaf422b3-Abstract-Datasets\_and\_Benchmarks.html}
  {A unified evaluation of textual backdoor learning: Frameworks and
  benchmarks}.
\newblock In \emph{NeurIPS}.

\bibitem[{Dan and Roth(2021)}]{DBLP:conf/emnlp/DanR21}
Soham Dan and Dan Roth. 2021.
\newblock \href {https://doi.org/10.18653/v1/2021.findings-emnlp.180} {On the
  effects of transformer size on in- and out-of-domain calibration}.
\newblock In \emph{Findings of the Association for Computational Linguistics:
  {EMNLP} 2021, Virtual Event / Punta Cana, Dominican Republic, 16-20 November,
  2021}, pages 2096--2101. Association for Computational Linguistics.

\bibitem[{de~Gibert et~al.(2018)de~Gibert, Perez, Garc{\'\i}a-Pablos, and
  Cuadros}]{hatespeech2018de-gibert}
Ona de~Gibert, Naiara Perez, Aitor Garc{\'\i}a-Pablos, and Montse Cuadros.
  2018.
\newblock \href {https://doi.org/10.18653/v1/W18-5102} {Hate speech dataset
  from a white supremacy forum}.
\newblock In \emph{Proceedings of the 2nd Workshop on Abusive Language Online
  ({ALW}2)}, pages 11--20, Brussels, Belgium. Association for Computational
  Linguistics.

\bibitem[{DeGroot and Fienberg(1983)}]{degroot1983comparison}
Morris~H DeGroot and Stephen~E Fienberg. 1983.
\newblock The comparison and evaluation of forecasters.
\newblock \emph{Journal of the Royal Statistical Society: Series D (The
  Statistician)}, 32(1-2):12--22.

\bibitem[{Desai and Durrett(2020)}]{DBLP:conf/emnlp/DesaiD20}
Shrey Desai and Greg Durrett. 2020.
\newblock \href {https://doi.org/10.18653/v1/2020.emnlp-main.21} {Calibration
  of pre-trained transformers}.
\newblock In \emph{Proceedings of the 2020 Conference on Empirical Methods in
  Natural Language Processing (EMNLP)}, pages 295--302, Online. Association for
  Computational Linguistics.

\bibitem[{Ding et~al.(2022)Ding, Qin, Yang, Wei, Yang, Su, Hu, Chen, Chan, Chen
  et~al.}]{ding2022delta}
Ning Ding, Yujia Qin, Guang Yang, Fuchao Wei, Zonghan Yang, Yusheng Su,
  Shengding Hu, Yulin Chen, Chi-Min Chan, Weize Chen, et~al. 2022.
\newblock \href {https://arxiv.org/abs/2203.06904} {Delta tuning: A
  comprehensive study of parameter efficient methods for pre-trained language
  models}.
\newblock \emph{ArXiv preprint}, abs/2203.06904.

\bibitem[{ElSherief et~al.(2021)ElSherief, Ziems, Muchlinski, Anupindi,
  Seybolt, De~Choudhury, and Yang}]{implicithate2021elsherief}
Mai ElSherief, Caleb Ziems, David Muchlinski, Vaishnavi Anupindi, Jordyn
  Seybolt, Munmun De~Choudhury, and Diyi Yang. 2021.
\newblock \href {https://doi.org/10.18653/v1/2021.emnlp-main.29} {Latent
  hatred: A benchmark for understanding implicit hate speech}.
\newblock In \emph{Proceedings of the 2021 Conference on Empirical Methods in
  Natural Language Processing}, pages 345--363, Online and Punta Cana,
  Dominican Republic. Association for Computational Linguistics.

\bibitem[{Gal and Ghahramani(2016)}]{DBLP:conf/icml/GalG16}
Yarin Gal and Zoubin Ghahramani. 2016.
\newblock \href {http://proceedings.mlr.press/v48/gal16.html} {Dropout as a
  bayesian approximation: Representing model uncertainty in deep learning}.
\newblock In \emph{Proceedings of the 33nd International Conference on Machine
  Learning, {ICML} 2016, New York City, NY, USA, June 19-24, 2016}, volume~48
  of \emph{{JMLR} Workshop and Conference Proceedings}, pages 1050--1059.
  JMLR.org.

\bibitem[{Gu et~al.(2019)Gu, Wang, Cho, and Li}]{DBLP:conf/acl/GuWCL19}
Jiatao Gu, Yong Wang, Kyunghyun Cho, and Victor O.~K. Li. 2019.
\newblock \href {https://doi.org/10.18653/v1/p19-1121} {Improved zero-shot
  neural machine translation via ignoring spurious correlations}.
\newblock In \emph{Proceedings of the 57th Conference of the Association for
  Computational Linguistics, {ACL} 2019, Florence, Italy, July 28- August 2,
  2019, Volume 1: Long Papers}, pages 1258--1268. Association for Computational
  Linguistics.

\bibitem[{Guo et~al.(2017)Guo, Pleiss, Sun, and
  Weinberger}]{DBLP:conf/icml/GuoPSW17}
Chuan Guo, Geoff Pleiss, Yu~Sun, and Kilian~Q. Weinberger. 2017.
\newblock \href {http://proceedings.mlr.press/v70/guo17a.html} {On calibration
  of modern neural networks}.
\newblock In \emph{Proceedings of the 34th International Conference on Machine
  Learning, {ICML} 2017, Sydney, NSW, Australia, 6-11 August 2017}, volume~70
  of \emph{Proceedings of Machine Learning Research}, pages 1321--1330. {PMLR}.

\bibitem[{Gupta et~al.(2021)Gupta, Rahimi, Ajanthan, Mensink, Sminchisescu, and
  Hartley}]{DBLP:conf/iclr/GuptaRAMS021}
Kartik Gupta, Amir Rahimi, Thalaiyasingam Ajanthan, Thomas Mensink, Cristian
  Sminchisescu, and Richard Hartley. 2021.
\newblock \href {https://openreview.net/forum?id=eQe8DEWNN2W} {Calibration of
  neural networks using splines}.
\newblock In \emph{9th International Conference on Learning Representations,
  {ICLR} 2021, Virtual Event, Austria, May 3-7, 2021}. OpenReview.net.

\bibitem[{Harris(1954)}]{harris1954distributional}
Zellig~S Harris. 1954.
\newblock Distributional structure.
\newblock \emph{Word}, 10(2-3):146--162.

\bibitem[{Hendrycks et~al.(2019{\natexlab{a}})Hendrycks, Basart, Mazeika,
  Mostajabi, Steinhardt, and Song}]{hendrycks2019scaling}
Dan Hendrycks, Steven Basart, Mantas Mazeika, Mohammadreza Mostajabi, Jacob
  Steinhardt, and Dawn Song. 2019{\natexlab{a}}.
\newblock \href {https://arxiv.org/abs/1911.11132} {Scaling out-of-distribution
  detection for real-world settings}.
\newblock \emph{ArXiv preprint}, abs/1911.11132.

\bibitem[{Hendrycks et~al.(2019{\natexlab{b}})Hendrycks, Lee, and
  Mazeika}]{hendrycks2019using}
Dan Hendrycks, Kimin Lee, and Mantas Mazeika. 2019{\natexlab{b}}.
\newblock \href {http://proceedings.mlr.press/v97/hendrycks19a.html} {Using
  pre-training can improve model robustness and uncertainty}.
\newblock In \emph{Proceedings of the 36th International Conference on Machine
  Learning, {ICML} 2019, 9-15 June 2019, Long Beach, California, {USA}},
  volume~97 of \emph{Proceedings of Machine Learning Research}, pages
  2712--2721. {PMLR}.

\bibitem[{Hendrycks et~al.(2020)Hendrycks, Mu, Cubuk, Zoph, Gilmer, and
  Lakshminarayanan}]{DBLP:conf/iclr/HendrycksMCZGL20}
Dan Hendrycks, Norman Mu, Ekin~Dogus Cubuk, Barret Zoph, Justin Gilmer, and
  Balaji Lakshminarayanan. 2020.
\newblock \href {https://openreview.net/forum?id=S1gmrxHFvB} {Augmix: {A}
  simple data processing method to improve robustness and uncertainty}.
\newblock In \emph{8th International Conference on Learning Representations,
  {ICLR} 2020, Addis Ababa, Ethiopia, April 26-30, 2020}. OpenReview.net.

\bibitem[{Hochreiter and Schmidhuber(1997)}]{hochreiter1997long}
Sepp Hochreiter and J{\"u}rgen Schmidhuber. 1997.
\newblock Long short-term memory.
\newblock \emph{Neural computation}, 9(8):1735--1780.

\bibitem[{Houben et~al.(2021)Houben, Abrecht, Akila, B{\"{a}}r, Brockherde,
  Feifel, Fingscheidt, Gannamaneni, Ghobadi, Hammam, Haselhoff, Hauser,
  Heinzemann, Hoffmann, Kapoor, Kappel, Klingner, Kronenberger, K{\"{u}}ppers,
  L{\"{o}}hdefink, Mlynarski, Mock, Mualla, Pavlitskaya, Poretschkin, Pohl,
  Kumar, Rosenzweig, Rottmann, R{\"{u}}ping, S{\"{a}}mann, Schneider, Schulz,
  Schwalbe, Sicking, Srivastava, Varghese, Weber, Wirkert, Wirtz, and
  Woehrle}]{DBLP:journals/corr/abs-2104-14235}
Sebastian Houben, Stephanie Abrecht, Maram Akila, Andreas B{\"{a}}r, Felix
  Brockherde, Patrick Feifel, Tim Fingscheidt, Sujan~Sai Gannamaneni,
  Seyed~Eghbal Ghobadi, Ahmed Hammam, Anselm Haselhoff, Felix Hauser, Christian
  Heinzemann, Marco Hoffmann, Nikhil Kapoor, Falk Kappel, Marvin Klingner, Jan
  Kronenberger, Fabian K{\"{u}}ppers, Jonas L{\"{o}}hdefink, Michael Mlynarski,
  Michael Mock, Firas Mualla, Svetlana Pavlitskaya, Maximilian Poretschkin,
  Alexander Pohl, Varun~Ravi Kumar, Julia Rosenzweig, Matthias Rottmann, Stefan
  R{\"{u}}ping, Timo S{\"{a}}mann, Jan~David Schneider, Elena Schulz, Gesina
  Schwalbe, Joachim Sicking, Toshika Srivastava, Serin Varghese, Michael Weber,
  Sebastian Wirkert, Tim Wirtz, and Matthias Woehrle. 2021.
\newblock \href {http://arxiv.org/abs/2104.14235} {Inspect, understand,
  overcome: {A} survey of practical methods for {AI} safety}.
\newblock \emph{CoRR}, abs/2104.14235.

\bibitem[{Houlsby et~al.(2019)Houlsby, Giurgiu, Jastrzebski, Morrone,
  de~Laroussilhe, Gesmundo, Attariyan, and
  Gelly}]{DBLP:conf/icml/HoulsbyGJMLGAG19}
Neil Houlsby, Andrei Giurgiu, Stanislaw Jastrzebski, Bruna Morrone, Quentin
  de~Laroussilhe, Andrea Gesmundo, Mona Attariyan, and Sylvain Gelly. 2019.
\newblock \href {http://proceedings.mlr.press/v97/houlsby19a.html}
  {Parameter-efficient transfer learning for {NLP}}.
\newblock In \emph{Proceedings of the 36th International Conference on Machine
  Learning, {ICML} 2019, 9-15 June 2019, Long Beach, California, {USA}},
  volume~97 of \emph{Proceedings of Machine Learning Research}, pages
  2790--2799. {PMLR}.

\bibitem[{Kadavath et~al.(2022)Kadavath, Conerly, Askell, Henighan, Drain,
  Perez, Schiefer, Dodds, DasSarma, Tran-Johnson et~al.}]{kadavath2022language}
Saurav Kadavath, Tom Conerly, Amanda Askell, Tom Henighan, Dawn Drain, Ethan
  Perez, Nicholas Schiefer, Zac~Hatfield Dodds, Nova DasSarma, Eli
  Tran-Johnson, et~al. 2022.
\newblock \href {https://arxiv.org/abs/2207.05221} {Language models (mostly)
  know what they know}.
\newblock \emph{ArXiv preprint}, abs/2207.05221.

\bibitem[{Kong et~al.(2020)Kong, Jiang, Zhuang, Lyu, Zhao, and
  Zhang}]{DBLP:conf/emnlp/KongJZLZZ20}
Lingkai Kong, Haoming Jiang, Yuchen Zhuang, Jie Lyu, Tuo Zhao, and Chao Zhang.
  2020.
\newblock \href {https://doi.org/10.18653/v1/2020.emnlp-main.102} {Calibrated
  language model fine-tuning for in- and out-of-distribution data}.
\newblock In \emph{Proceedings of the 2020 Conference on Empirical Methods in
  Natural Language Processing (EMNLP)}, pages 1326--1340, Online. Association
  for Computational Linguistics.

\bibitem[{Kumar et~al.(2022)Kumar, Ma, Liang, and
  Raghunathan}]{kumar2022calibrated}
Ananya Kumar, Tengyu Ma, Percy Liang, and Aditi Raghunathan. 2022.
\newblock Calibrated ensembles can mitigate accuracy tradeoffs under
  distribution shift.
\newblock In \emph{The 38th Conference on Uncertainty in Artificial
  Intelligence}.

\bibitem[{Lakshminarayanan et~al.(2017)Lakshminarayanan, Pritzel, and
  Blundell}]{DBLP:conf/nips/Lakshminarayanan17}
Balaji Lakshminarayanan, Alexander Pritzel, and Charles Blundell. 2017.
\newblock \href
  {https://proceedings.neurips.cc/paper/2017/hash/9ef2ed4b7fd2c810847ffa5fa85bce38-Abstract.html}
  {Simple and scalable predictive uncertainty estimation using deep ensembles}.
\newblock In \emph{Advances in Neural Information Processing Systems 30: Annual
  Conference on Neural Information Processing Systems 2017, December 4-9, 2017,
  Long Beach, CA, {USA}}, pages 6402--6413.

\bibitem[{Lester et~al.(2021)Lester, Al{-}Rfou, and
  Constant}]{DBLP:conf/emnlp/LesterAC21}
Brian Lester, Rami Al{-}Rfou, and Noah Constant. 2021.
\newblock \href {https://doi.org/10.18653/v1/2021.emnlp-main.243} {The power of
  scale for parameter-efficient prompt tuning}.
\newblock In \emph{Proceedings of the 2021 Conference on Empirical Methods in
  Natural Language Processing, {EMNLP} 2021, Virtual Event / Punta Cana,
  Dominican Republic, 7-11 November, 2021}, pages 3045--3059. Association for
  Computational Linguistics.

\bibitem[{Lin et~al.(2022)Lin, Hilton, and Evans}]{lin2022teaching}
Stephanie Lin, Jacob Hilton, and Owain Evans. 2022.
\newblock \href {https://arxiv.org/abs/2205.14334} {Teaching models to express
  their uncertainty in words}.
\newblock \emph{ArXiv preprint}, abs/2205.14334.

\bibitem[{Liu et~al.(2021)Liu, Yuan, Fu, Jiang, Hayashi, and
  Neubig}]{liu2021pre}
Pengfei Liu, Weizhe Yuan, Jinlan Fu, Zhengbao Jiang, Hiroaki Hayashi, and
  Graham Neubig. 2021.
\newblock Pre-train, prompt, and predict: A systematic survey of prompting
  methods in natural language processing.
\newblock \emph{arXiv preprint arXiv:2107.13586}.

\bibitem[{Liu et~al.(2019)Liu, Ott, Goyal, Du, Joshi, Chen, Levy, Lewis,
  Zettlemoyer, and Stoyanov}]{liu2019roberta}
Yinhan Liu, Myle Ott, Naman Goyal, Jingfei Du, Mandar Joshi, Danqi Chen, Omer
  Levy, Mike Lewis, Luke Zettlemoyer, and Veselin Stoyanov. 2019.
\newblock Roberta: A robustly optimized bert pretraining approach.
\newblock \emph{arXiv preprint arXiv:1907.11692}.

\bibitem[{Luhn(1957)}]{luhn1957statistical}
Hans~Peter Luhn. 1957.
\newblock A statistical approach to mechanized encoding and searching of
  literary information.
\newblock \emph{IBM Journal of research and development}, 1(4):309--317.

\bibitem[{McAuley and Leskovec(2013)}]{amazon2013mcauley}
Julian~J. McAuley and Jure Leskovec. 2013.
\newblock \href {https://doi.org/10.1145/2488388.2488466} {From amateurs to
  connoisseurs: modeling the evolution of user expertise through online
  reviews}.
\newblock In \emph{22nd International World Wide Web Conference, {WWW} '13, Rio
  de Janeiro, Brazil, May 13-17, 2013}, pages 897--908. International World
  Wide Web Conferences Steering Committee / {ACM}.

\bibitem[{McCoy et~al.(2019)McCoy, Pavlick, and Linzen}]{hans2019mccoy}
Tom McCoy, Ellie Pavlick, and Tal Linzen. 2019.
\newblock \href {https://doi.org/10.18653/v1/P19-1334} {Right for the wrong
  reasons: Diagnosing syntactic heuristics in natural language inference}.
\newblock In \emph{Proceedings of the 57th Annual Meeting of the Association
  for Computational Linguistics}, pages 3428--3448, Florence, Italy.
  Association for Computational Linguistics.

\bibitem[{Minderer et~al.(2021)Minderer, Djolonga, Romijnders, Hubis, Zhai,
  Houlsby, Tran, and Lucic}]{DBLP:conf/nips/MindererDRHZHTL21}
Matthias Minderer, Josip Djolonga, Rob Romijnders, Frances Hubis, Xiaohua Zhai,
  Neil Houlsby, Dustin Tran, and Mario Lucic. 2021.
\newblock \href
  {https://proceedings.neurips.cc/paper/2021/hash/8420d359404024567b5aefda1231af24-Abstract.html}
  {Revisiting the calibration of modern neural networks}.
\newblock In \emph{Advances in Neural Information Processing Systems 34: Annual
  Conference on Neural Information Processing Systems 2021, NeurIPS 2021,
  December 6-14, 2021, virtual}, pages 15682--15694.

\bibitem[{Naeini et~al.(2015)Naeini, Cooper, and
  Hauskrecht}]{DBLP:conf/aaai/NaeiniCH15}
Mahdi~Pakdaman Naeini, Gregory~F. Cooper, and Milos Hauskrecht. 2015.
\newblock \href {http://www.aaai.org/ocs/index.php/AAAI/AAAI15/paper/view/9667}
  {Obtaining well calibrated probabilities using bayesian binning}.
\newblock In \emph{Proceedings of the Twenty-Ninth {AAAI} Conference on
  Artificial Intelligence, January 25-30, 2015, Austin, Texas, {USA}}, pages
  2901--2907. {AAAI} Press.

\bibitem[{Nakov et~al.(2013)Nakov, Rosenthal, Kozareva, Stoyanov, Ritter, and
  Wilson}]{semeval2016nakov}
Preslav Nakov, Sara Rosenthal, Zornitsa Kozareva, Veselin Stoyanov, Alan
  Ritter, and Theresa Wilson. 2013.
\newblock \href {https://aclanthology.org/S13-2052} {{S}em{E}val-2013 task 2:
  Sentiment analysis in {T}witter}.
\newblock In \emph{Second Joint Conference on Lexical and Computational
  Semantics (*{SEM}), Volume 2: Proceedings of the Seventh International
  Workshop on Semantic Evaluation ({S}em{E}val 2013)}, pages 312--320, Atlanta,
  Georgia, USA. Association for Computational Linguistics.

\bibitem[{Nie et~al.(2020)Nie, Williams, Dinan, Bansal, Weston, and
  Kiela}]{anli2020nie}
Yixin Nie, Adina Williams, Emily Dinan, Mohit Bansal, Jason Weston, and Douwe
  Kiela. 2020.
\newblock \href {https://doi.org/10.18653/v1/2020.acl-main.441} {Adversarial
  {NLI}: A new benchmark for natural language understanding}.
\newblock In \emph{Proceedings of the 58th Annual Meeting of the Association
  for Computational Linguistics}, pages 4885--4901, Online. Association for
  Computational Linguistics.

\bibitem[{Nixon et~al.(2019)Nixon, Dusenberry, Zhang, Jerfel, and
  Tran}]{DBLP:conf/cvpr/NixonDZJT19}
Jeremy Nixon, Michael~W. Dusenberry, Linchuan Zhang, Ghassen Jerfel, and Dustin
  Tran. 2019.
\newblock \href
  {http://openaccess.thecvf.com/content\_CVPRW\_2019/html/Uncertainty\_and\_Robustness\_in\_Deep\_Visual\_Learning/Nixon\_Measuring\_Calibration\_in\_Deep\_Learning\_CVPRW\_2019\_paper.html}
  {Measuring calibration in deep learning}.
\newblock In \emph{{IEEE} Conference on Computer Vision and Pattern Recognition
  Workshops, {CVPR} Workshops 2019, Long Beach, CA, USA, June 16-20, 2019},
  pages 38--41. Computer Vision Foundation / {IEEE}.

\bibitem[{Pereyra et~al.(2017)Pereyra, Tucker, Chorowski, Kaiser, and
  Hinton}]{DBLP:conf/iclr/PereyraTCKH17}
Gabriel Pereyra, George Tucker, Jan Chorowski, Lukasz Kaiser, and Geoffrey~E.
  Hinton. 2017.
\newblock \href {https://openreview.net/forum?id=HyhbYrGYe} {Regularizing
  neural networks by penalizing confident output distributions}.
\newblock In \emph{5th International Conference on Learning Representations,
  {ICLR} 2017, Toulon, France, April 24-26, 2017, Workshop Track Proceedings}.
  OpenReview.net.

\bibitem[{Platt et~al.(1999)}]{platt1999probabilistic}
John Platt et~al. 1999.
\newblock Probabilistic outputs for support vector machines and comparisons to
  regularized likelihood methods.
\newblock \emph{Advances in large margin classifiers}, 10(3):61--74.

\bibitem[{Potts et~al.(2021)Potts, Wu, Geiger, and Kiela}]{dynasent2021potts}
Christopher Potts, Zhengxuan Wu, Atticus Geiger, and Douwe Kiela. 2021.
\newblock \href {https://doi.org/10.18653/v1/2021.acl-long.186} {{D}yna{S}ent:
  A dynamic benchmark for sentiment analysis}.
\newblock In \emph{Proceedings of the 59th Annual Meeting of the Association
  for Computational Linguistics and the 11th International Joint Conference on
  Natural Language Processing (Volume 1: Long Papers)}, pages 2388--2404,
  Online. Association for Computational Linguistics.

\bibitem[{Raffel et~al.(2020)Raffel, Shazeer, Roberts, Lee, Narang, Matena,
  Zhou, Li, and Liu}]{DBLP:journals/jmlr/RaffelSRLNMZLL20}
Colin Raffel, Noam Shazeer, Adam Roberts, Katherine Lee, Sharan Narang, Michael
  Matena, Yanqi Zhou, Wei Li, and Peter~J. Liu. 2020.
\newblock \href {http://jmlr.org/papers/v21/20-074.html} {Exploring the limits
  of transfer learning with a unified text-to-text transformer}.
\newblock \emph{J. Mach. Learn. Res.}, 21:140:1--140:67.

\bibitem[{Rizve et~al.(2021)Rizve, Duarte, Rawat, and Shah}]{rizve2021defense}
Mamshad~Nayeem Rizve, Kevin Duarte, Yogesh~S Rawat, and Mubarak Shah. 2021.
\newblock \href {https://arxiv.org/abs/2101.06329} {In defense of
  pseudo-labeling: An uncertainty-aware pseudo-label selection framework for
  semi-supervised learning}.
\newblock \emph{ArXiv preprint}, abs/2101.06329.

\bibitem[{Sheng et~al.(2022)Sheng, Han, Li, and
  Chang}]{DBLP:conf/qrs/ShengHLC22}
Xuan Sheng, Zhaoyang Han, Piji Li, and Xiangmao Chang. 2022.
\newblock \href {https://doi.org/10.1109/QRS57517.2022.00086} {A survey on
  backdoor attack and defense in natural language processing}.
\newblock In \emph{22nd {IEEE} International Conference on Software Quality,
  Reliability and Security, {QRS} 2022, Guangzhou, China, December 5-9, 2022},
  pages 809--820. {IEEE}.

\bibitem[{Si et~al.(2022)Si, Zhao, Min, and Boyd-Graber}]{si2022revisiting}
Chenglei Si, Chen Zhao, Sewon Min, and Jordan Boyd-Graber. 2022.
\newblock \href {https://arxiv.org/abs/2205.12507} {Revisiting calibration for
  question answering}.
\newblock \emph{ArXiv preprint}, abs/2205.12507.

\bibitem[{Snoek et~al.(2019)Snoek, Ovadia, Fertig, Lakshminarayanan, Nowozin,
  Sculley, Dillon, Ren, and Nado}]{DBLP:conf/nips/SnoekOFLNSDRN19}
Jasper Snoek, Yaniv Ovadia, Emily Fertig, Balaji Lakshminarayanan, Sebastian
  Nowozin, D.~Sculley, Joshua~V. Dillon, Jie Ren, and Zachary Nado. 2019.
\newblock \href
  {https://proceedings.neurips.cc/paper/2019/hash/8558cb408c1d76621371888657d2eb1d-Abstract.html}
  {Can you trust your model's uncertainty? evaluating predictive uncertainty
  under dataset shift}.
\newblock In \emph{Advances in Neural Information Processing Systems 32: Annual
  Conference on Neural Information Processing Systems 2019, NeurIPS 2019,
  December 8-14, 2019, Vancouver, BC, Canada}, pages 13969--13980.

\bibitem[{Socher et~al.(2013{\natexlab{a}})Socher, Perelygin, Wu, Chuang,
  Manning, Ng, and Potts}]{DBLP:conf/emnlp/SocherPWCMNP13}
Richard Socher, Alex Perelygin, Jean Wu, Jason Chuang, Christopher~D. Manning,
  Andrew Ng, and Christopher Potts. 2013{\natexlab{a}}.
\newblock \href {https://aclanthology.org/D13-1170} {Recursive deep models for
  semantic compositionality over a sentiment treebank}.
\newblock In \emph{Proceedings of the 2013 Conference on Empirical Methods in
  Natural Language Processing}, pages 1631--1642, Seattle, Washington, USA.
  Association for Computational Linguistics.

\bibitem[{Socher et~al.(2013{\natexlab{b}})Socher, Perelygin, Wu, Chuang,
  Manning, Ng, and Potts}]{sst2013socher}
Richard Socher, Alex Perelygin, Jean Wu, Jason Chuang, Christopher~D. Manning,
  Andrew Ng, and Christopher Potts. 2013{\natexlab{b}}.
\newblock \href {https://aclanthology.org/D13-1170} {Recursive deep models for
  semantic compositionality over a sentiment treebank}.
\newblock In \emph{Proceedings of the 2013 Conference on Empirical Methods in
  Natural Language Processing}, pages 1631--1642, Seattle, Washington, USA.
  Association for Computational Linguistics.

\bibitem[{Srivastava et~al.(2022)Srivastava, Rastogi, Rao, Shoeb, Abid, Fisch,
  Brown, Santoro, Gupta, Garriga-Alonso et~al.}]{srivastava2022beyond}
Aarohi Srivastava, Abhinav Rastogi, Abhishek Rao, Abu Awal~Md Shoeb, Abubakar
  Abid, Adam Fisch, Adam~R Brown, Adam Santoro, Aditya Gupta, Adri{\`a}
  Garriga-Alonso, et~al. 2022.
\newblock \href {https://arxiv.org/abs/2206.04615} {Beyond the imitation game:
  Quantifying and extrapolating the capabilities of language models}.
\newblock \emph{ArXiv preprint}, abs/2206.04615.

\bibitem[{Szegedy et~al.(2016)Szegedy, Vanhoucke, Ioffe, Shlens, and
  Wojna}]{DBLP:conf/cvpr/SzegedyVISW16}
Christian Szegedy, Vincent Vanhoucke, Sergey Ioffe, Jonathon Shlens, and
  Zbigniew Wojna. 2016.
\newblock \href {https://doi.org/10.1109/CVPR.2016.308} {Rethinking the
  inception architecture for computer vision}.
\newblock In \emph{2016 {IEEE} Conference on Computer Vision and Pattern
  Recognition, {CVPR} 2016, Las Vegas, NV, USA, June 27-30, 2016}, pages
  2818--2826. {IEEE} Computer Society.

\bibitem[{Ulmer et~al.(2022)Ulmer, Frellsen, and
  Hardmeier}]{DBLP:conf/emnlp/UlmerFH22}
Dennis Ulmer, Jes Frellsen, and Christian Hardmeier. 2022.
\newblock \href {https://aclanthology.org/2022.findings-emnlp.198} {Exploring
  predictive uncertainty and calibration in {NLP:} {A} study on the impact of
  method {\&} data scarcity}.
\newblock In \emph{Findings of the Association for Computational Linguistics:
  {EMNLP} 2022, Abu Dhabi, United Arab Emirates, December 7-11, 2022}, pages
  2707--2735. Association for Computational Linguistics.

\bibitem[{Vaicenavicius et~al.(2019)Vaicenavicius, Widmann, Andersson,
  Lindsten, Roll, and Sch{\"{o}}n}]{DBLP:conf/aistats/VaicenaviciusWA19}
Juozas Vaicenavicius, David Widmann, Carl~R. Andersson, Fredrik Lindsten, Jacob
  Roll, and Thomas~B. Sch{\"{o}}n. 2019.
\newblock \href {http://proceedings.mlr.press/v89/vaicenavicius19a.html}
  {Evaluating model calibration in classification}.
\newblock In \emph{The 22nd International Conference on Artificial Intelligence
  and Statistics, {AISTATS} 2019, 16-18 April 2019, Naha, Okinawa, Japan},
  volume~89 of \emph{Proceedings of Machine Learning Research}, pages
  3459--3467. {PMLR}.

\bibitem[{Varshney et~al.(2022)Varshney, Mishra, and
  Baral}]{varshney2022investigating}
Neeraj Varshney, Swaroop Mishra, and Chitta Baral. 2022.
\newblock \href {https://arxiv.org/abs/2203.00211} {Investigating selective
  prediction approaches across several tasks in iid, ood, and adversarial
  settings}.
\newblock \emph{ArXiv preprint}, abs/2203.00211.

\bibitem[{Wang et~al.(2019)Wang, Singh, Michael, Hill, Levy, and
  Bowman}]{DBLP:conf/iclr/WangSMHLB19}
Alex Wang, Amanpreet Singh, Julian Michael, Felix Hill, Omer Levy, and
  Samuel~R. Bowman. 2019.
\newblock \href {https://openreview.net/forum?id=rJ4km2R5t7} {{GLUE:} {A}
  multi-task benchmark and analysis platform for natural language
  understanding}.
\newblock In \emph{7th International Conference on Learning Representations,
  {ICLR} 2019, New Orleans, LA, USA, May 6-9, 2019}. OpenReview.net.

\bibitem[{Wang et~al.(2021)Wang, Xiao, Kossaifi, Yu, Anandkumar, and
  Wang}]{DBLP:conf/nips/WangXKYAW21}
Haotao Wang, Chaowei Xiao, Jean Kossaifi, Zhiding Yu, Anima Anandkumar, and
  Zhangyang Wang. 2021.
\newblock \href
  {https://proceedings.neurips.cc/paper/2021/hash/01e9565cecc4e989123f9620c1d09c09-Abstract.html}
  {Augmax: Adversarial composition of random augmentations for robust
  training}.
\newblock In \emph{Advances in Neural Information Processing Systems 34: Annual
  Conference on Neural Information Processing Systems 2021, NeurIPS 2021,
  December 6-14, 2021, virtual}, pages 237--250.

\bibitem[{Wang et~al.(2022)Wang, Sridhar, Yang, and
  Wang}]{DBLP:conf/naacl/WangSY022}
Tianlu Wang, Rohit Sridhar, Diyi Yang, and Xuezhi Wang. 2022.
\newblock \href {https://doi.org/10.18653/v1/2022.findings-naacl.130}
  {Identifying and mitigating spurious correlations for improving robustness in
  {NLP} models}.
\newblock In \emph{Findings of the Association for Computational Linguistics:
  {NAACL} 2022, Seattle, WA, United States, July 10-15, 2022}, pages
  1719--1729. Association for Computational Linguistics.

\bibitem[{Wei and Zou(2019)}]{DBLP:conf/emnlp/WeiZ19}
Jason Wei and Kai Zou. 2019.
\newblock \href {https://doi.org/10.18653/v1/D19-1670} {{EDA}: Easy data
  augmentation techniques for boosting performance on text classification
  tasks}.
\newblock In \emph{Proceedings of the 2019 Conference on Empirical Methods in
  Natural Language Processing and the 9th International Joint Conference on
  Natural Language Processing (EMNLP-IJCNLP)}, pages 6382--6388, Hong Kong,
  China. Association for Computational Linguistics.

\bibitem[{Williams et~al.(2018{\natexlab{a}})Williams, Nangia, and
  Bowman}]{DBLP:conf/naacl/WilliamsNB18}
Adina Williams, Nikita Nangia, and Samuel Bowman. 2018{\natexlab{a}}.
\newblock \href {https://doi.org/10.18653/v1/N18-1101} {A broad-coverage
  challenge corpus for sentence understanding through inference}.
\newblock In \emph{Proceedings of the 2018 Conference of the North {A}merican
  Chapter of the Association for Computational Linguistics: Human Language
  Technologies, Volume 1 (Long Papers)}, pages 1112--1122, New Orleans,
  Louisiana. Association for Computational Linguistics.

\bibitem[{Williams et~al.(2018{\natexlab{b}})Williams, Nangia, and
  Bowman}]{mnli2018williams}
Adina Williams, Nikita Nangia, and Samuel Bowman. 2018{\natexlab{b}}.
\newblock \href {https://doi.org/10.18653/v1/N18-1101} {A broad-coverage
  challenge corpus for sentence understanding through inference}.
\newblock In \emph{Proceedings of the 2018 Conference of the North {A}merican
  Chapter of the Association for Computational Linguistics: Human Language
  Technologies, Volume 1 (Long Papers)}, pages 1112--1122, New Orleans,
  Louisiana. Association for Computational Linguistics.

\bibitem[{Xiao et~al.(2022)Xiao, Liang, Bhatt, Neiswanger, Salakhutdinov, and
  Morency}]{xiao2022uncertainty}
Yuxin Xiao, Paul~Pu Liang, Umang Bhatt, Willie Neiswanger, Ruslan
  Salakhutdinov, and Louis-Philippe Morency. 2022.
\newblock Uncertainty quantification with pre-trained language models: A
  large-scale empirical analysis.
\newblock \emph{arXiv preprint arXiv:2210.04714}.

\bibitem[{Ye and Durrett(2022)}]{DBLP:conf/acl/YeD22}
Xi~Ye and Greg Durrett. 2022.
\newblock \href {https://doi.org/10.18653/v1/2022.acl-long.429} {Can
  explanations be useful for calibrating black box models?}
\newblock In \emph{Proceedings of the 60th Annual Meeting of the Association
  for Computational Linguistics (Volume 1: Long Papers), {ACL} 2022, Dublin,
  Ireland, May 22-27, 2022}, pages 6199--6212. Association for Computational
  Linguistics.

\bibitem[{Yuan et~al.(2021)Yuan, Zhang, Chen, and
  Wei}]{DBLP:journals/corr/abs-2110-15317}
Lifan Yuan, Yichi Zhang, Yangyi Chen, and Wei Wei. 2021.
\newblock \href {http://arxiv.org/abs/2110.15317} {Bridge the gap between {CV}
  and nlp! {A} gradient-based textual adversarial attack framework}.
\newblock \emph{CoRR}, abs/2110.15317.

\bibitem[{Zaken et~al.(2022)Zaken, Goldberg, and
  Ravfogel}]{DBLP:conf/acl/ZakenGR22}
Elad~Ben Zaken, Yoav Goldberg, and Shauli Ravfogel. 2022.
\newblock \href {https://aclanthology.org/2022.acl-short.1} {Bitfit: Simple
  parameter-efficient fine-tuning for transformer-based masked
  language-models}.
\newblock In \emph{Proceedings of the 60th Annual Meeting of the Association
  for Computational Linguistics (Volume 2: Short Papers), {ACL} 2022, Dublin,
  Ireland, May 22-27, 2022}, pages 1--9. Association for Computational
  Linguistics.

\bibitem[{Zhang et~al.(2021{\natexlab{a}})Zhang, Bengio, Hardt, Recht, and
  Vinyals}]{zhang2021understanding}
Chiyuan Zhang, Samy Bengio, Moritz Hardt, Benjamin Recht, and Oriol Vinyals.
  2021{\natexlab{a}}.
\newblock Understanding deep learning (still) requires rethinking
  generalization.
\newblock \emph{Communications of the ACM}, 64(3):107--115.

\bibitem[{Zhang et~al.(2021{\natexlab{b}})Zhang, Gong, and
  Choi}]{DBLP:conf/acl/ZhangGC21}
Shujian Zhang, Chengyue Gong, and Eunsol Choi. 2021{\natexlab{b}}.
\newblock \href {https://doi.org/10.18653/v1/2021.findings-acl.172} {Knowing
  more about questions can help: Improving calibration in question answering}.
\newblock In \emph{Findings of the Association for Computational Linguistics:
  ACL-IJCNLP 2021}, pages 1958--1970, Online. Association for Computational
  Linguistics.

\bibitem[{Zhang et~al.(2015)Zhang, Zhao, and LeCun}]{DBLP:conf/nips/ZhangZL15}
Xiang Zhang, Junbo~Jake Zhao, and Yann LeCun. 2015.
\newblock \href
  {https://proceedings.neurips.cc/paper/2015/hash/250cf8b51c773f3f8dc8b4be867a9a02-Abstract.html}
  {Character-level convolutional networks for text classification}.
\newblock In \emph{Advances in Neural Information Processing Systems 28: Annual
  Conference on Neural Information Processing Systems 2015, December 7-12,
  2015, Montreal, Quebec, Canada}, pages 649--657.

\bibitem[{Zhang et~al.(2022)Zhang, Pan, Tan, and Kan}]{DBLP:conf/acl/0002PTK22}
Yunxiang Zhang, Liangming Pan, Samson Tan, and Min{-}Yen Kan. 2022.
\newblock \href {https://doi.org/10.18653/v1/2022.findings-acl.315}
  {Interpreting the robustness of neural {NLP} models to textual
  perturbations}.
\newblock In \emph{Findings of the Association for Computational Linguistics:
  {ACL} 2022, Dublin, Ireland, May 22-27, 2022}, pages 3993--4007. Association
  for Computational Linguistics.

\bibitem[{Zhu et~al.(2015)Zhu, Kiros, Zemel, Salakhutdinov, Urtasun, Torralba,
  and Fidler}]{zhu2015bookcorpus}
Yukun Zhu, Ryan Kiros, Richard~S. Zemel, Ruslan Salakhutdinov, Raquel Urtasun,
  Antonio Torralba, and Sanja Fidler. 2015.
\newblock \href {https://doi.org/10.1109/ICCV.2015.11} {Aligning books and
  movies: Towards story-like visual explanations by watching movies and reading
  books}.
\newblock In \emph{2015 {IEEE} International Conference on Computer Vision,
  {ICCV} 2015, Santiago, Chile, December 7-13, 2015}, pages 19--27. {IEEE}
  Computer Society.

\end{thebibliography}
\bibliographystyle{acl_natbib}

\appendix
\newpage

\begin{figure*}[t]
    \centering
    \begin{subfigure}[b]{0.3\textwidth}
    \centering
    \includegraphics[trim=0 0 0 0, clip, width=\textwidth]{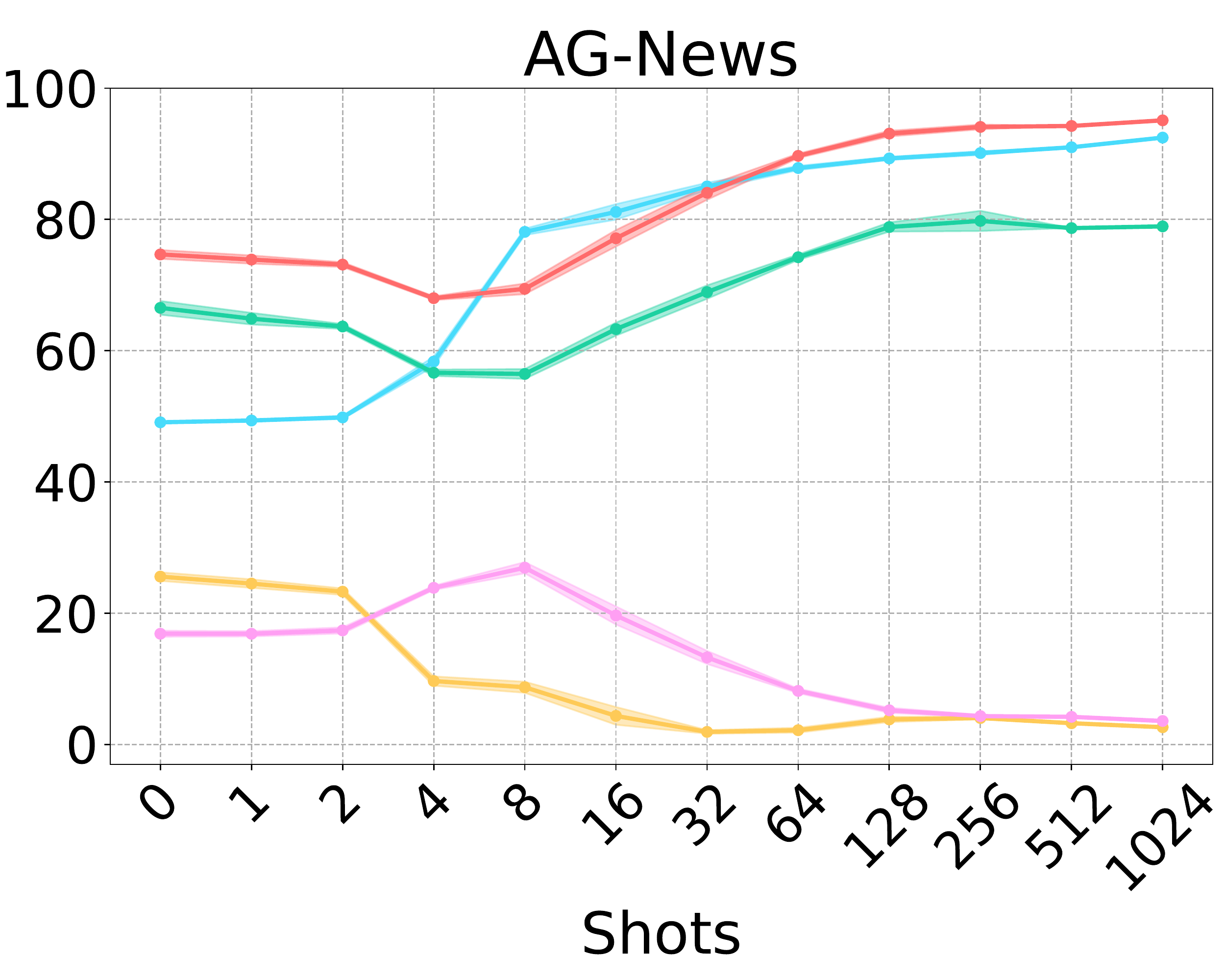}
     \label{fig:kshots-t5-agnews}
    \end{subfigure}
    \begin{subfigure}[b]{0.32\textwidth}
    \centering
    \includegraphics[trim=0 0 0 0, clip,width=\textwidth]{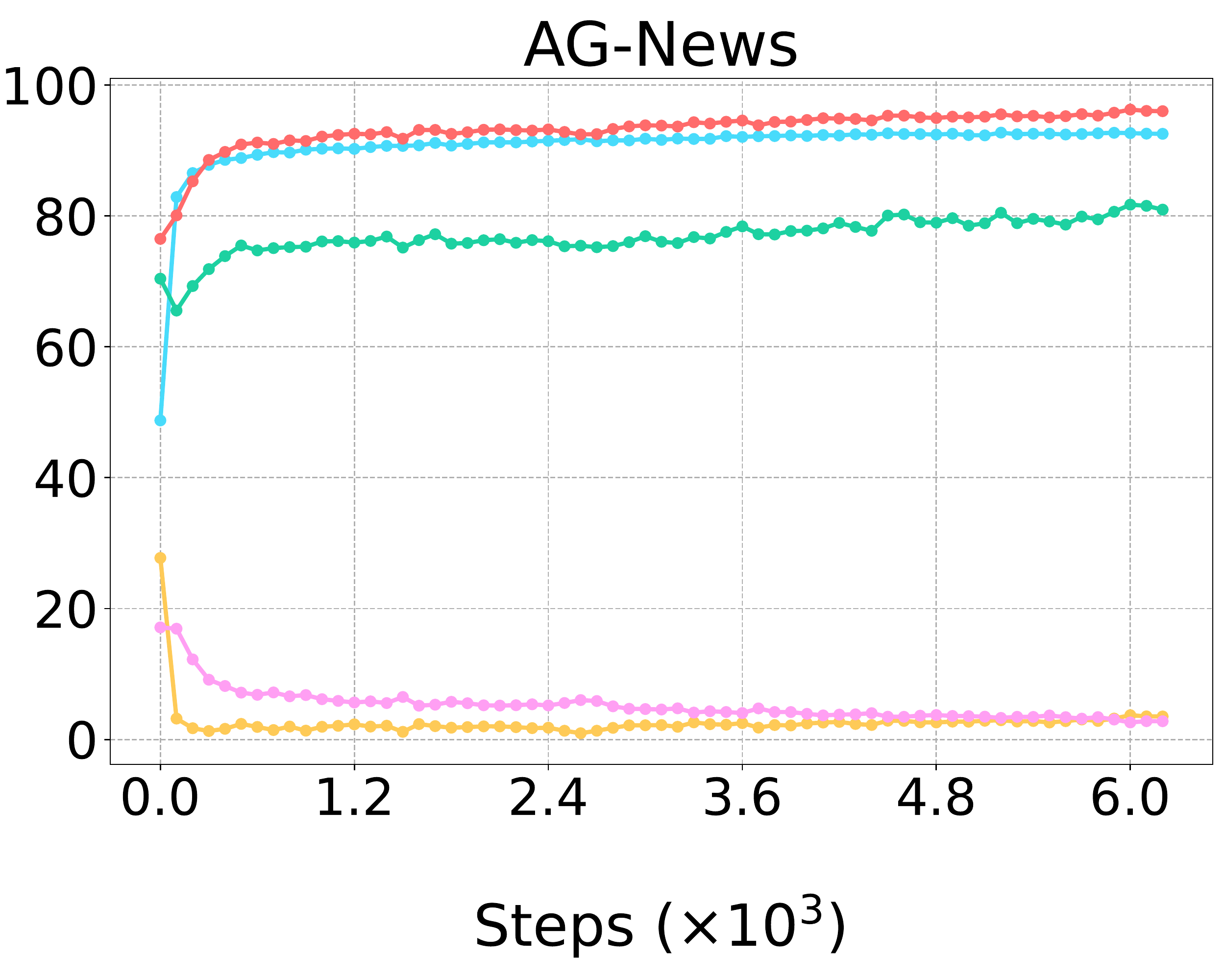}
     \label{fig:dynamics-t5-agnews}
    \end{subfigure}     
     \begin{subfigure}[b]{0.322\textwidth}
     \includegraphics[trim=0 0 0 0, clip,width=\textwidth]{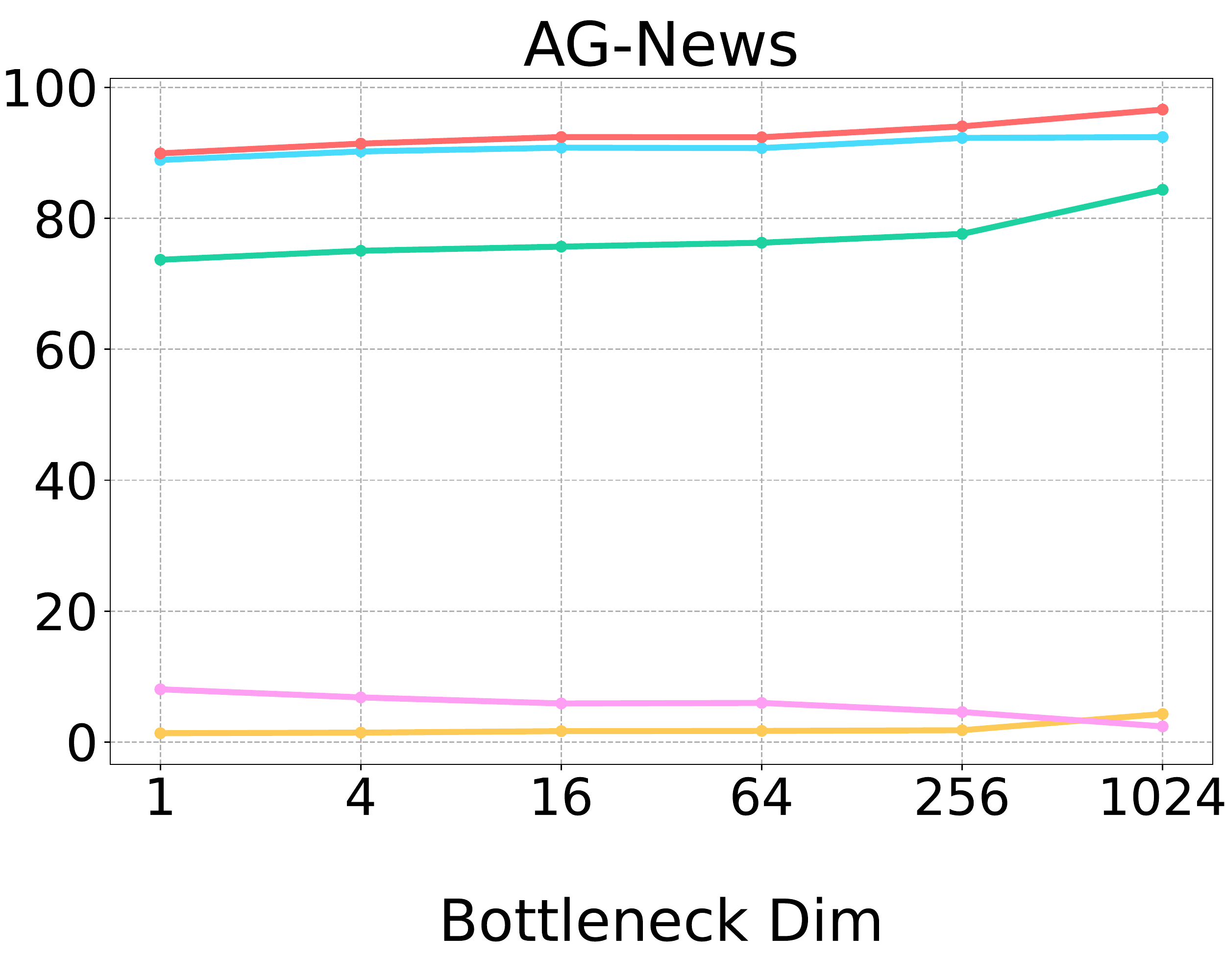}
     \label{fig:adapter-t5-agnews}
     \end{subfigure}

    \vspace{-7pt}
     \begin{subfigure}[b]{0.6\textwidth}
         \centering
         \includegraphics[trim=0 0 0 0, clip, width=\textwidth]{figures/legend.pdf}
     \end{subfigure}
     
     \vspace{-6pt}
     \caption{Additional results of T5 on AG-News including the influence of the number of training samples, training steps, and the tunable parameters number.}
     \label{fig:t5-agnews}
\end{figure*}

\begin{figure*}
\centering
    \begin{minipage}[b]{0.46\textwidth}
        \includegraphics[trim=0 0 0 0 , clip,width=0.93\textwidth]{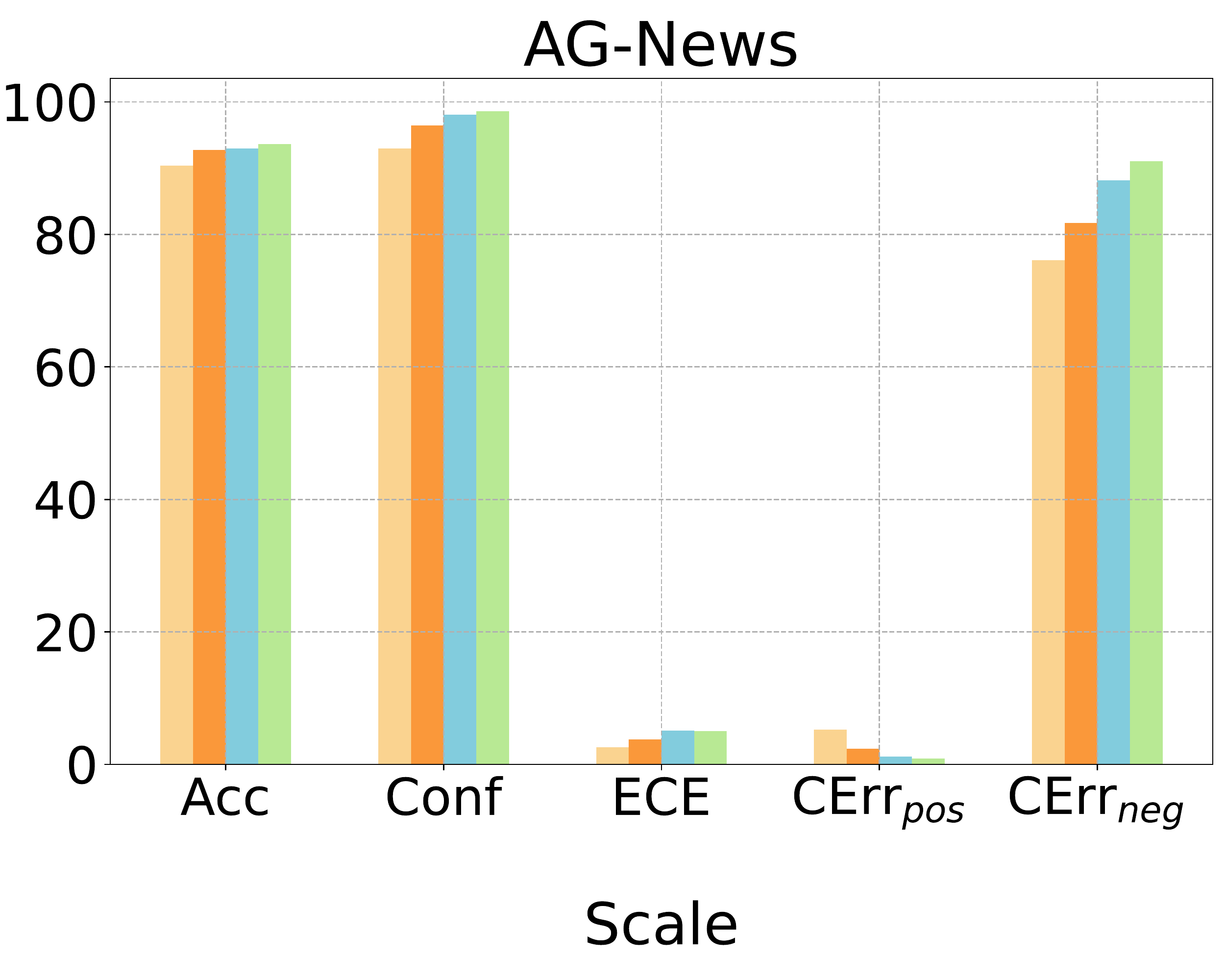}
        \caption{Additional results of different PLMs scales with T5 on AG-News.}
         \label{fig:scale-t5-agnews}
    \end{minipage}
    \quad
    \begin{minipage}[b]{0.43\textwidth}
        \includegraphics[width=\textwidth]{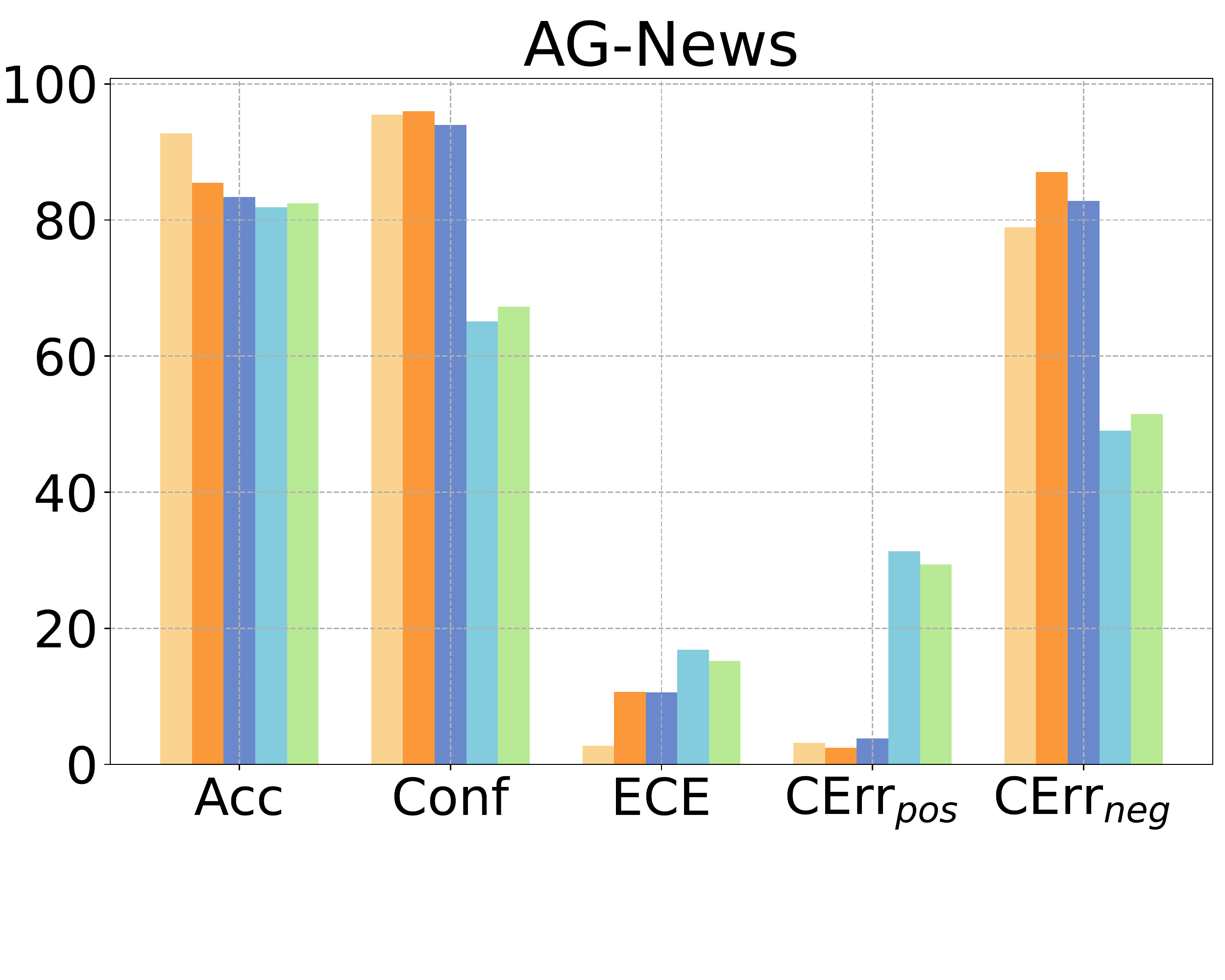}
        \caption{Additional results of the pretraining influence with T5 on AG-News.}
         \label{fig:pretrain-t5-agnews}
    \end{minipage}
\end{figure*}

\begin{figure*}[!h]
     \centering
     \begin{subfigure}[b]{0.32\textwidth}
         \centering
         \includegraphics[trim=0 0 0 0, clip, width=\textwidth]{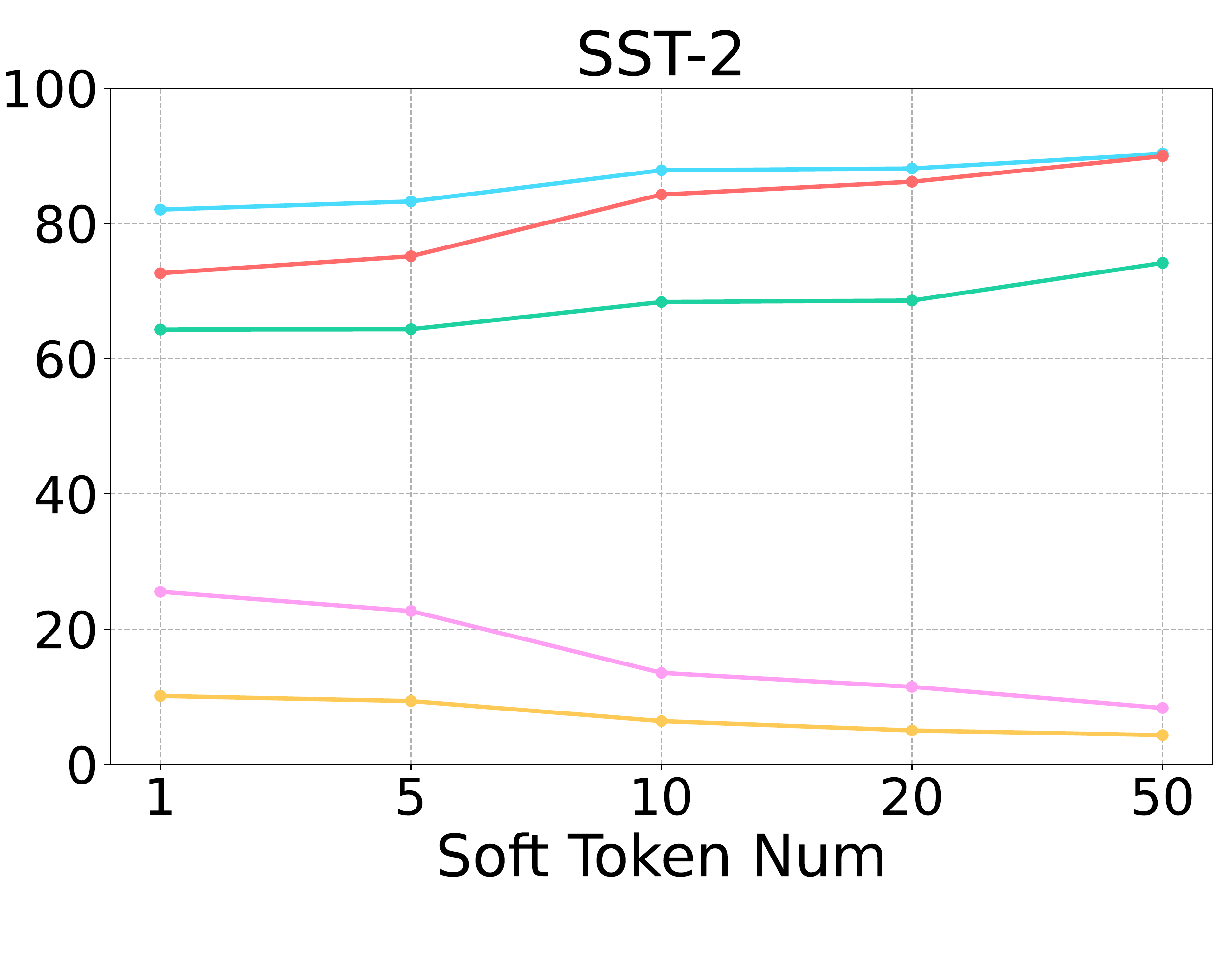}
         \label{fig:soft_prompt-t5-sst2}
     \end{subfigure}
     \begin{subfigure}[b]{0.32\textwidth}
         \centering
         \includegraphics[trim=0 0 0 0, clip,width=\textwidth]{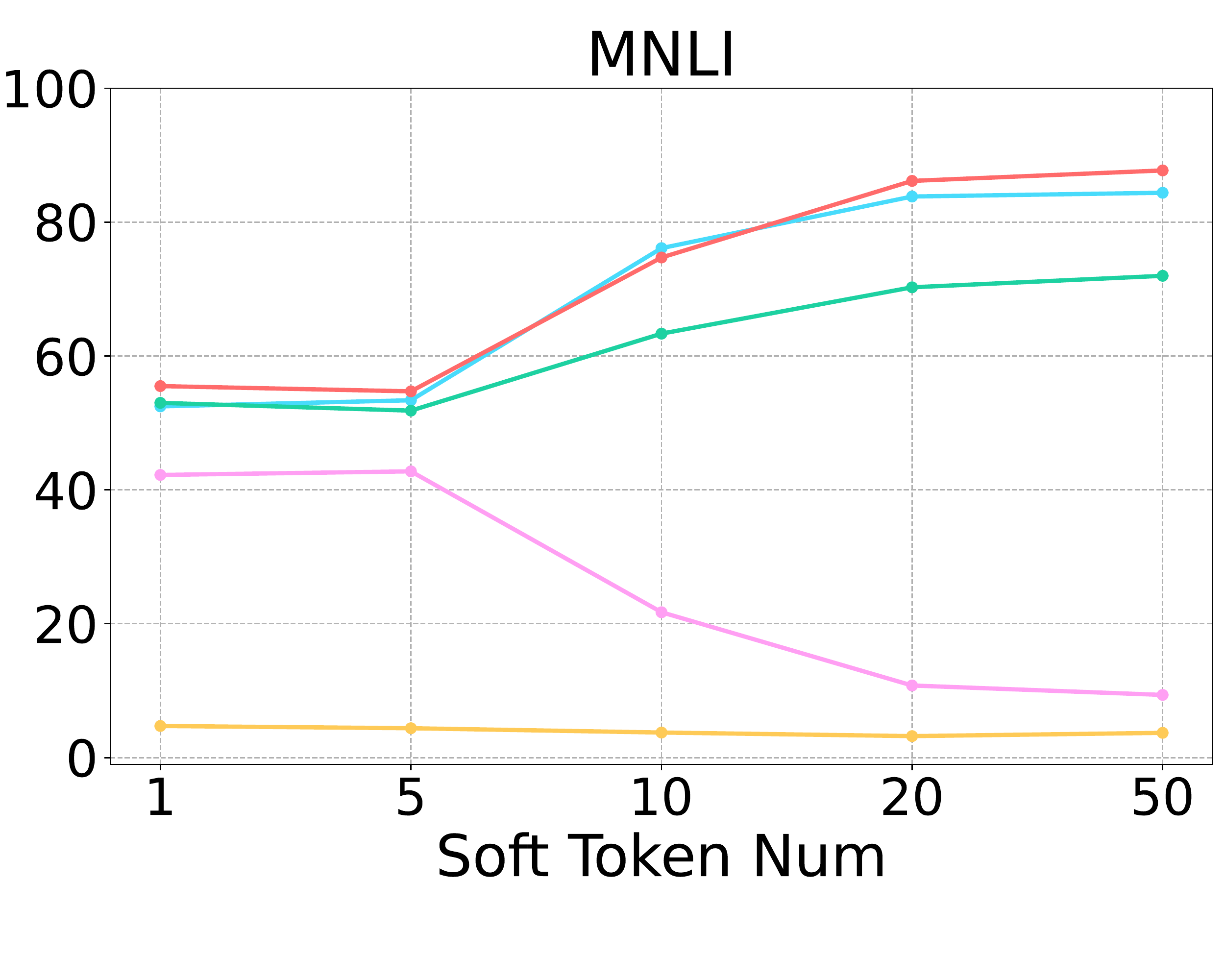}
         \label{fig:soft_prompt-t5-mnli}
     \end{subfigure}

     \vspace{-5pt}
     \begin{subfigure}[b]{0.32\textwidth}
         \centering
         \includegraphics[trim=0 0 0 10, clip,width=\textwidth]{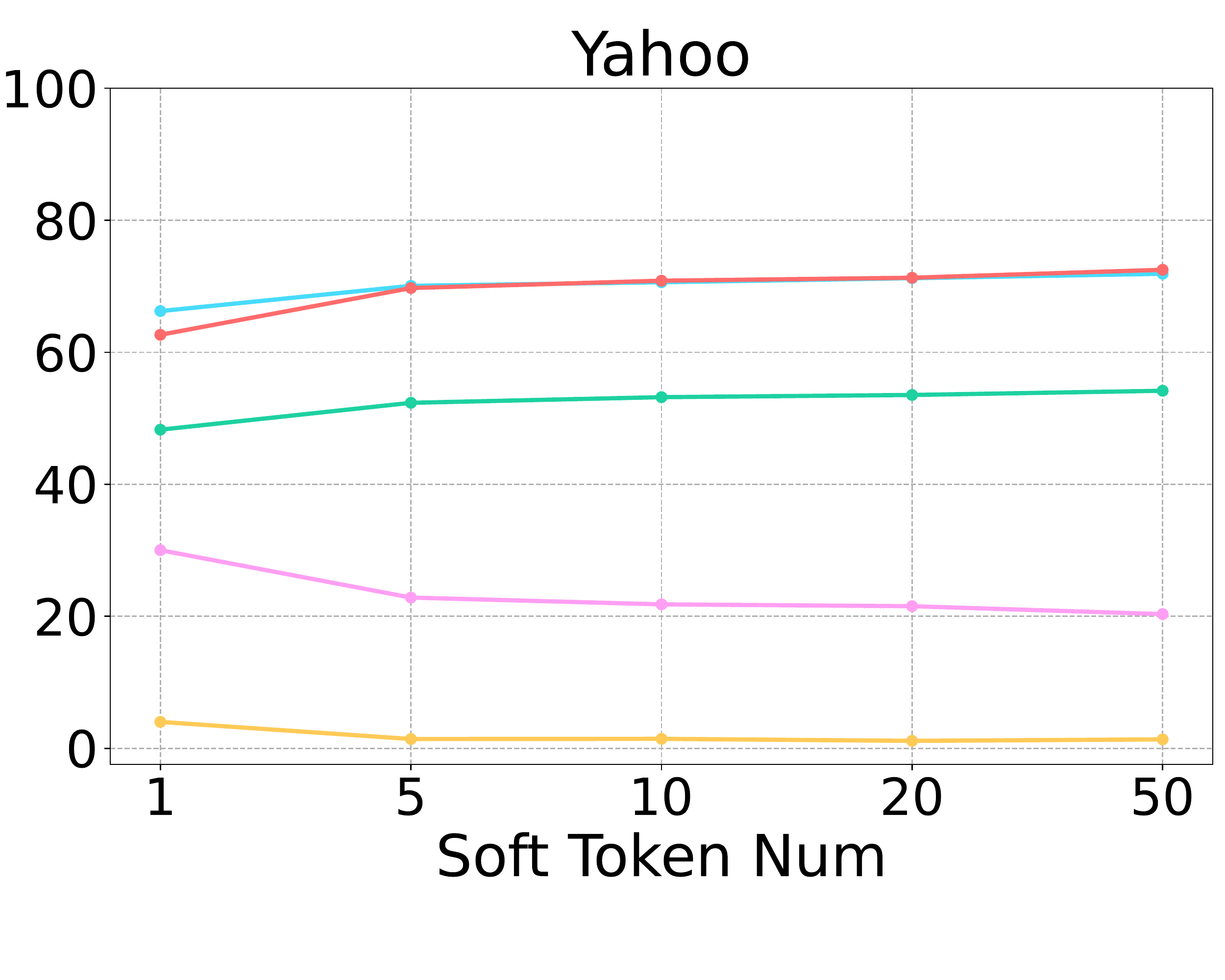}
         \label{fig:soft_prompt-t5-yahoo}
     \end{subfigure}
    \begin{subfigure}[b]{0.32\textwidth}
         \centering
         \includegraphics[trim=0 0 0 10, clip,width=\textwidth]{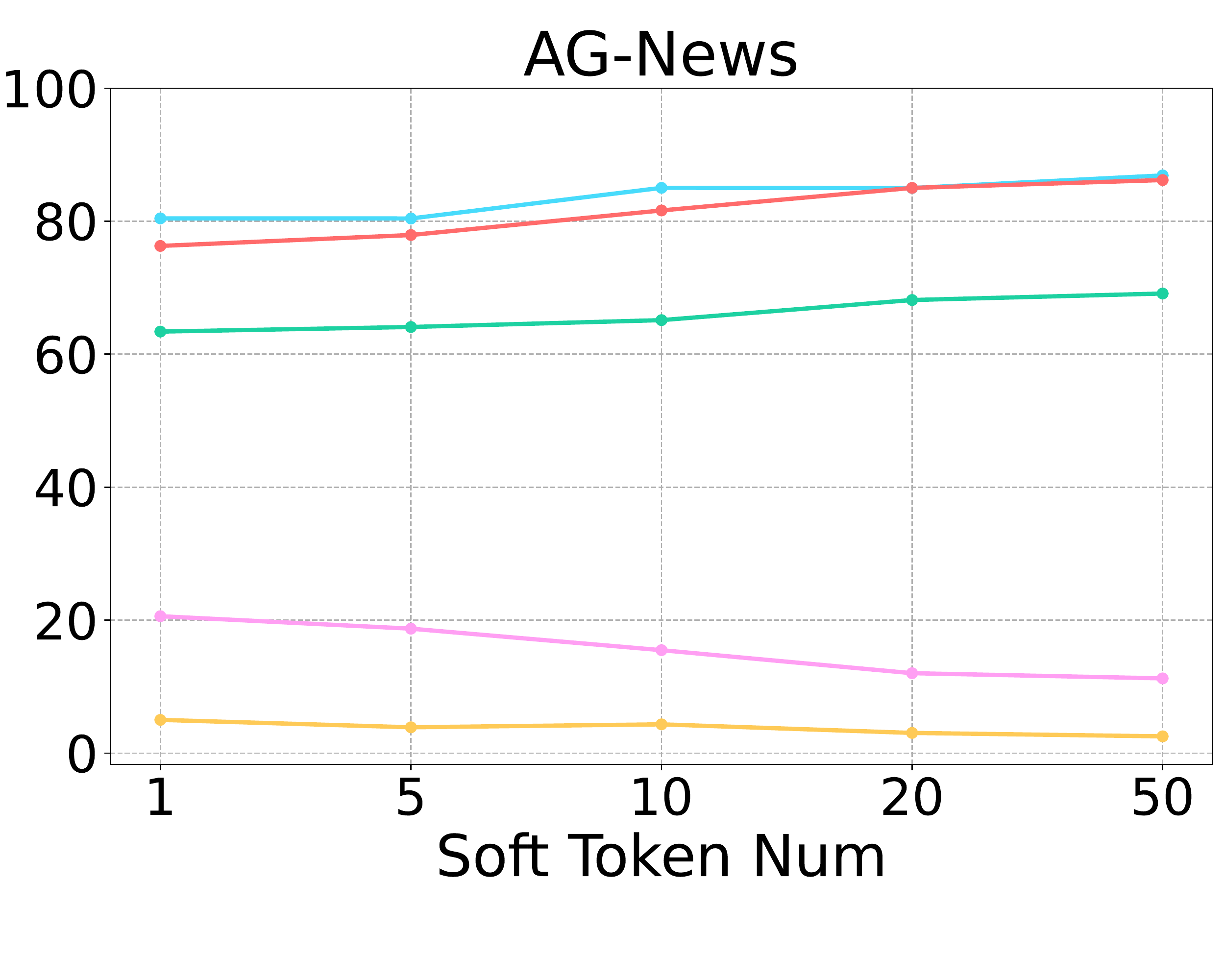}
         \label{fig:soft_prompt-t5-agnews}
     \end{subfigure}

     \vspace{-7pt}
     \begin{subfigure}[b]{0.6\textwidth}
         \centering
         \includegraphics[trim=0 0 0 10, clip, width=\textwidth]{figures/legend.pdf}
     \end{subfigure}
        
    \vspace{-5pt}
    \caption{Results of tunable parameters with T5 (Soft-prompt).}
    \label{fig:soft_prompt-t5}
    \vspace{-5pt}
\end{figure*}

\begin{figure*}[t]
     \centering
     \begin{subfigure}[b]{0.33\textwidth}
         \centering
         \includegraphics[width=\textwidth]{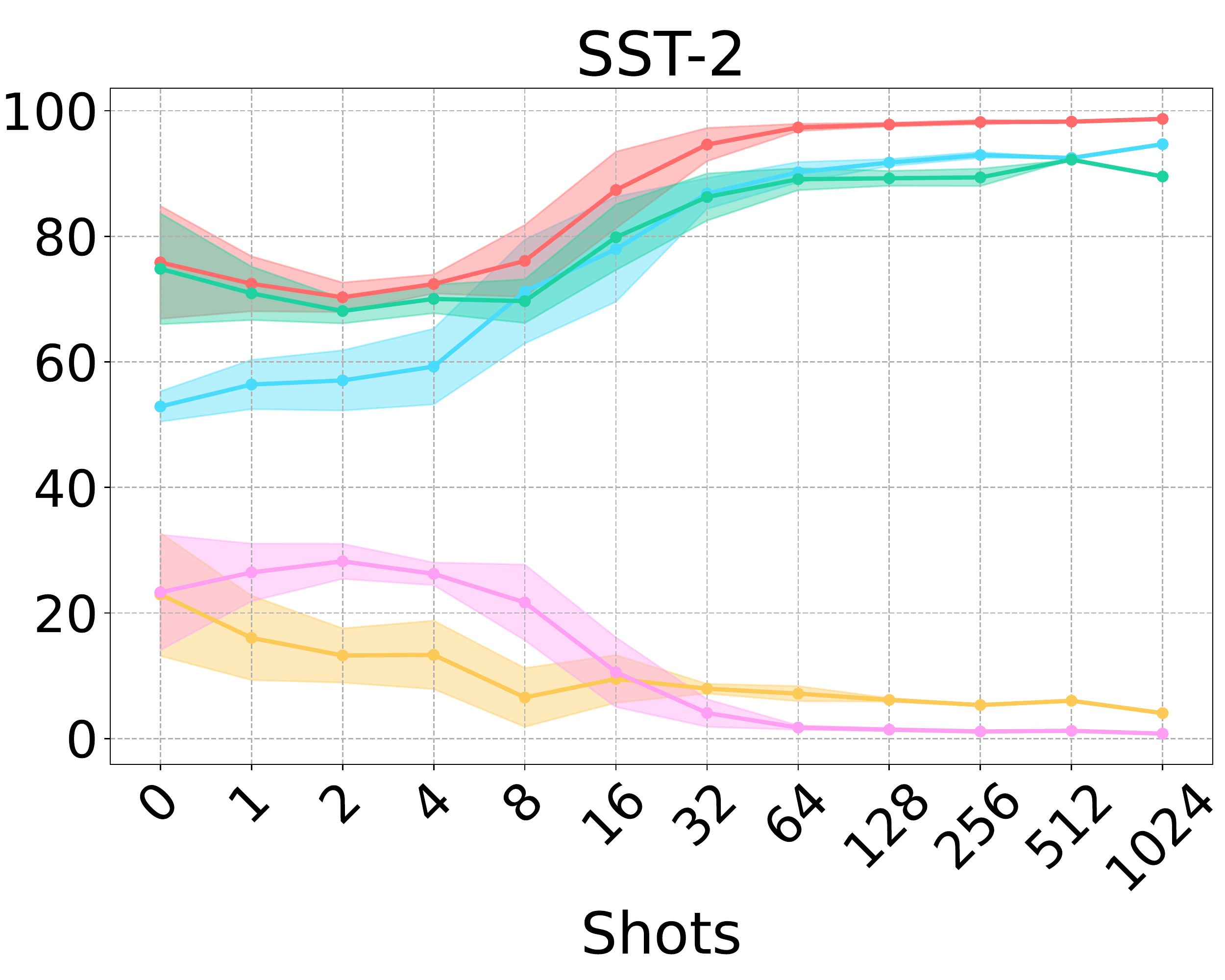}
         \label{fig:kshots-roberta-sst2}
     \end{subfigure}
     \begin{subfigure}[b]{0.33\textwidth}
         \centering
         \includegraphics[width=\textwidth]{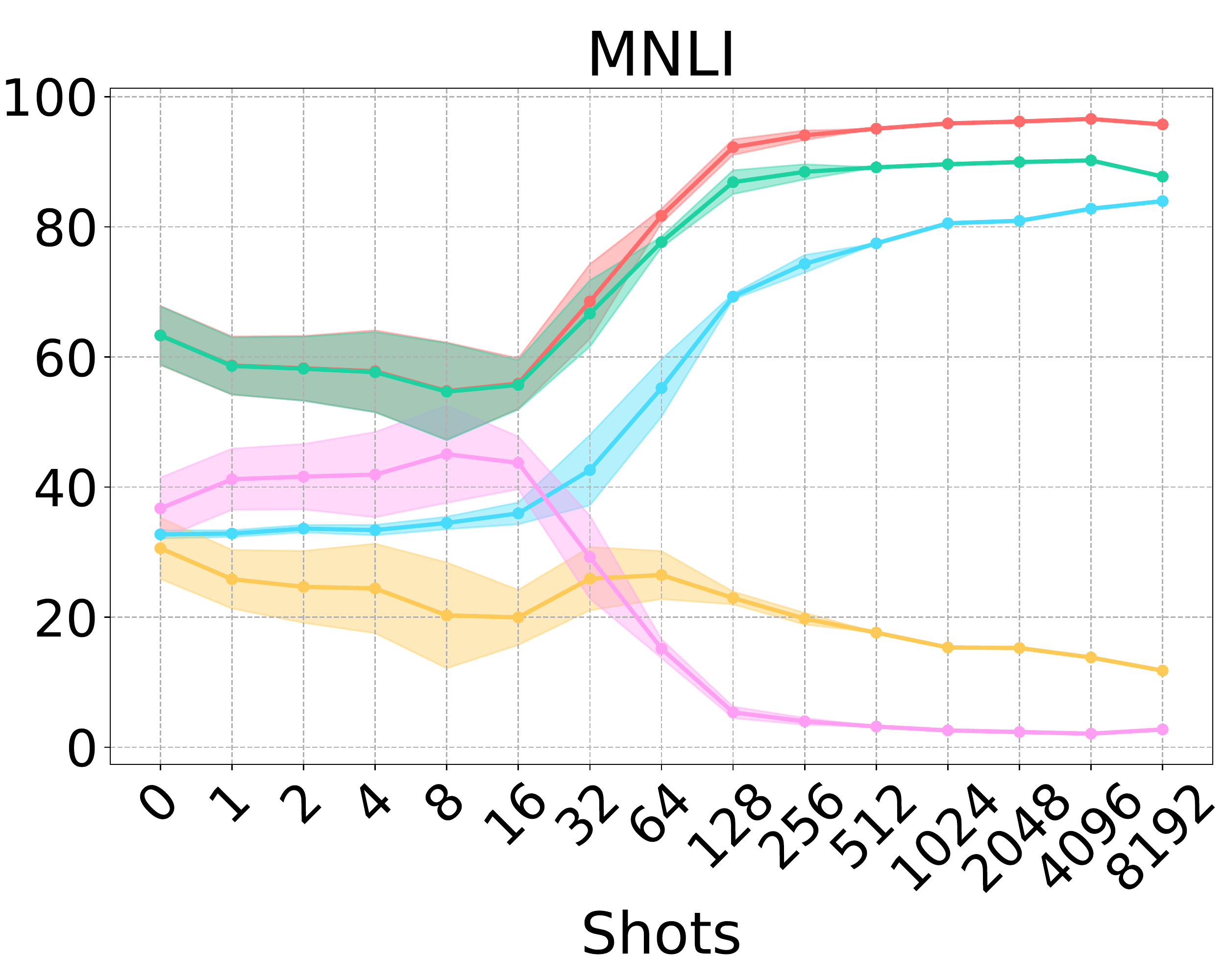}
         \label{fig:kshots-roberta-mnli}
     \end{subfigure}
    
    \vspace{-5pt}
    \centering
     \begin{subfigure}[b]{0.33\textwidth}
         \centering
         \includegraphics[width=\textwidth]{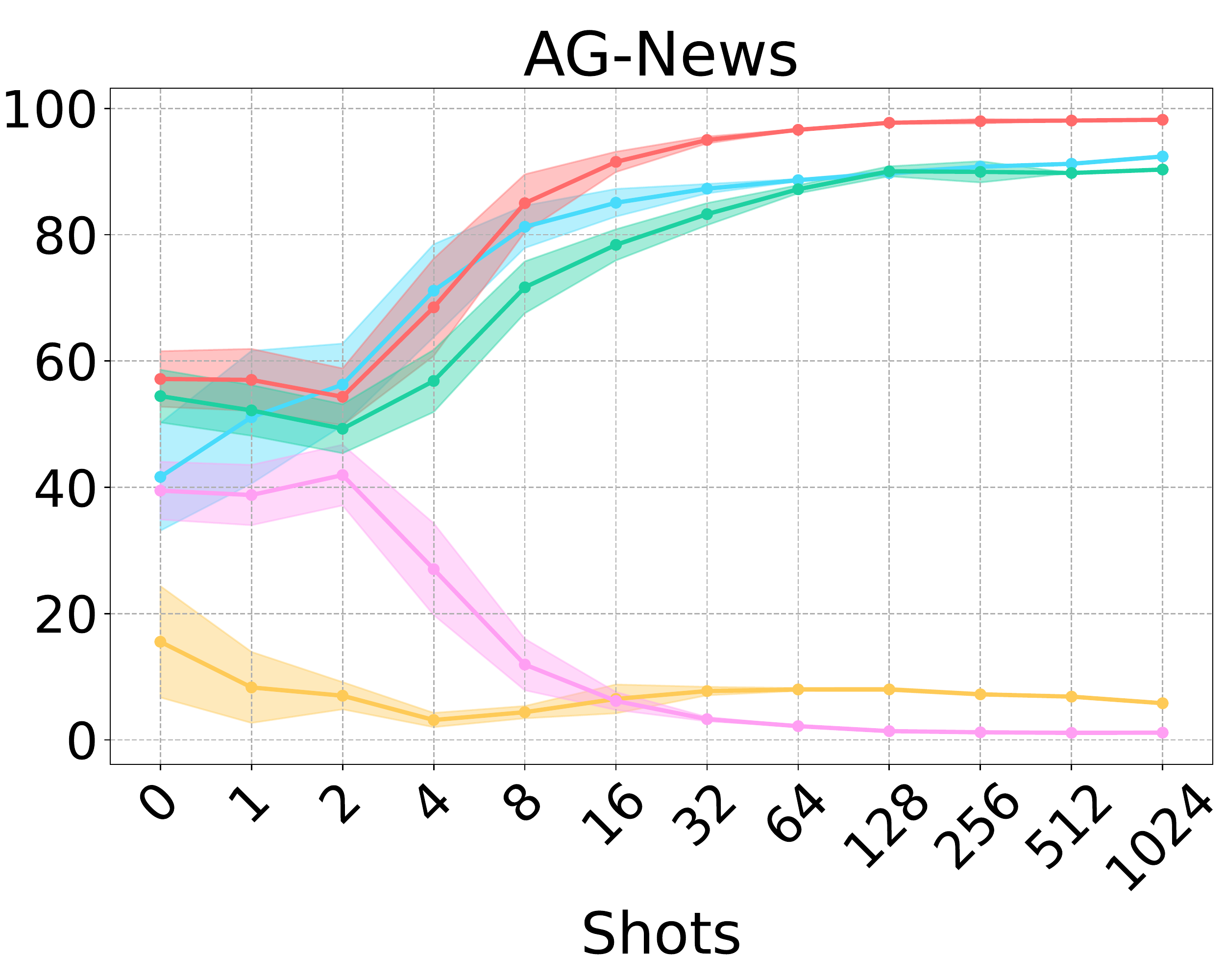}
         \label{fig:kshots-roberta-agnews}
     \end{subfigure}
     \begin{subfigure}[b]{0.33\textwidth}
         \centering
         \includegraphics[trim=0 0 0 0, clip, width=\textwidth]{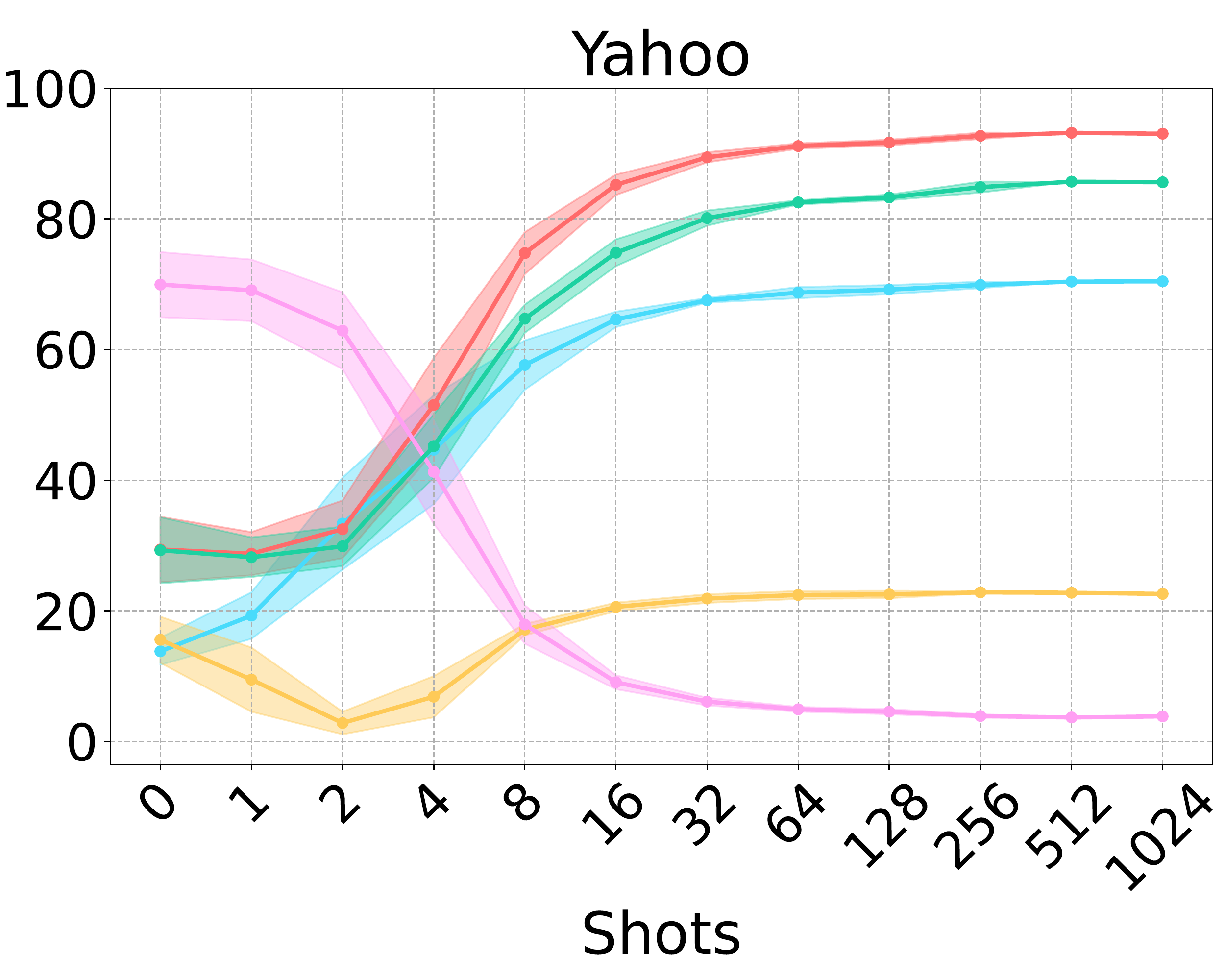}
         \label{fig:kshots-roberta-yahoo}
     \end{subfigure}

     \vspace{-15pt}
     \begin{subfigure}[b]{0.6\textwidth}
         \centering
         \includegraphics[width=\textwidth]{figures/legend.pdf}
     \end{subfigure}
     
    \vspace{-5pt}
    \caption{Results of available training samples with RoBERTa.}
    \label{fig:kshots-roberta}
\end{figure*}
\begin{figure*}[!h]
     \centering
     \begin{subfigure}[b]{0.33\textwidth}
         \centering
         \includegraphics[width=\textwidth]{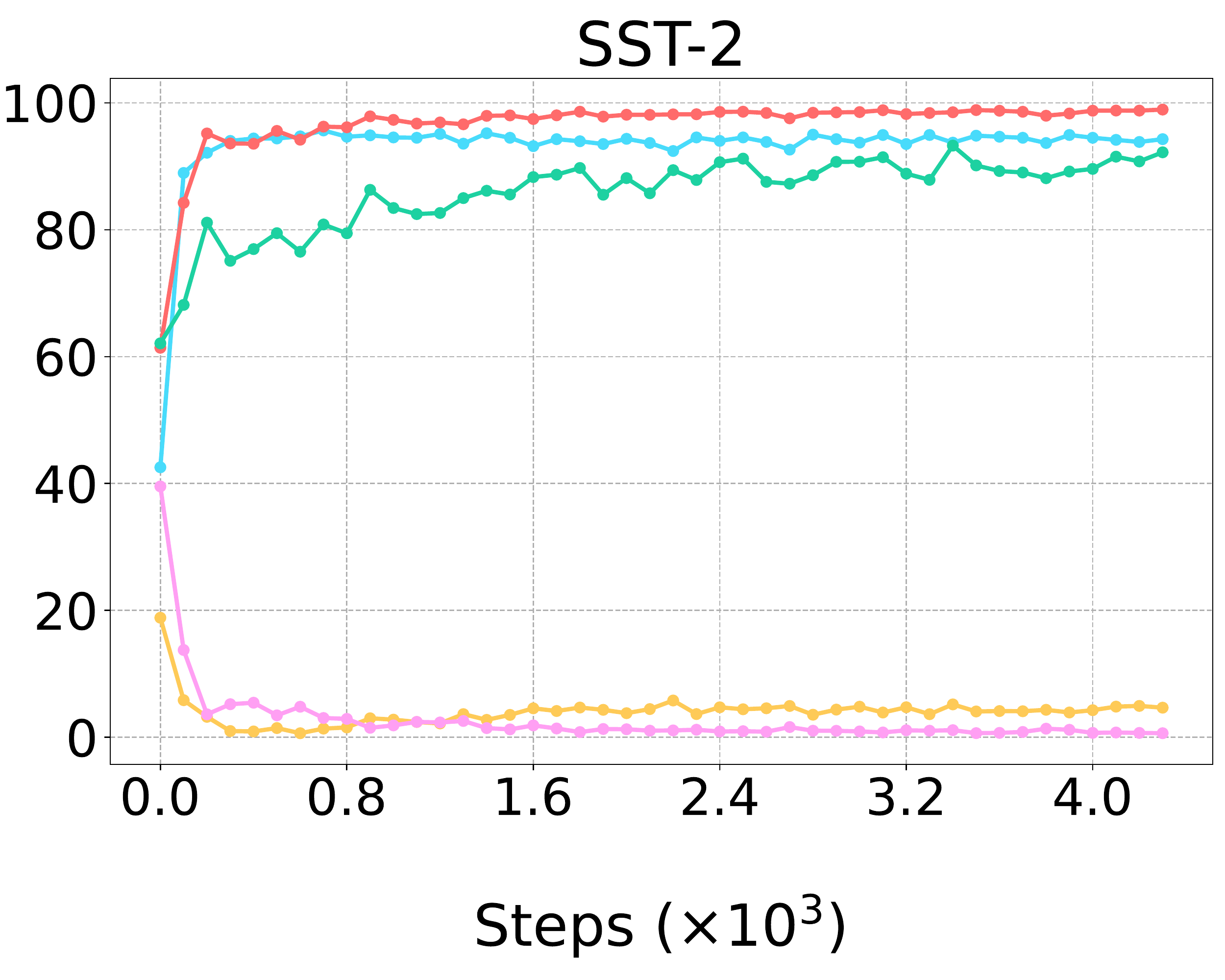}
         \label{fig:dynamics-roberta-sst2}
     \end{subfigure}
     \begin{subfigure}[b]{0.33\textwidth}
         \centering
         \includegraphics[width=\textwidth]{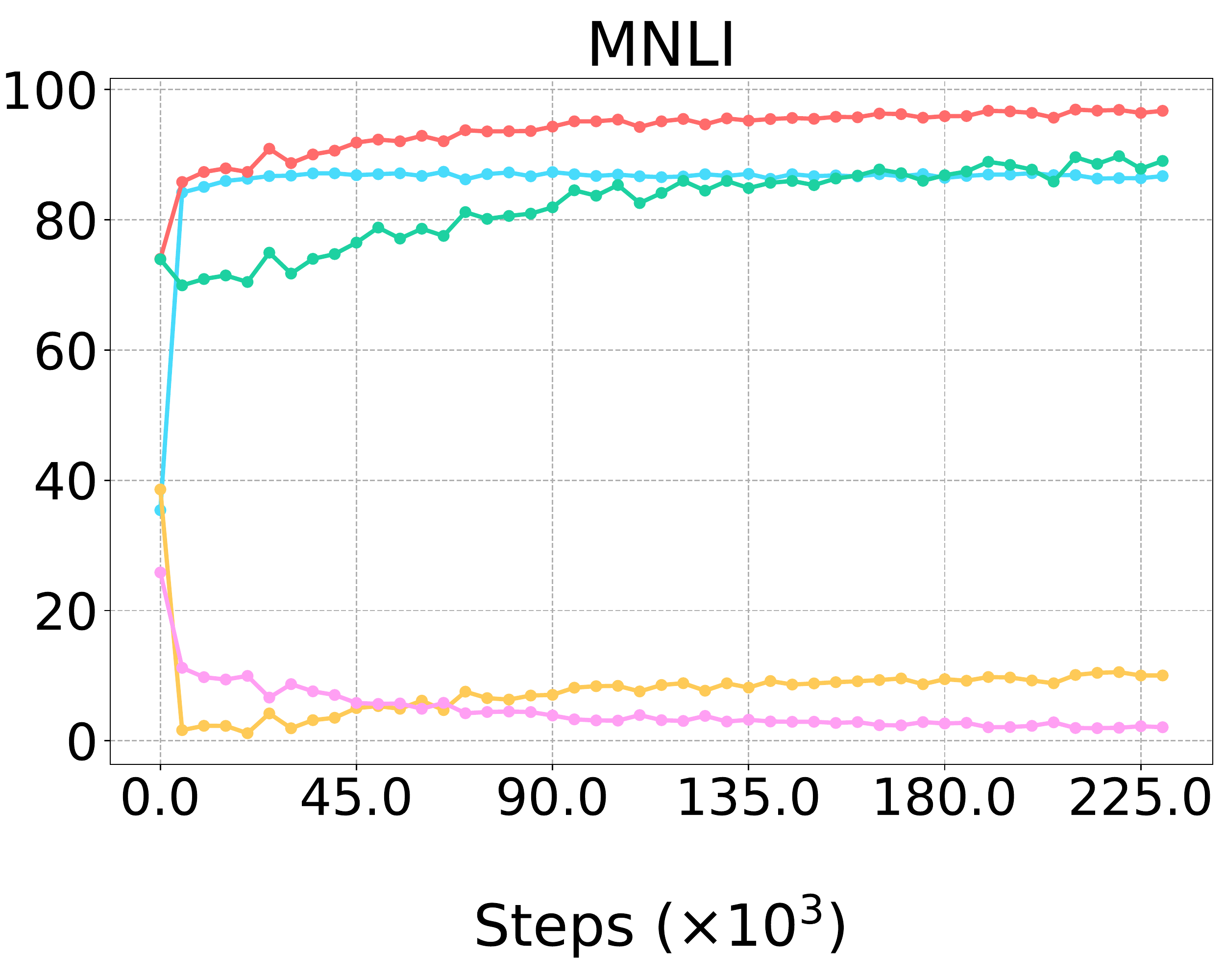}
         \label{fig:dynamics-roberta-mnli}
     \end{subfigure}
    
    \vspace{-5pt}
    \centering
     \begin{subfigure}[b]{0.33\textwidth}
         \centering
         \includegraphics[width=\textwidth]{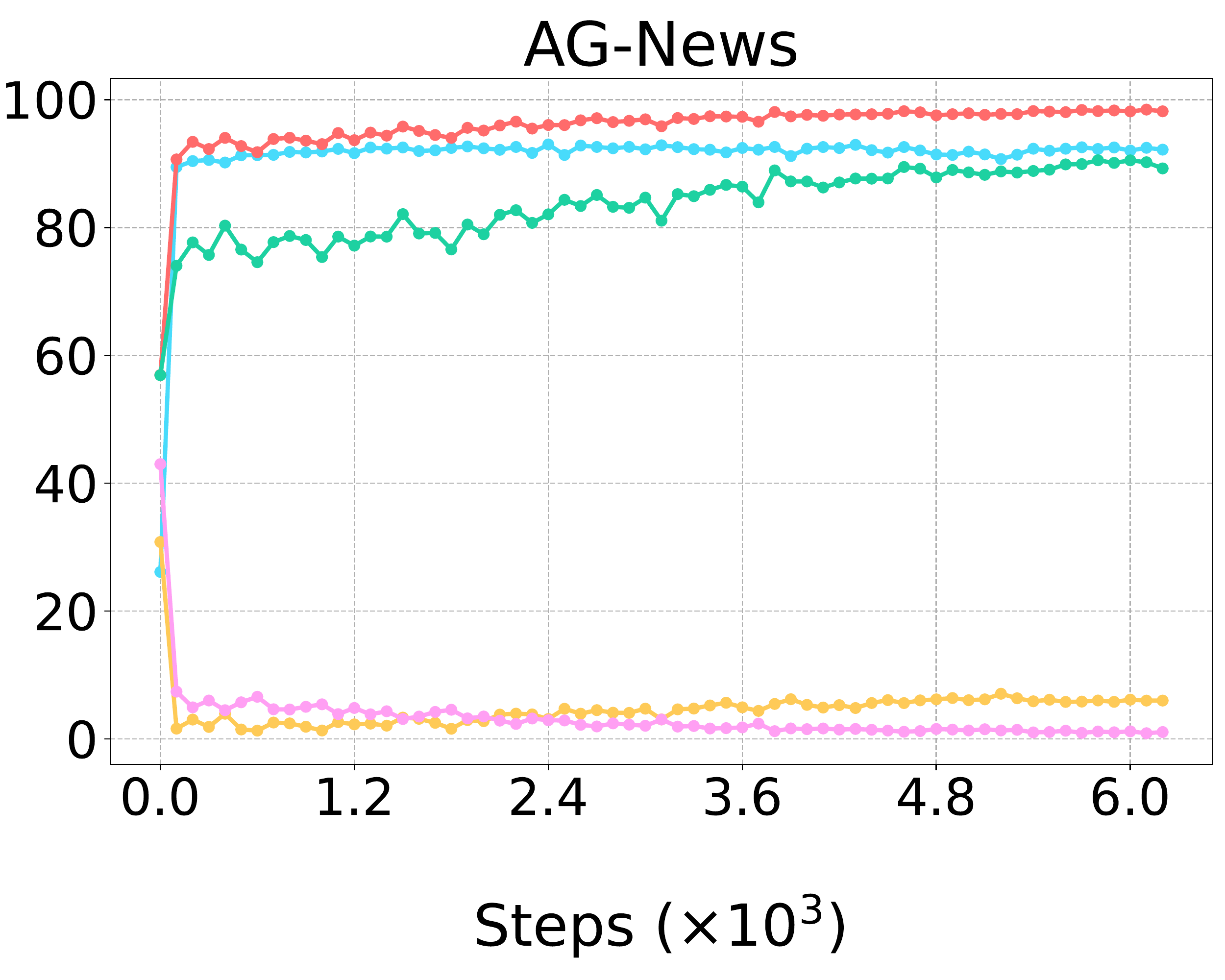}
         \label{fig:dynamics-roberta-agnews}
     \end{subfigure}
     \begin{subfigure}[b]{0.33\textwidth}
         \centering
         \includegraphics[width=\textwidth]{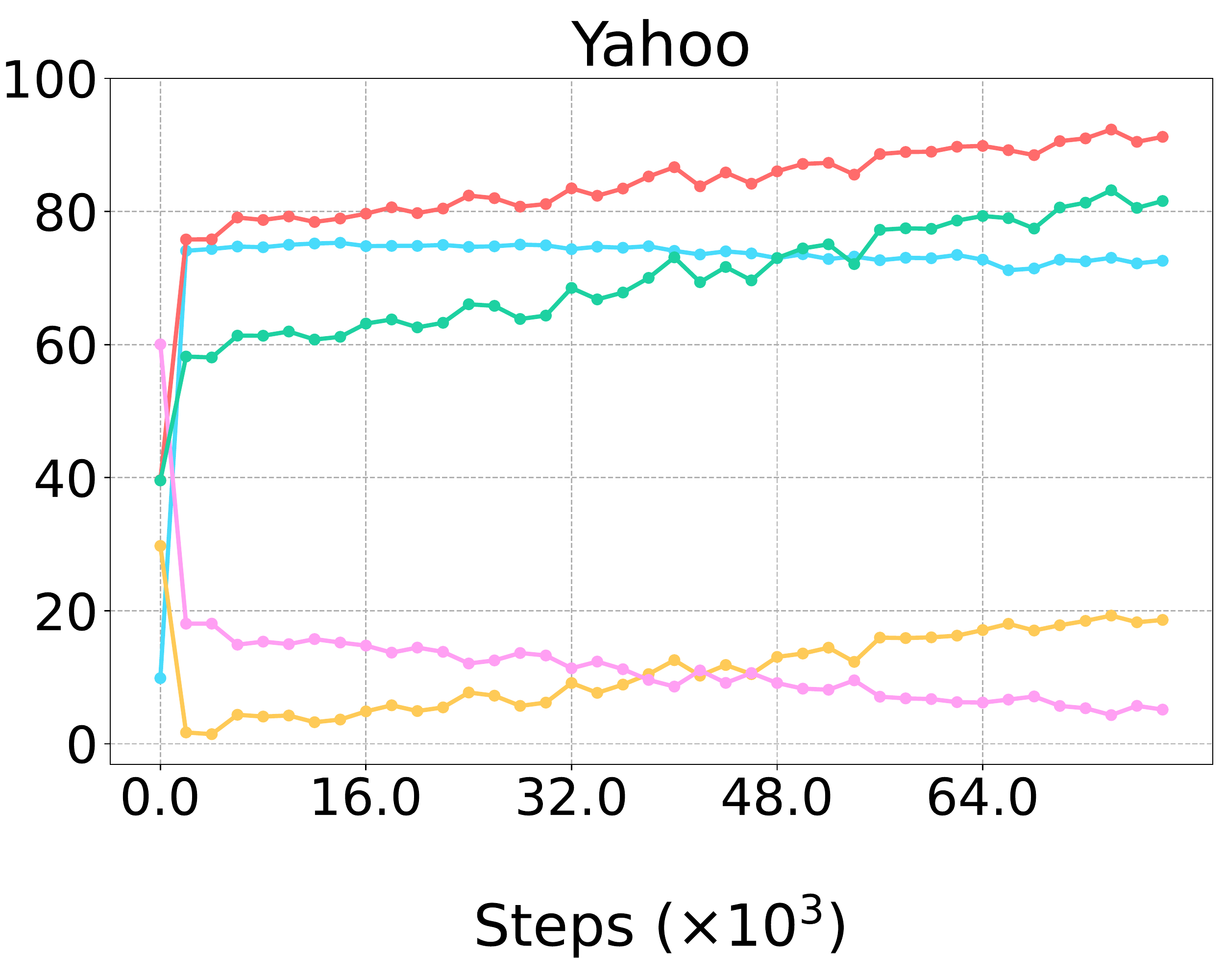}
         \label{fig:dynamics-roberta-yahoo}
     \end{subfigure}

     \vspace{-18pt}
     \begin{subfigure}[b]{0.6\textwidth}
         \centering
         \includegraphics[width=\textwidth]{figures/legend.pdf}
     \end{subfigure}
     
    \vspace{-5pt}
    \caption{Results of training steps with RoBERTa.}
    \label{fig:dynamics-roberta}
\end{figure*}

\begin{figure*}[h]
     \centering
     \begin{subfigure}[b]{0.33\textwidth}
         \centering
         \includegraphics[trim=0 0 0 0, clip, width=\textwidth]{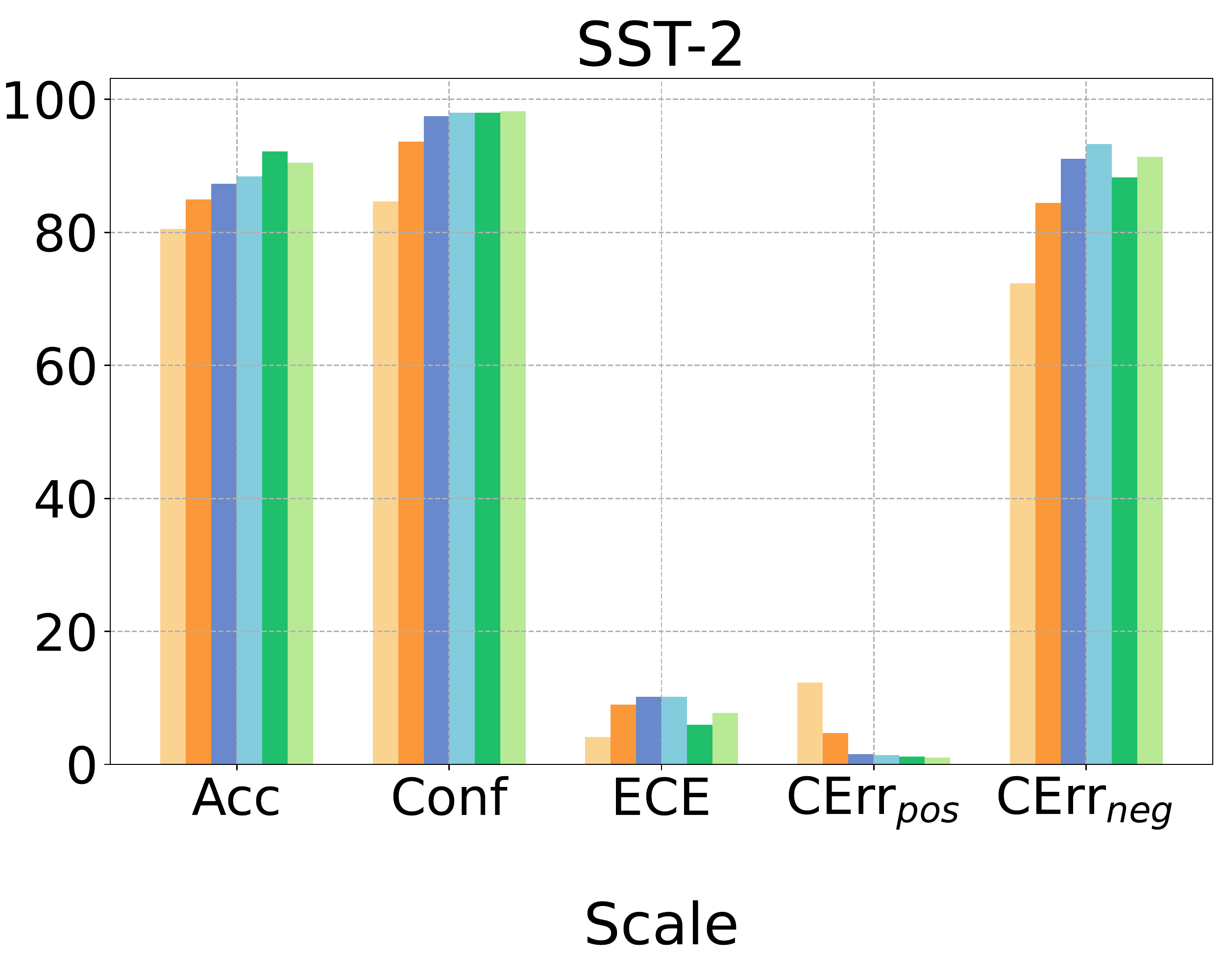}
         \label{fig:scale-bert-sst2}
     \end{subfigure}
     \begin{subfigure}[b]{0.33\textwidth}
         \centering
         \includegraphics[trim=0 0 0 0, clip, width=\textwidth]{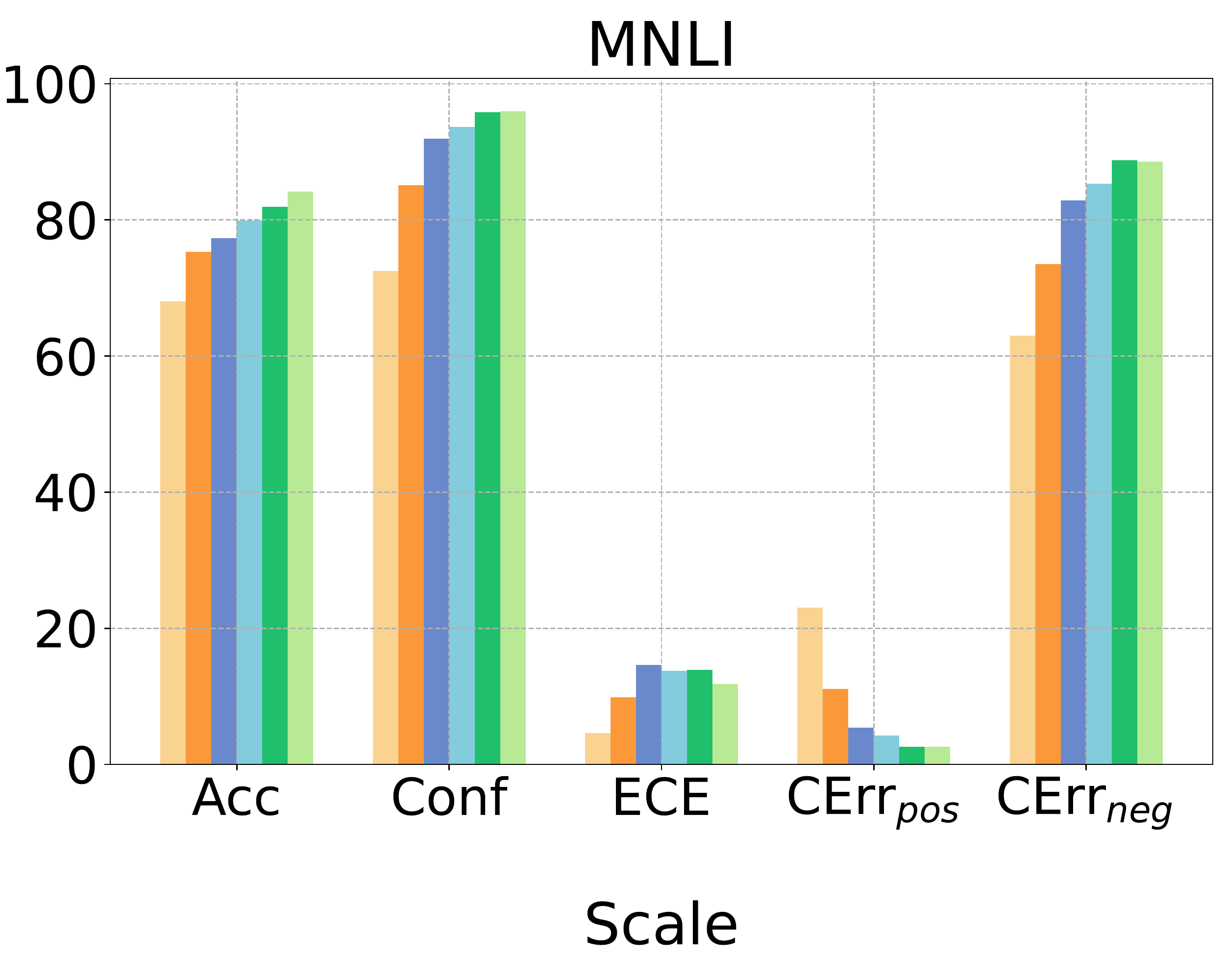}
         \label{fig:scale-bert-mnli}
     \end{subfigure}
    
    \vspace{-5pt}
    \centering
     \begin{subfigure}[b]{0.33\textwidth}
         \centering
         \includegraphics[trim=0 0 0 0, clip, width=\textwidth]{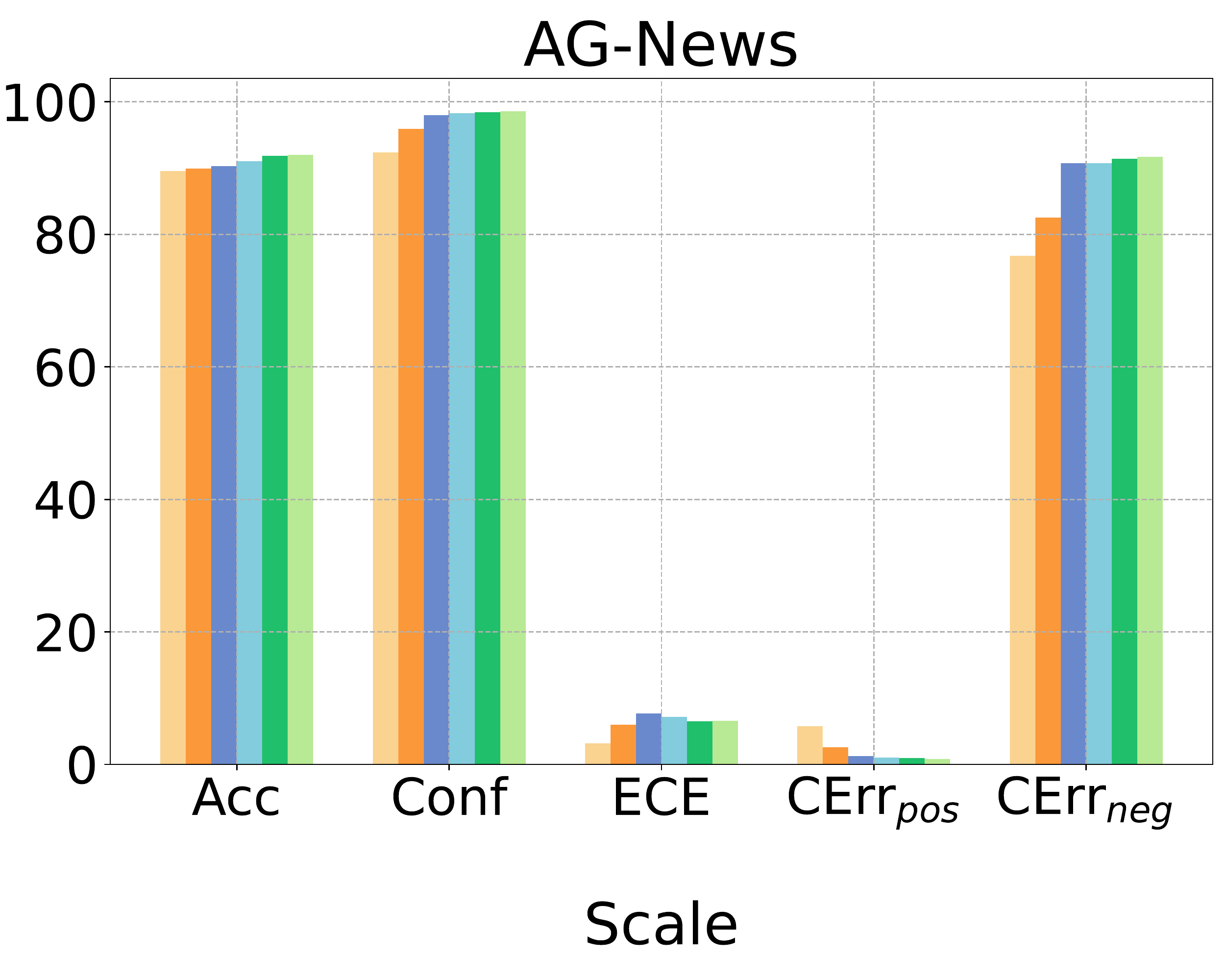}
         \label{fig:scale-bert-agnews}
     \end{subfigure}
     \begin{subfigure}[b]{0.33\textwidth}
         \centering
     \includegraphics[trim=0 0 0 0, clip, width=\textwidth]{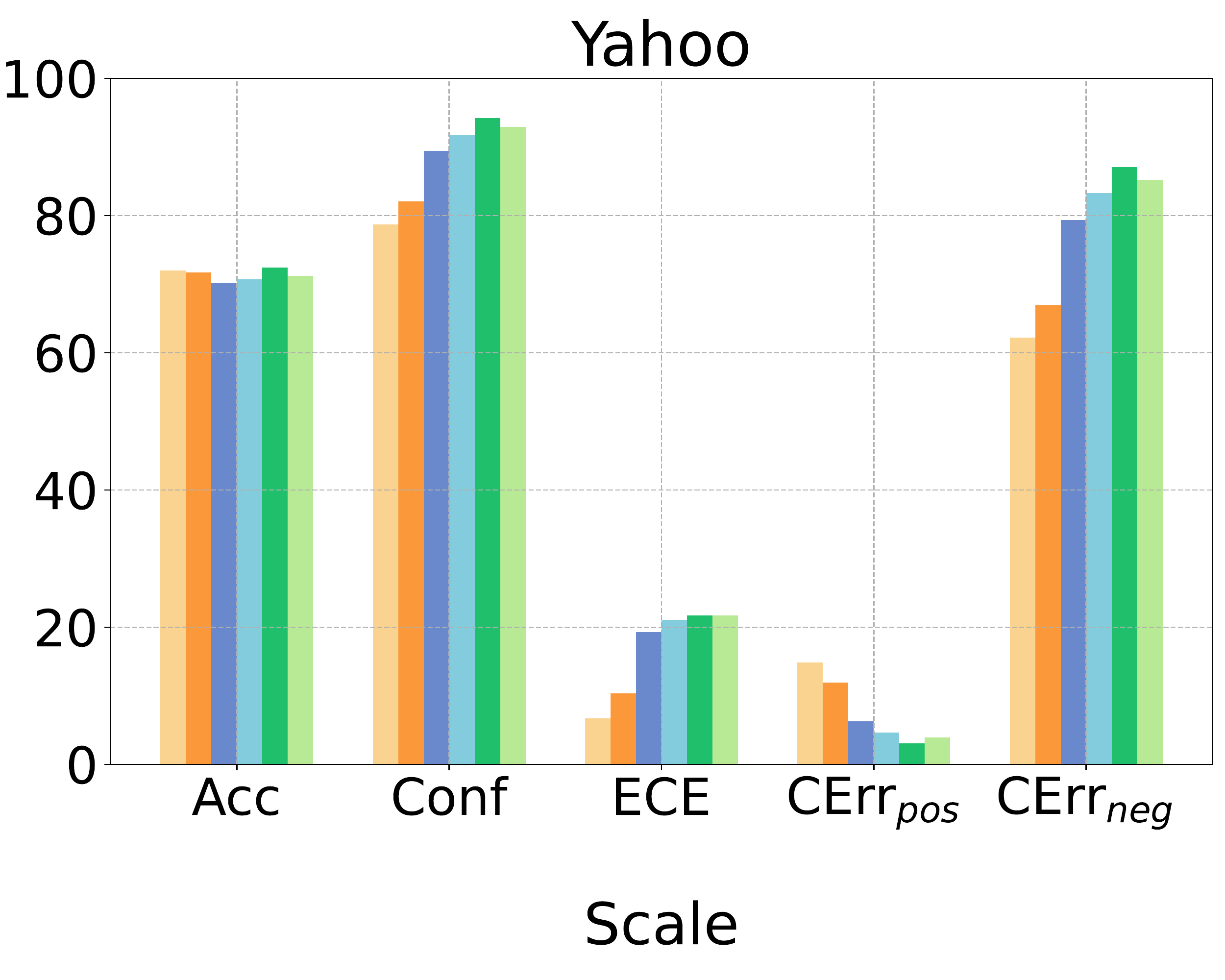}
     \label{fig:scale-bert-yahoo}
     \end{subfigure}

    \vspace{-15pt}
     \begin{subfigure}[b]{0.64\textwidth}
         \centering
     \includegraphics[width=\textwidth]{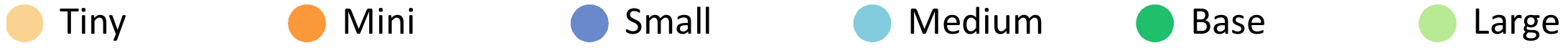}
     \end{subfigure}
     
    \vspace{-5pt}
    \caption{Results of increasing PLMs scales with RoBERTa.}
    \label{fig:scale-bert}
\end{figure*}

\begin{figure*}[!h]
     \centering
     \begin{subfigure}[b]{0.33\textwidth}
         \centering
         \includegraphics[trim=10 20 0 0, clip, width=\textwidth]{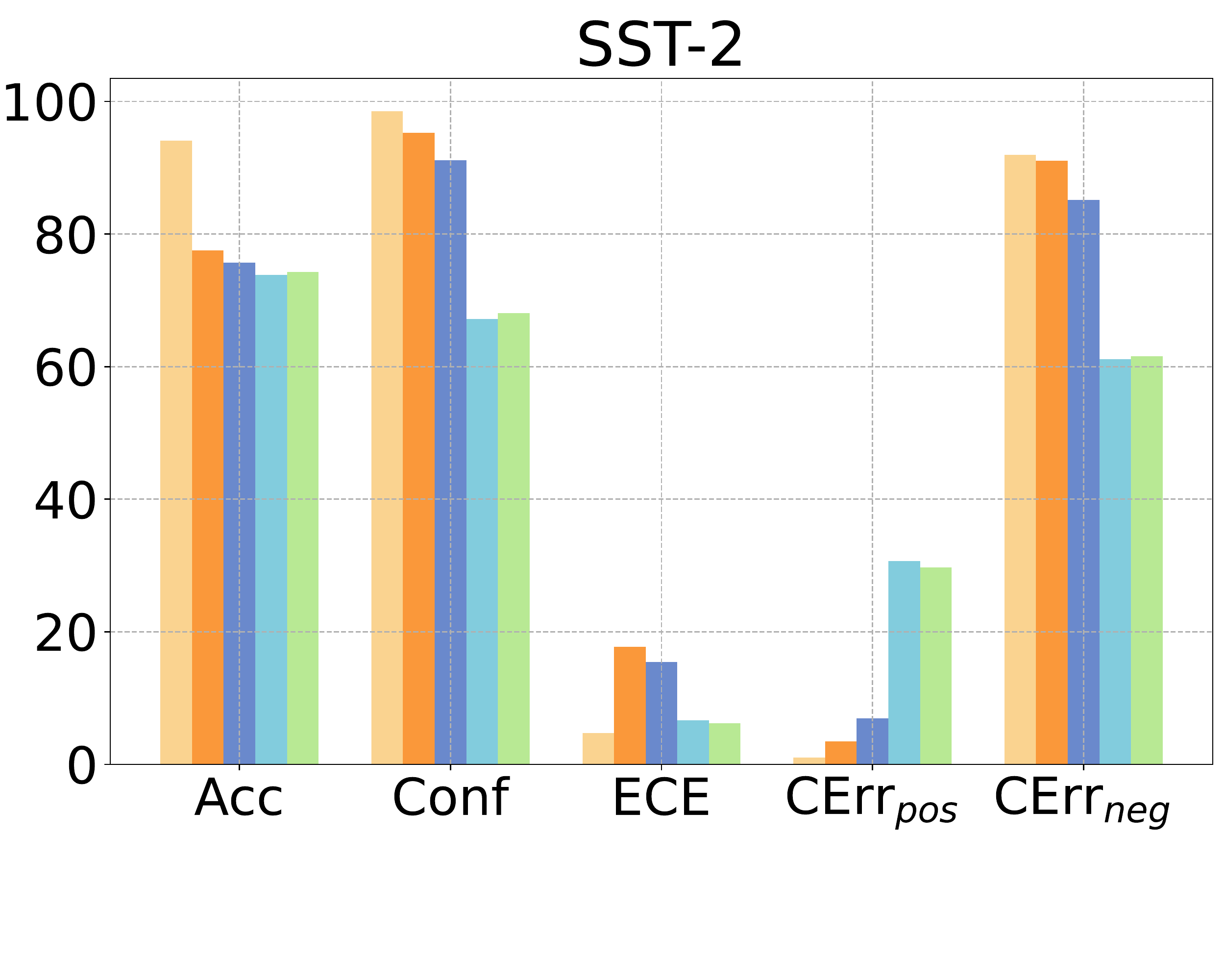}
         \label{fig:pretrain-roberta-sst2}
     \end{subfigure}
     \begin{subfigure}[b]{0.33\textwidth}
         \centering
         \includegraphics[trim=10 20 0 0, clip, width=\textwidth]{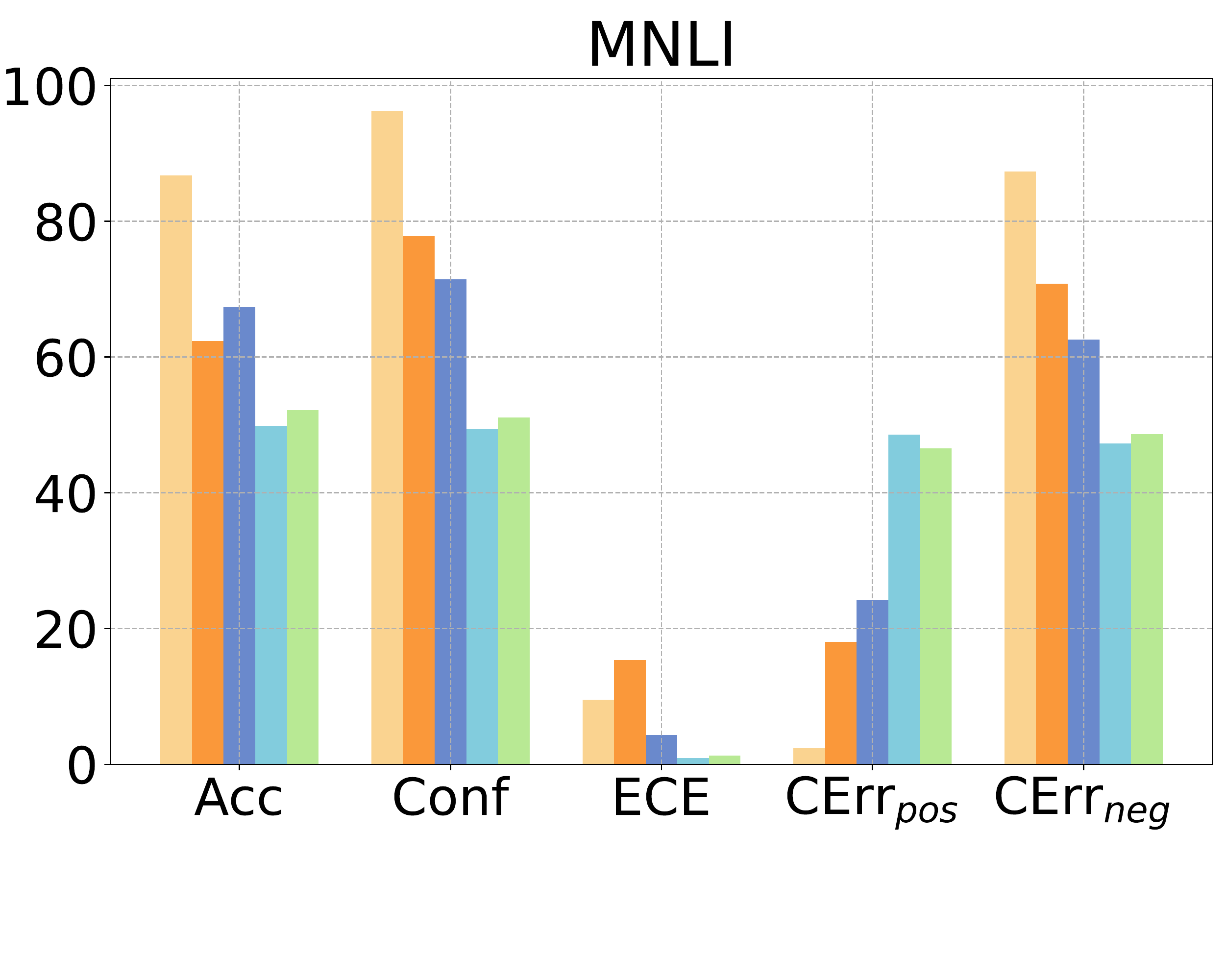}
         \label{fig:pretrain-roberta-mnli}
     \end{subfigure}

     \vspace{-10pt}
     \begin{subfigure}[b]{0.33\textwidth}
         \centering
         \includegraphics[trim=10 20 0 0, clip, width=\textwidth]{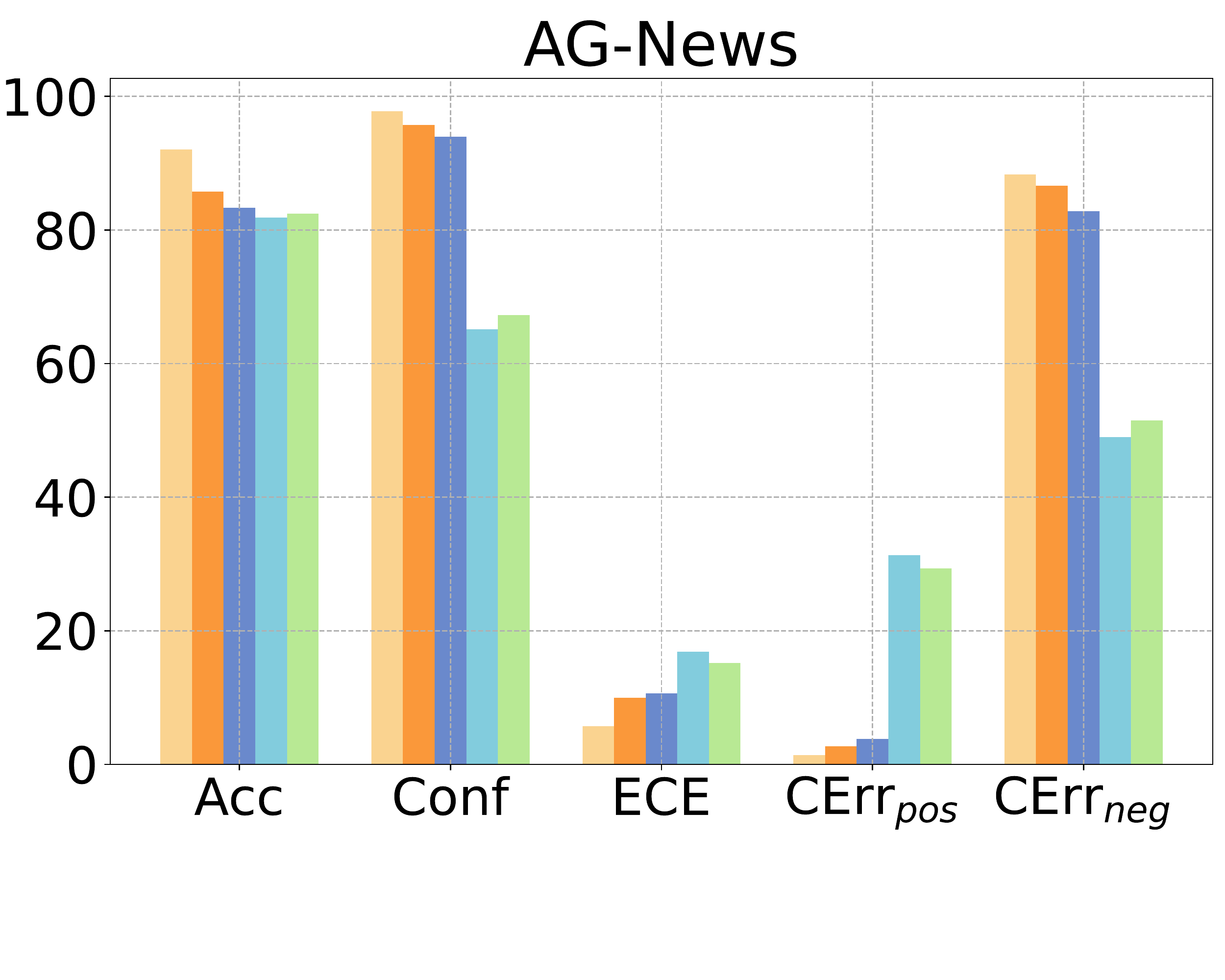}
         \label{fig:pretrain-roberta-agnews}
     \end{subfigure}
     \begin{subfigure}[b]{0.33\textwidth}
         \centering
         \includegraphics[trim=10 20 0 0, clip, width=\textwidth]{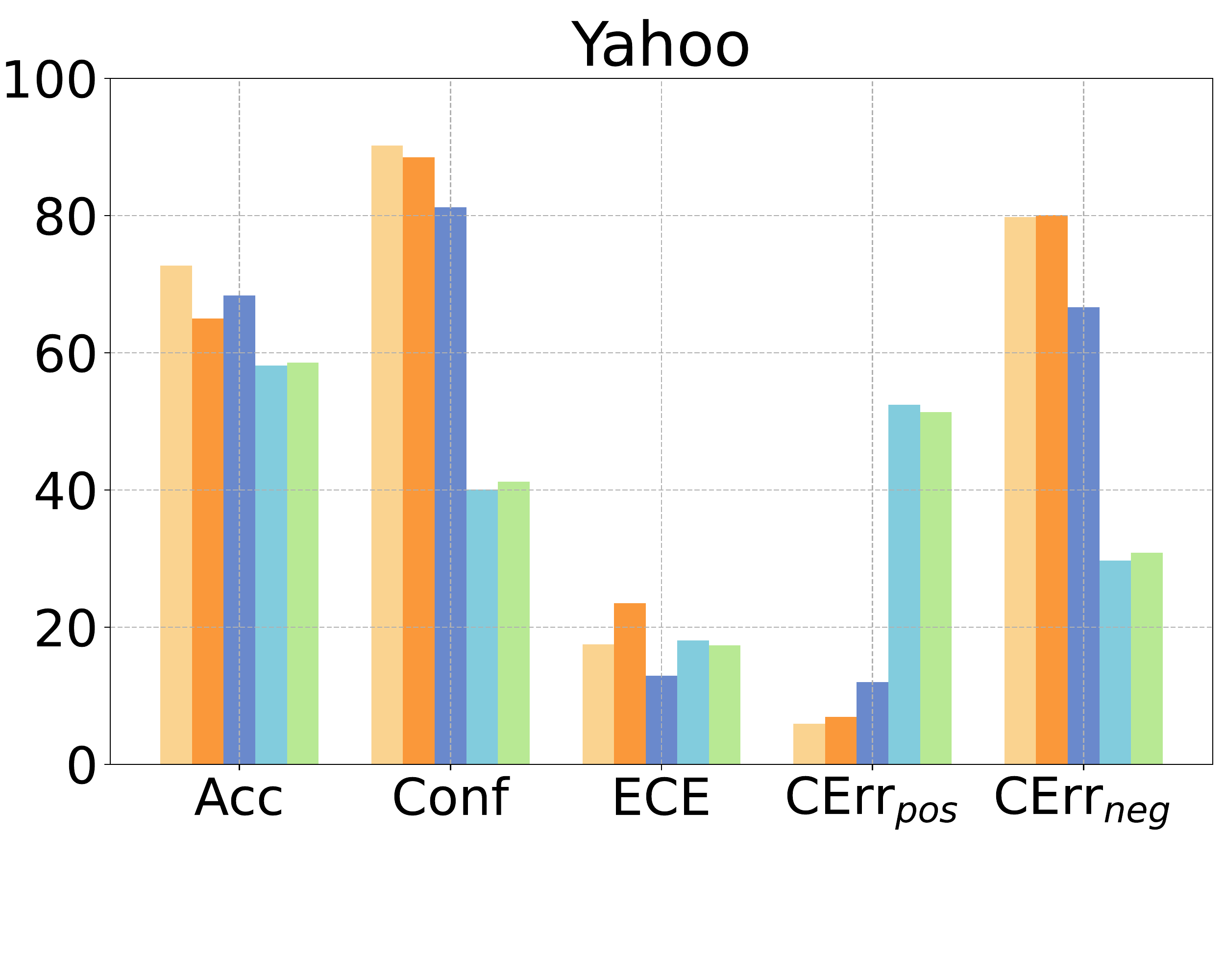}
         \label{fig:pretrain-roberta-yahoo}
     \end{subfigure}
     
    \vspace{-17pt}
     \begin{subfigure}[b]{0.64\textwidth}
         \centering
         \includegraphics[width=\textwidth]{ figures/pretrain_legend.pdf}
     \end{subfigure}
        \vspace{-5pt}
        
        \caption{Results of the pretraining influence with RoBERTa.}
        \label{fig:pretrain_roberta}
\end{figure*}

\begin{figure*}[t]
     \centering
     \begin{subfigure}[b]{0.33\textwidth}
         \centering
         \includegraphics[width=\textwidth]{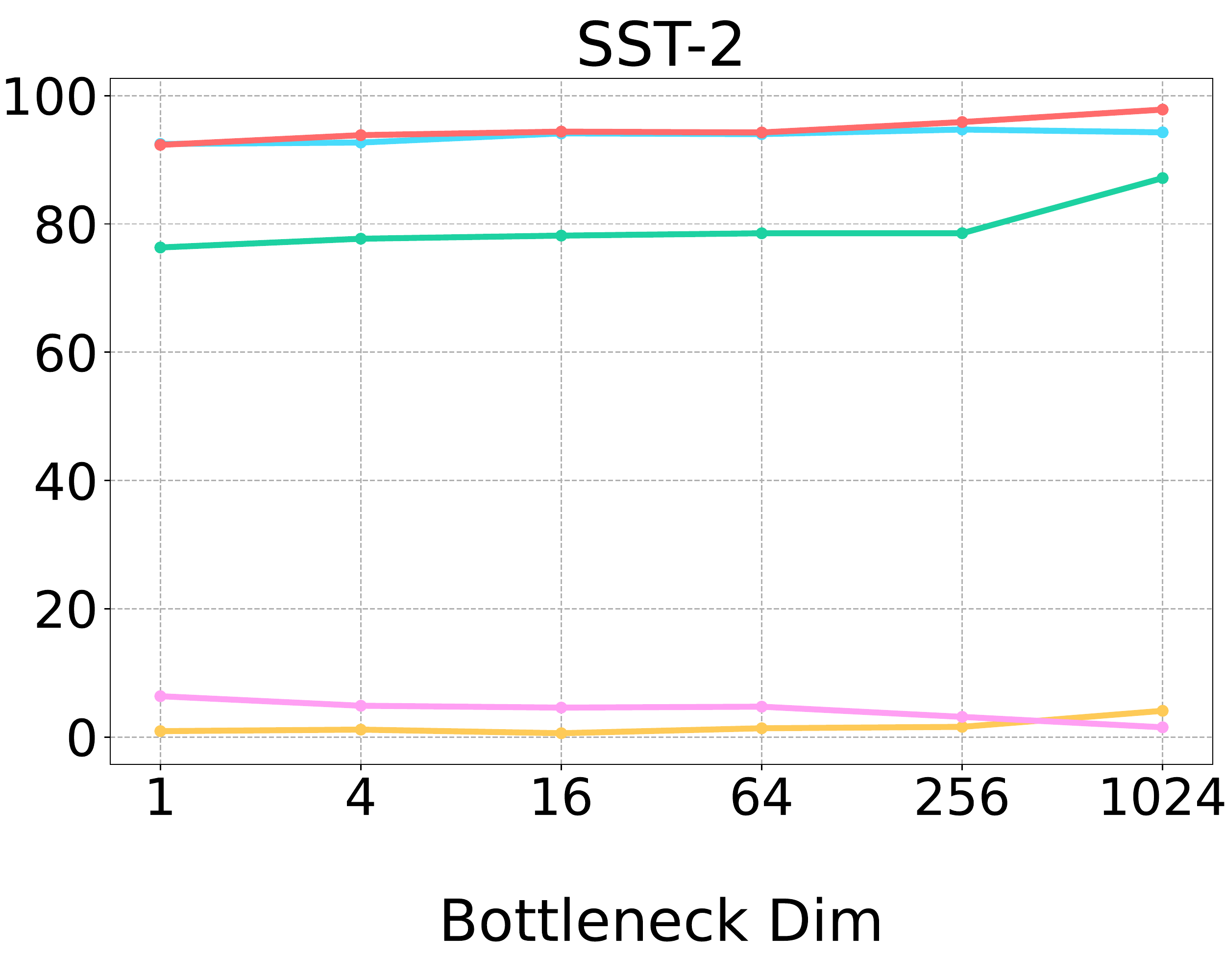}
         \label{fig:parameter-roberta-sst2}
     \end{subfigure}
     \begin{subfigure}[b]{0.33\textwidth}
         \centering
         \includegraphics[width=\textwidth]{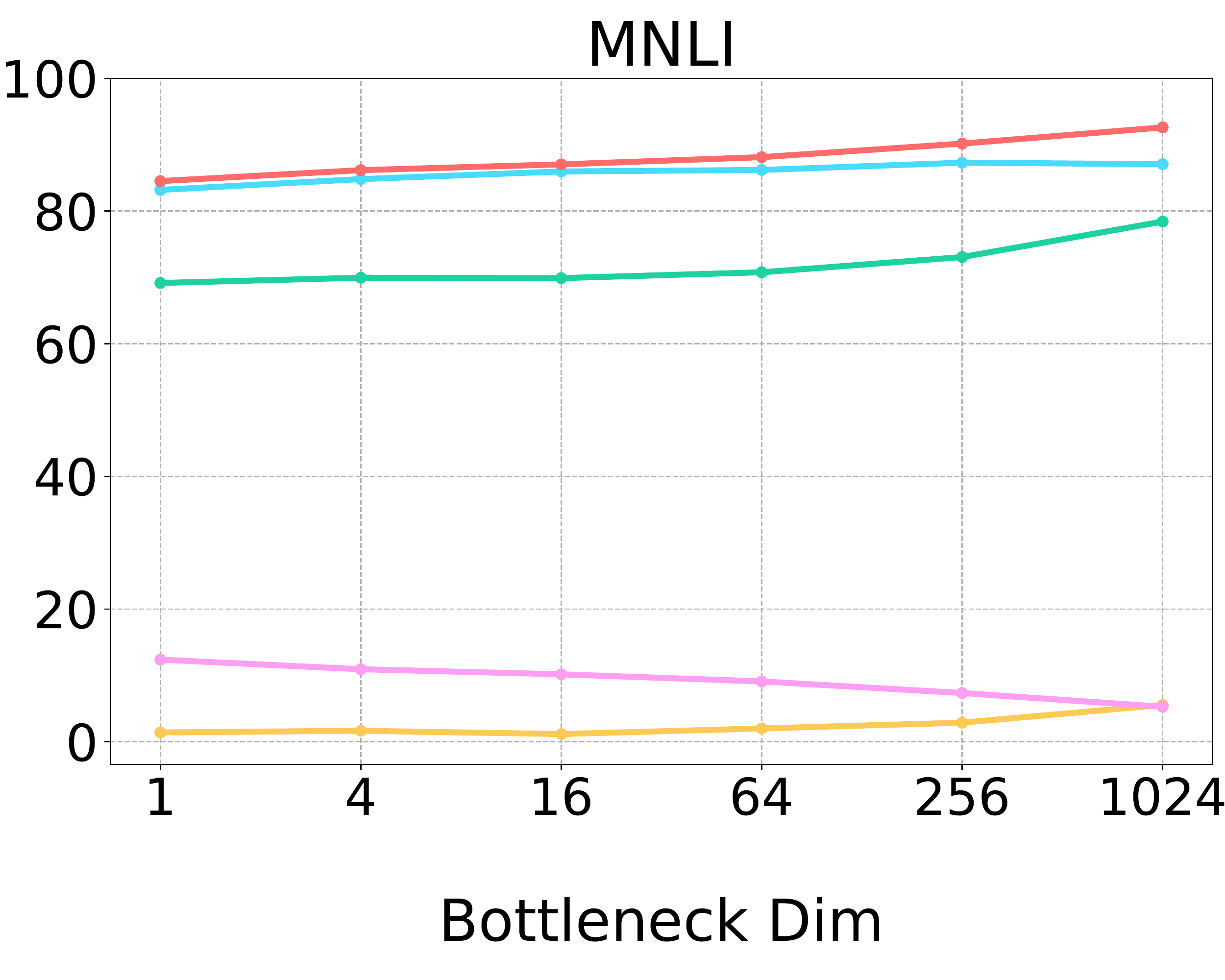}
         \label{fig:parameter-roberta-mnli}
     \end{subfigure}
    
    \vspace{-5pt}
    \centering
     \begin{subfigure}[b]{0.33\textwidth}
         \centering
         \includegraphics[width=\textwidth]{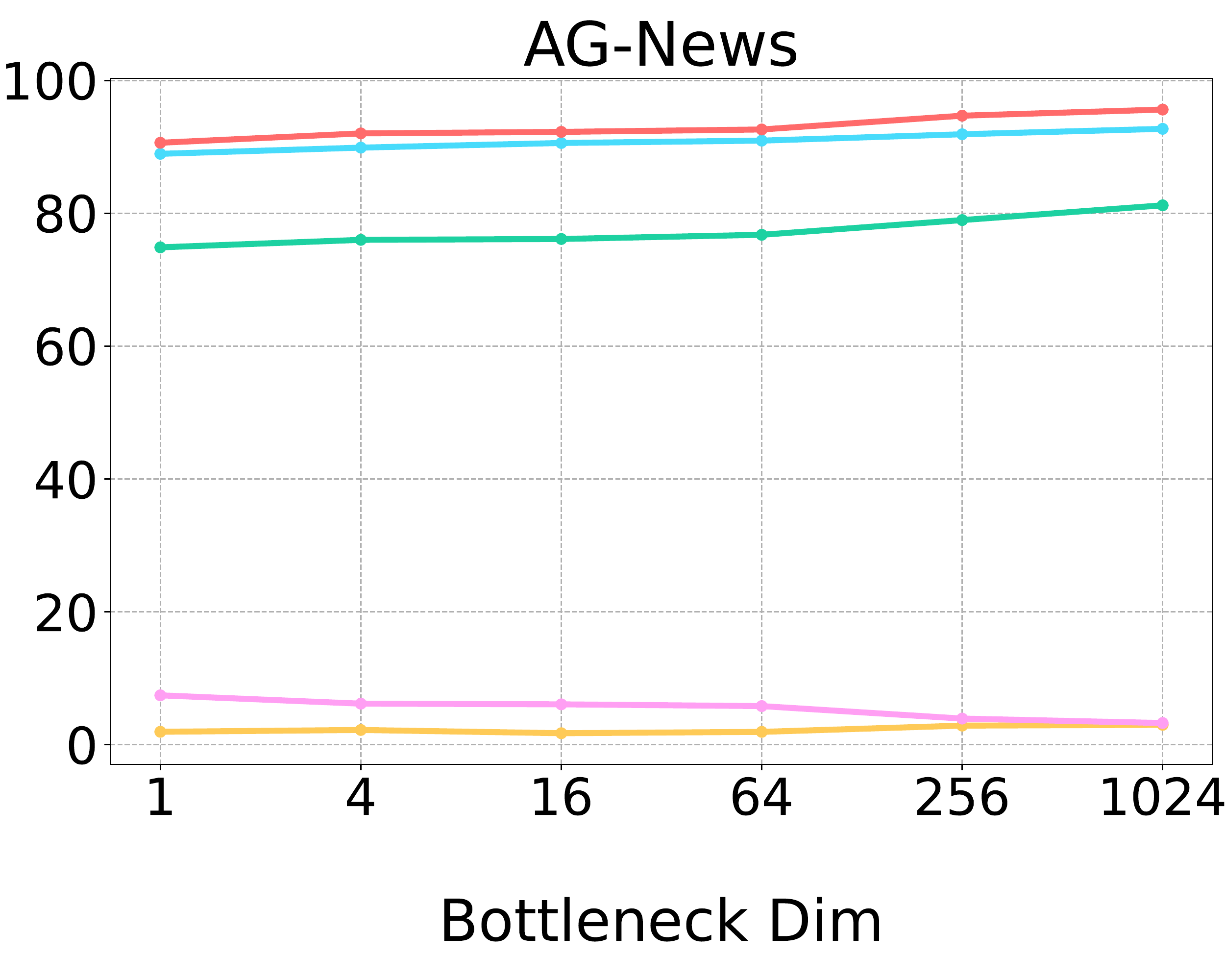}
         \label{fig:parameter-roberta-agnews}
     \end{subfigure}
     \begin{subfigure}[b]{0.33\textwidth}
         \centering
         \includegraphics[width=\textwidth]{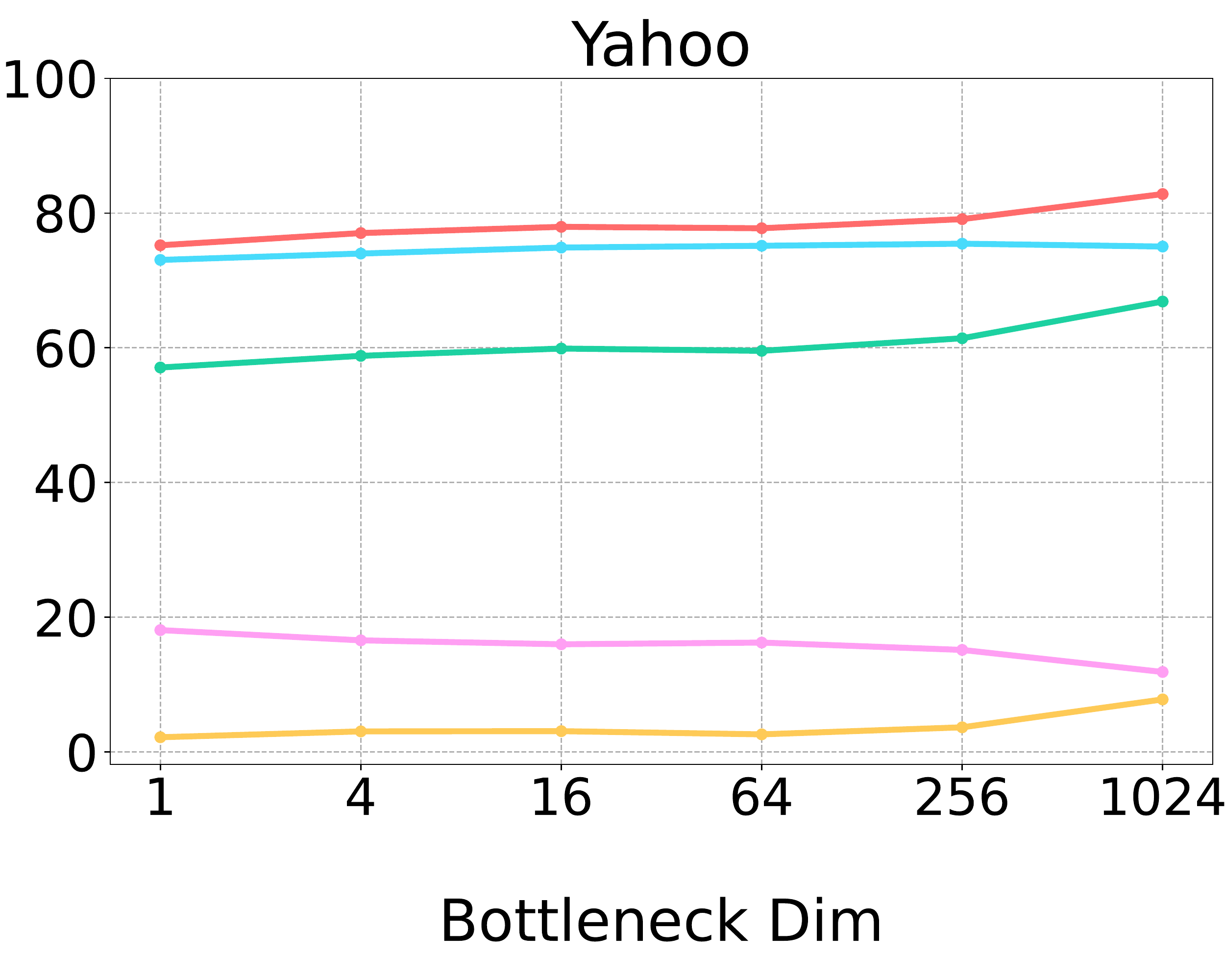}
         \label{fig:parameter-roberta-yahoo}
     \end{subfigure}

     \vspace{-18pt}
     \begin{subfigure}[b]{0.6\textwidth}
         \centering
         \includegraphics[width=\textwidth]{figures/legend.pdf}
     \end{subfigure}
     
    \vspace{-5pt}
    \caption{Results of tunable parameters with RoBERTa (Adapter).}
    \label{fig:parameter-roberta}
\end{figure*}

\begin{figure*}[!h]
     \centering
     \begin{subfigure}[b]{0.33\textwidth}
         \centering
         \includegraphics[width=\textwidth]{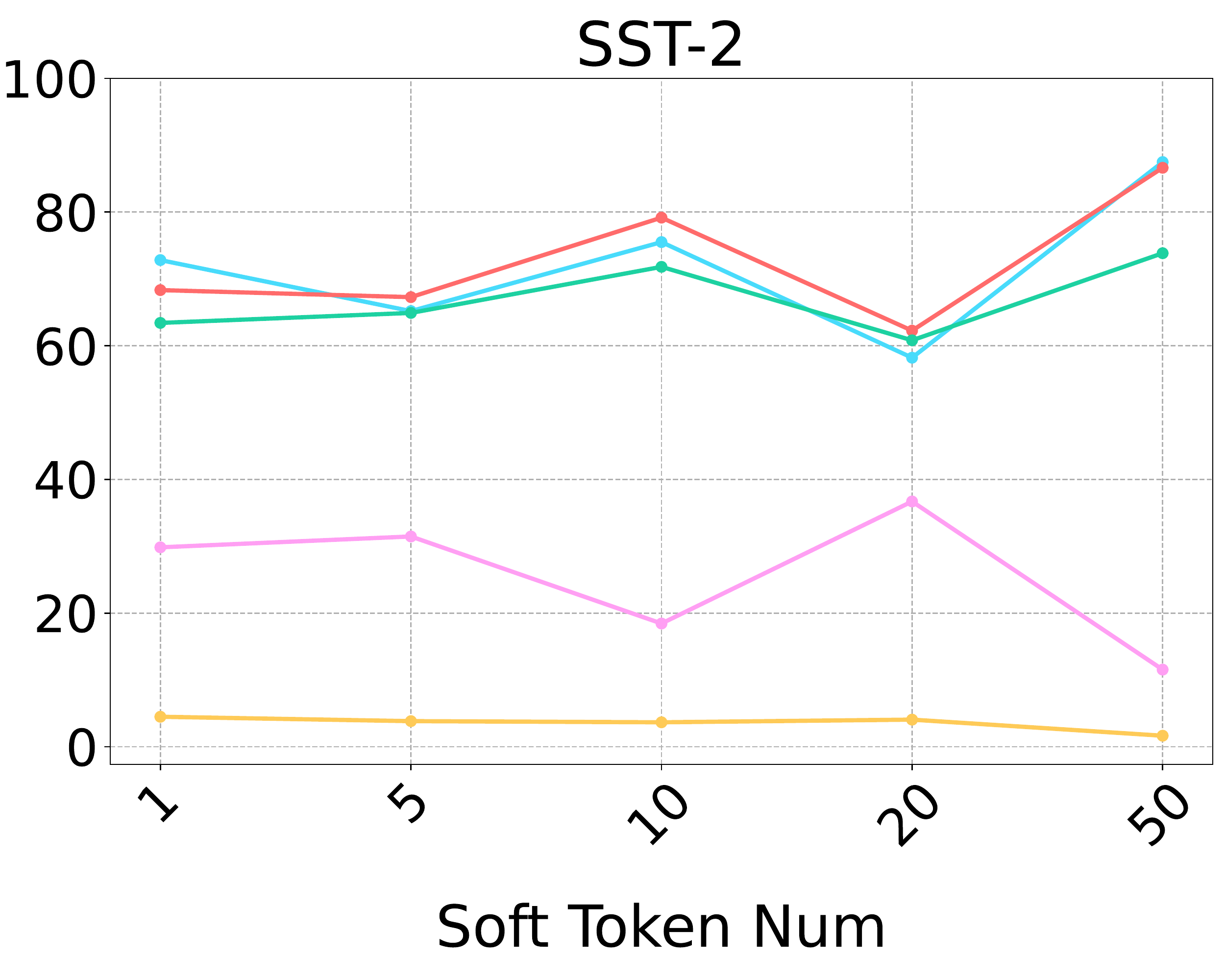}
         \label{fig:parameter-roberta-sst2}
     \end{subfigure}
     \begin{subfigure}[b]{0.33\textwidth}
         \centering
         \includegraphics[width=\textwidth]{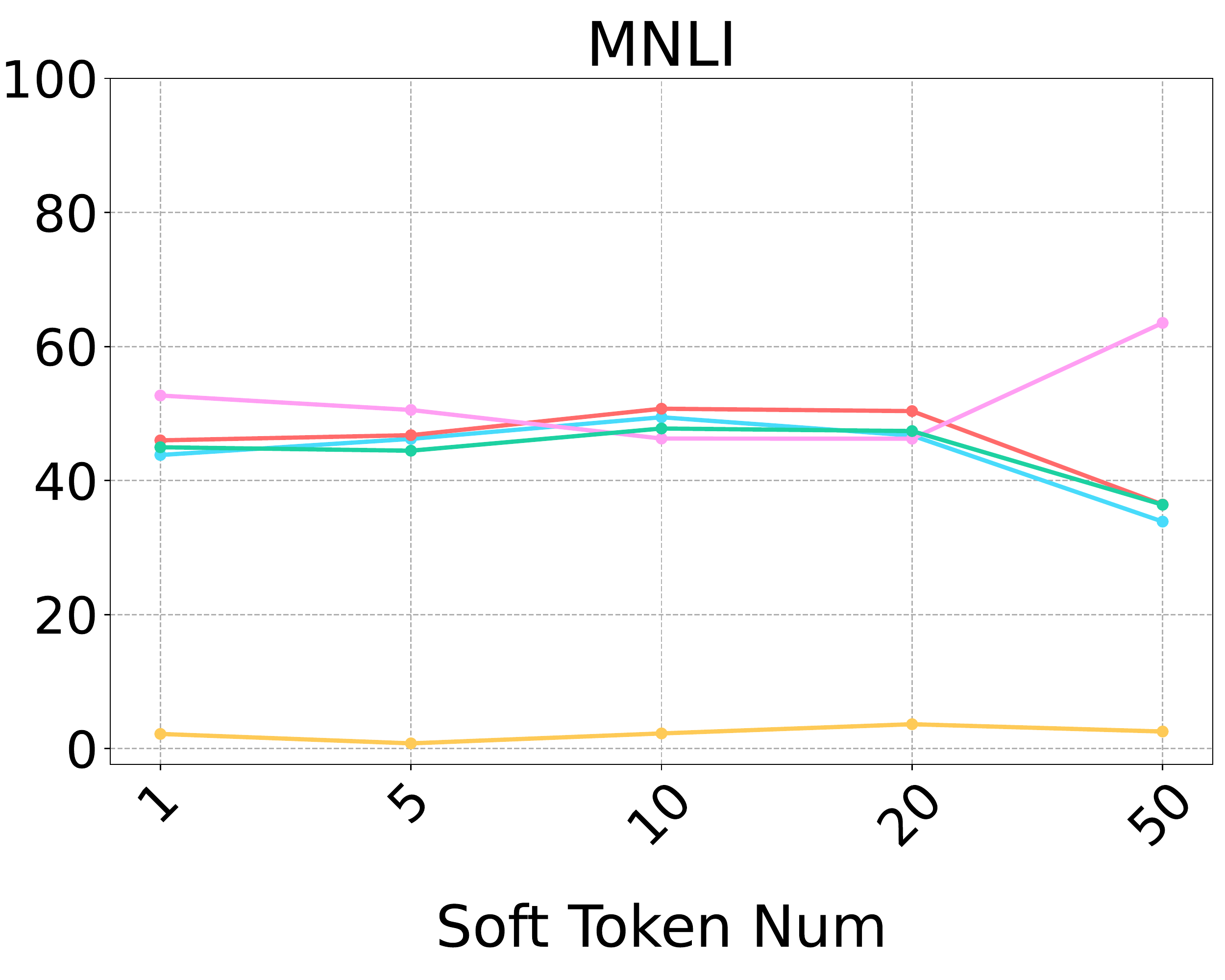}
         \label{fig:parameter-roberta-mnli}
     \end{subfigure}
    
    \vspace{-5pt}
    \centering
     \begin{subfigure}[b]{0.33\textwidth}
         \centering
         \includegraphics[width=\textwidth]{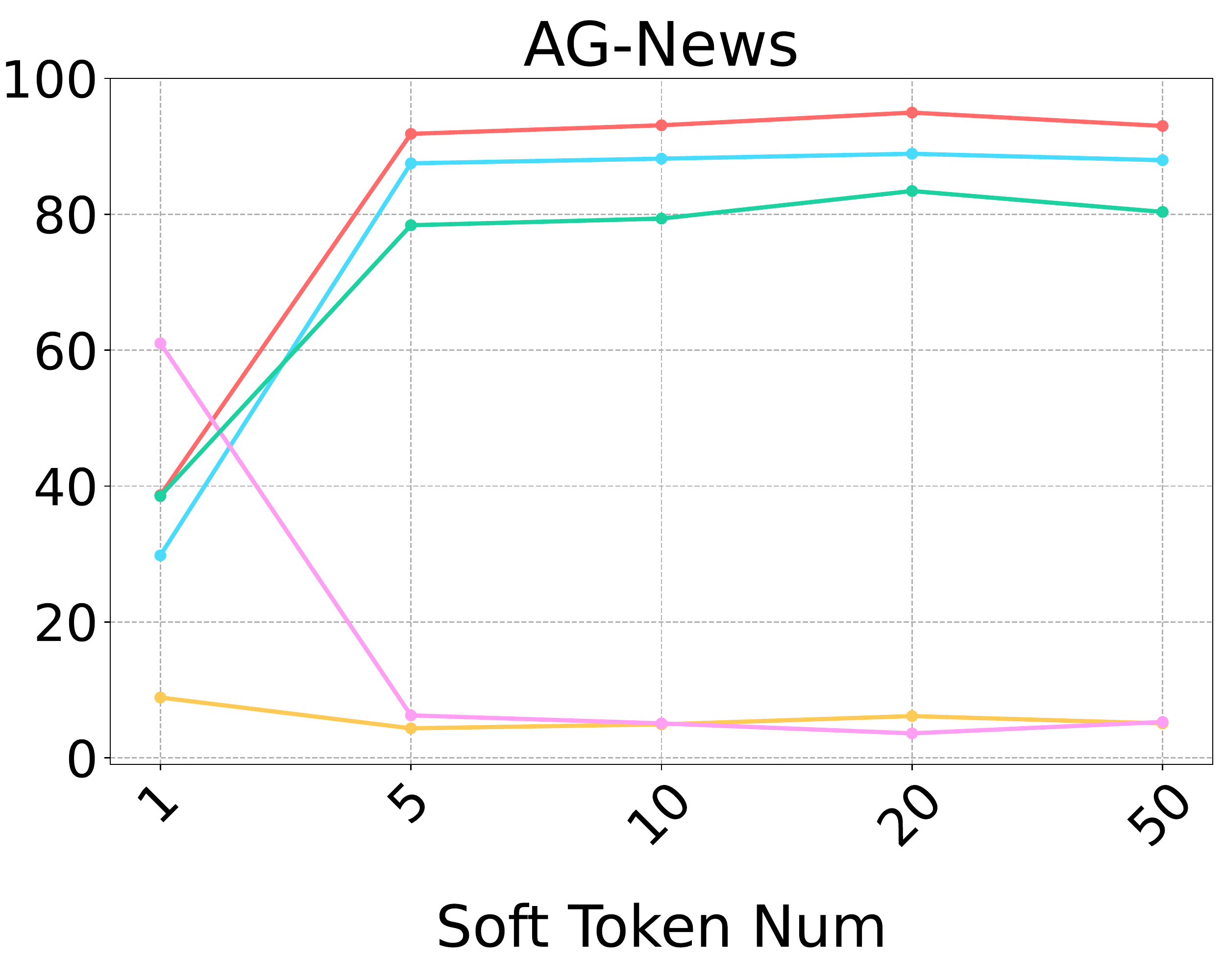}
         \label{fig:parameter-roberta-agnews}
     \end{subfigure}
     \begin{subfigure}[b]{0.33\textwidth}
         \centering
         \includegraphics[width=\textwidth]{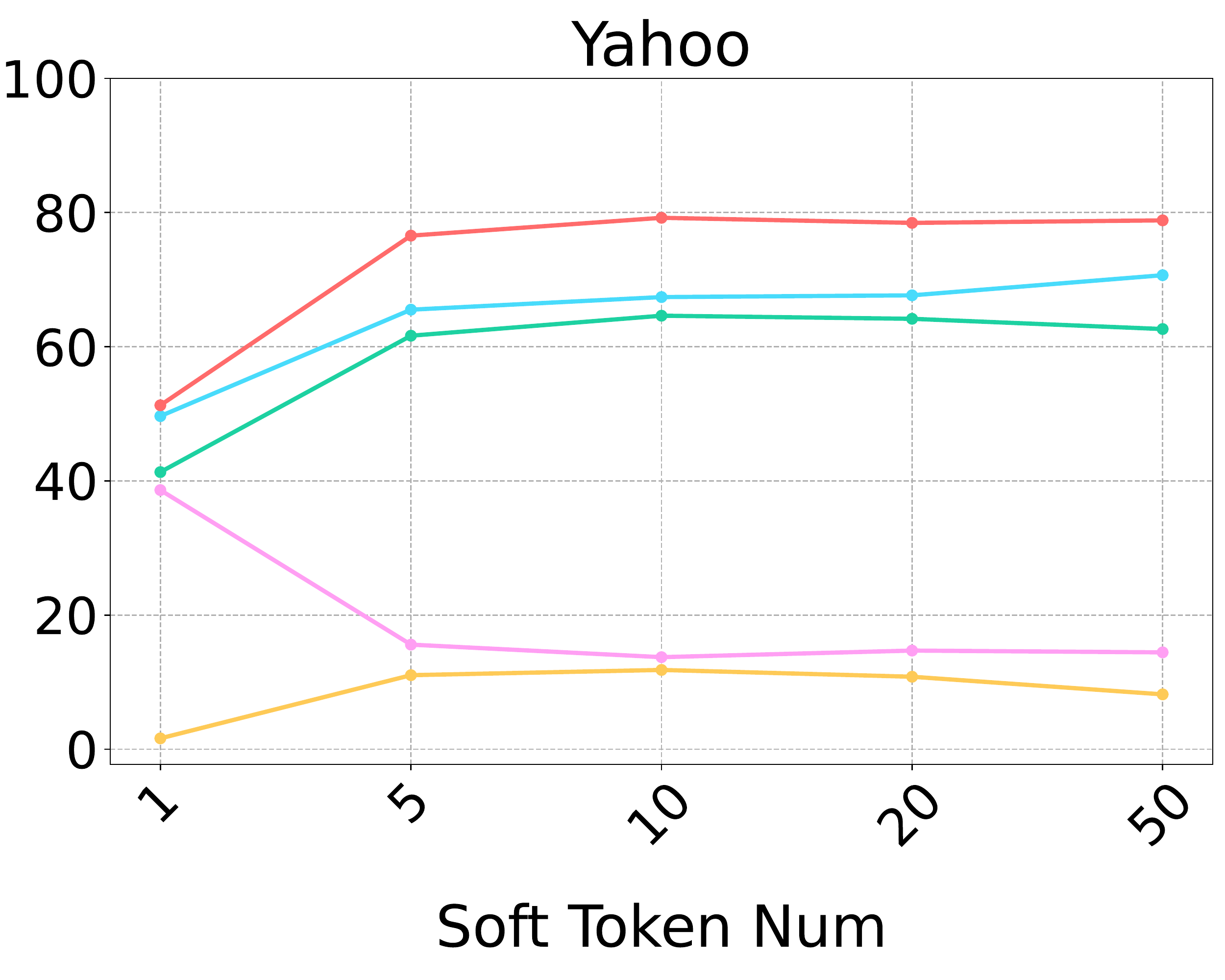}
         \label{fig:parameter-roberta-yahoo}
     \end{subfigure}

     \vspace{-18pt}
     \begin{subfigure}[b]{0.6\textwidth}
         \centering
         \includegraphics[width=\textwidth]{figures/legend.pdf}
     \end{subfigure}
     
    \vspace{-5pt}
    \caption{Results of tunable parameters with RoBERTa (Soft-prompt).}
    \label{fig:parameter-roberta_soft}
\end{figure*}

\section{Datasets}
\label{sec:appendix_dataset}
In this section, we describe the datasets adopted in experiments by tasks. 
The dataset statistics are shown in Table \ref{tab:statistics}.
The manual templates and verbalizers are presented in Table~\ref{tab:template}.


\paragraph{Sentiment analysis.}
\textbf{SST} \citep{sst2013socher} is a sentence-level corpus of movie reviews, where each sentence is labeled as \textit{negative}, \textit{somewhat negative}, \textit{neutral}, \textit{somewhat positive}, or \textit{positive}. \textbf{SST-5} contains the complete corpus with all five labels, while \textbf{SST-2} discards the label \textit{neutral} and polarizes the remaining 4 classes, i.e., negative or somewhat negative vs.~somewhat positive or positive.
\textbf{Amazon Fine Foods} \citep{amazon2013mcauley}, denoted as \textbf{Amazon} for simplicity throughout the paper, is a sentiment analysis dataset of reviews on fine foods from Amazon. Due to the enormous dataset size in the dataset, we sample 10k samples per class from the dataset.
\textbf{SemEval} 2016 Task 4 \citep{semeval2016nakov} is the sentiment analysis in the Twitter task. We consider Subtask A, where all Twitter texts are labeled as negative, neutral, or positive. 
\textbf{Dynasent} \citep{dynasent2021potts} is a challenging and dynamically evolved dataset, adopting human-in-the-loop efforts in dataset construction. We merge the data of round 1 and round 2 in our experiments.
\paragraph{Natural language inference.}
\textbf{MNLI} \citep{mnli2018williams} consists of 10 types of written and spoken English data and has two versions called matched and mismatched respectively, according to whether the domain of the train set and dev/test set is matched. We use the matched version in our experiment.
\textbf{HANS} \citep{hans2019mccoy} is a heuristic analysis dataset for NLI systems, based on the specific hypotheses about invalid heuristics that may be captured by the NLI model.
\textbf{ANLI} \citep{anli2020nie} is an adversarial NLI dataset, created by an iterative (three rounds in total), human-and-model-in-the-loop procedure.  We merge the data from all three rounds in our experiments.
\paragraph{Topic classification.}
\textbf{Yahoo Topic Answers} \citep{DBLP:conf/nips/ZhangZL15} contains 10 categories of questions and their corresponding answers from the Yahoo! Webscope program. For each sample, the title and content of the question are concatenated as one text, and the best answer to the question is used as a label. Since the original training dataset is extremely large (1.4 million samples for each category), we randomly sample 140,000 samples for simplicity.
\textbf{AG News} \citep{DBLP:conf/nips/ZhangZL15} is a corpus of news articles consisting of 4 classes: World, Sports, Business, and Science/Technology. For each article, we construct the text by concatenating the title and description.

\paragraph{Toxic detection.}
\textbf{Civil Comments}\footnote{\url{https://www.kaggle.com/competitions/jigsaw-unintended-bias-in-toxicity-\\classification}} is collected from the Civil Comments platform. Each comment is annotated with a float toxicity score, scaling from 0 to 1. We follow the official instructions to set samples with a toxicity score smaller than 0.5 as label 0 and vice versa.
\textbf{Hate Speech} \citep{hatespeech2018de-gibert}, the arguably most popular dataset in toxic detection, is collected from Stormfront, a large forum of white nationalists. The test set we use is sampled by the author in the official Github repository. 
\textbf{Implicit Hate} \citep{implicithate2021elsherief} consists of hate tweets from extremist groups in the US. Notably, a part of the hate tweets is implicit, which contains some subtle tricks to conceal the toxicity and evade keyword detection.
\paragraph{Plain text.}
\textbf{BookCorpus} \citep{zhu2015bookcorpus} collects a tremendous number of free novel books and thus is used in the pre-training stage of pre-trained language models. We sample 10k texts for evaluation.
\textbf{Random Words} contains 1k meaningless texts, each synthesized by concatenating 20 random words.

\section{Additional Results of Control Experiments}
\label{sec:appendix_additional_first}
For the empirical control study in the influence of six factors on PLMs calibration, we provide additional experimental results.
%
The results of T5-base on AG News are shown in Fig.\ref{fig:t5-agnews}, Fig.\ref{fig:scale-t5-agnews}, Fig.\ref{fig:pretrain-t5-agnews}, and Fig.\ref{fig:soft_prompt-t5}.
The results of RoBERTa-base are shown in Fig.\ref{fig:kshots-roberta}, Fig.\ref{fig:dynamics-roberta}, Fig.\ref{fig:scale-bert}, Fig.\ref{fig:pretrain_roberta}, Fig.\ref{fig:parameter-roberta}, and Fig.\ref{fig:parameter-roberta_soft}.
We discuss detailed experimental settings and conclusions for each considered factor.


\paragraph{Available training samples.}
We adopt K-shot learning, where K is the number of samples per class.
We experiment with each K five times on each dataset and report the average performance due to the potential variance in the few-shot setting. 
In this dimension, we additionally find that the trends in average confidence are different in the two model architectures. 
While T5 has an obvious confidence drop in the early stage, the confidence of RoBERTa seems to continually increase along with the number of available training samples. This can be partially explained by the stronger few-shot adaptation of RoBERTa since we observe that the performance of RoBERTa is significantly higher in extreme cases (e.g., K=1,2,4).

\label{sec:exp_shot}

\paragraph{Training dynamics.}
\looseness=-1
We decompose the whole training process into steps, and measure five metrics during some fixed intervals. 
In this dimension, the conclusion is consistent with the general one.

\paragraph{Number of tunable parameters.}
To quantitatively explore the influence of the number of tunable parameters on PLMs calibration, we employ the parameter efficient tuning methods in NLP~\citep{DBLP:conf/icml/HoulsbyGJMLGAG19, DBLP:conf/acl/ZakenGR22, ding2022delta}. 
We adopt Soft-prompt~\citep{DBLP:conf/emnlp/LesterAC21} and Adapter~\citep{DBLP:conf/icml/HoulsbyGJMLGAG19} tuning due to their simplicity, stability, and practicality. 
We experiment with various numbers of soft tokens and bottleneck dimensions of the inserted adapter modules. 
Only the parameters in the soft tokens and adapter module are tunable.

We summarize the extra findings as follows:~(1) Soft-prompt and Adapter tuning show different trends spanning four datasets;
(2) For Soft-prompt tuning, the model performance and confidence increase continually with more tunable parameters. We can observe that the increasing rates are nearly matched, thus decreasing ECE continually.  
The negative effect is also the increase in CErr$_{neg}$ due to the overconfidence in wrong predictions. This is consistent with the trend we observed in the under-fitted state;
(3) The world in Adapter tuning is different, where increasing capacity cannot bring substantial performance gains. This is due to the strong capacity of Adapter.
However, the overall confidence continues to increase given more capacity, resulting in increasing ECE and CErr$_{neg}$, while the performance stays constant.  This is consistent with the trend we observed in the over-fittied state;
(4) The implication of experimental results is that blindly increasing model capacity may negatively impact PLMs calibration, especially at the critical point when current capacity is sufficient to solve the task well. 
\label{sec:increasing_scale}

\begin{table*}[]

\centering
\resizebox{\textwidth}{!}{
\begin{tabular}{@{}l|ll|ccccc|ccccc|ccccc@{}}
\toprule
\multirow{12}{*}{Dynasent} & \multicolumn{2}{l|}{Dataset}                                  & \multicolumn{5}{c|}{Dynasent}                                           & \multicolumn{5}{c|}{Amazon}                                             & \multicolumn{5}{c}{DSC}                                                 \\ \cmidrule(l){2-18} 
                           & \multicolumn{2}{l|}{Method}                                   & Acc            & Conf  & ECE           & CErr$_{pos}$  & CErr$_{neg}$   & Acc            & Conf  & ECE           & CErr$_{pos}$  & CErr$_{neg}$   & Acc            & Conf  & ECE           & CErr$_{pos}$  & CErr$_{neg}$   \\ \cmidrule(l){2-18} 
                           & \multicolumn{1}{l|}{\multirow{5}{*}{Unlearnable}} & Vanilla   & 78.45          & 86.83 & 8.38          & 9.94          & 75.07          & 86.57          & 95.28 & 8.71          & 3.44          & 87.02          & \textbf{90.00} & 94.40 & 4.48          & 4.10          & 80.85          \\
                           & \multicolumn{1}{l|}{}                             & TS        & 78.45          & 79.10 & \textbf{1.02} & 17.37         & 66.27          & 86.57          & 89.92 & 3.36          & 8.59          & 80.31          & \textbf{90.00} & 89.26 & \textbf{0.78} & 8.90          & 72.68          \\
                           & \multicolumn{1}{l|}{}                             & LS        & \textbf{78.47} & 78.22 & 3.64          & 18.89         & 67.69          & 86.55          & 85.48 & 3.42          & 13.35         & 77.91          & 89.75          & 84.61 & 5.31          & 13.95         & 72.02          \\
                           & \multicolumn{1}{l|}{}                             & EDA       & 76.30          & 89.20 & 12.91         & \textbf{7.76} & 79.44          & \textbf{87.19} & 97.07 & 9.88          & \textbf{1.75} & 89.04          & 88.05          & 95.50 & 7.45          & \textbf{2.81} & 83.03          \\
                           & \multicolumn{1}{l|}{}                             & Ensemble  & 78.18          & 86.76 & 8.58          & 9.89          & 74.75          & 86.37          & 95.02 & 8.66          & 3.71          & 86.99          & 89.74          & 94.27 & 4.56          & 4.17          & 80.67          \\ \cmidrule(l){2-18} 
                           & \multicolumn{1}{l|}{\multirow{5}{*}{Learnable}}   & E-MLP     & 78.45          & 78.99 & 4.45          & 21.05         & 79.11          & 86.57          & 83.15 & \textbf{2.92} & 16.85         & 83.14          & \textbf{90.00} & 82.53 & 7.17          & 17.48         & 82.63          \\
                           & \multicolumn{1}{l|}{}                             & E-T5 \ours     & 78.45          & 61.63 & 18.26         & 33.00         & 42.07          & 86.57          & 89.99 & 6.51          & 6.94          & 71.00          & \textbf{90.00} & 86.14 & 6.19          & 11.03         & 61.60          \\
                           & \multicolumn{1}{l|}{}                             & I-Vanilla & \textbf{78.47} & 61.95 & 17.91         & 32.77         & 42.72          & 84.44          & 89.89 & 6.52          & 6.18          & 68.52          & 88.84          & 86.15 & 5.76          & 10.77         & 61.69          \\
                           & \multicolumn{1}{l|}{}                             & I-Iter \ours   & 77.92          & 61.45 & 16.47         & 33.26         & 42.78          & 86.03          & 86.92 & 2.99          & 9.99          & \textbf{67.91} & 89.45          & 84.72 & 4.88          & 12.54         & 61.55          \\
                           & \multicolumn{1}{l|}{}                             & I-Simul \ours  & 78.13          & 66.36 & 24.59         & 25.51         & \textbf{37.34} & 85.67          & 91.26 & 13.29         & 5.28          & 70.59          & 88.61          & 87.83 & 12.46         & 8.41          & \textbf{58.61} \\ \bottomrule
\end{tabular}
}

\caption{Results T5's calibration performance under hard-to-easy distribution shifts.}
\label{tab:hard2easyt5}
\vspace{-10pt}
\end{table*}


\begin{table*}[t!]
\centering

\resizebox{\textwidth}{!}{
\begin{tabular}{@{}l|ll|ccccc|ccccc|ccccc@{}}
\toprule
\multirow{12}{*}{MNLI}   & \multicolumn{2}{l|}{Dataset}                                  & \multicolumn{5}{c|}{MNLI}                                               & \multicolumn{5}{c|}{HANS}                                                & \multicolumn{5}{c}{ANLI}                                                 \\ \cmidrule(l){2-18} 
                         & \multicolumn{2}{l|}{Method}                                   & Acc            & Conf  & ECE           & CErr$_{pos}$  & CErr$_{neg}$   & Acc            & Conf  & ECE            & CErr$_{pos}$  & CErr$_{neg}$   & Acc            & Conf  & ECE            & CErr$_{pos}$  & CErr$_{neg}$   \\ \cmidrule(l){2-18} 
                         & \multicolumn{1}{l|}{\multirow{5}{*}{Unlearnable}} & Vanilla   & 85.90          & 96.24 & 9.50          & 2.40          & 87.36          & 54.17          & 95.09 & 39.68          & 2.71          & 92.36          & 29.78          & 90.94 & 61.14          & 11.28         & 91.90          \\
                         & \multicolumn{1}{l|}{}                             & TS        & 85.90          & 86.65 & \textbf{0.90} & 11.09         & 71.84          & 54.17          & 82.15 & 26.74          & 15.43         & 79.16          & 29.78          & 75.57 & 45.57          & 27.32         & 76.80          \\
                         & \multicolumn{1}{l|}{}                             & LS        & 86.28          & 86.88 & 4.43          & 11.92         & 79.31          & 55.59          & 86.96 & 31.37          & 11.47         & 85.00          & 29.25          & 81.59 & 52.37          & 20.23         & 82.34          \\
                         & \multicolumn{1}{l|}{}                             & EDA       & 85.99          & 97.07 & 11.09         & \textbf{1.78} & 90.05          & \textbf{58.24} & 96.87 & 38.63          & \textbf{1.91} & 95.16          & \textbf{31.34} & 92.00 & 60.66          & \textbf{8.81} & 92.38          \\
                         & \multicolumn{1}{l|}{}                             & Ensemble  & 86.60          & 96.32 & 9.74          & 2.37          & 87.90          & 56.09          & 96.44 & 40.35          & 2.00          & 94.45          & 30.06          & 90.47 & 60.46          & 11.38         & 91.26          \\ \cmidrule(l){2-18} 
                         & \multicolumn{1}{l|}{\multirow{5}{*}{Learnable}}   & E-MLP     & 85.90          & 85.82 & 13.73         & 14.16         & 85.67          & 54.17          & 81.92 & 29.36          & 17.87         & 81.66          & 29.78          & 81.49 & 51.71          & 18.88         & 81.65          \\
                         & \multicolumn{1}{l|}{}                             & E-T5 \ours     & 85.90          & 74.37 & 18.51         & 18.93         & 33.58          & 54.17          & 74.47 & 28.79          & 10.10         & 56.23          & 29.78          & 35.21 & 45.46          & 74.72         & 39.43          \\
                         & \multicolumn{1}{l|}{}                             & I-Vanilla & 85.76          & 75.23 & 18.25         & 18.32         & 36.45          & 57.28          & 77.14 & 32.26          & 13.23         & 64.23          & 28.63          & 37.14 & 44.78          & 71.91         & 40.77          \\
                         & \multicolumn{1}{l|}{}                             & I-Iter \ours    & \textbf{86.63} & 60.04 & 26.59         & 33.85         & \textbf{20.41} & 53.70          & 57.77 & \textbf{21.70} & 29.34         & \textbf{42.82} & 31.06          & 21.29 & \textbf{31.88} & 83.71         & \textbf{23.55} \\
                         & \multicolumn{1}{l|}{}                             & I-Simul \ours   & 86.46          & 74.81 & 18.91         & 18.49         & 32.01          & 56.65          & 75.84 & 33.83          & 13.79         & 62.28          & 29.16          & 38.67 & 45.44          & 66.86         & 40.95          \\ \midrule
\multirow{12}{*}{Amazon} & \multicolumn{2}{l|}{Dataset}                                  & \multicolumn{5}{c|}{Amazon}                                             & \multicolumn{5}{c|}{SST-5}                                               & \multicolumn{5}{c}{SemEval}                                              \\ \cmidrule(l){2-18} 
                         & \multicolumn{2}{l|}{Method}                                   & Acc            & Conf  & ECE           & CErr$_{pos}$  & CErr$_{neg}$   & Acc            & Conf  & ECE            & CErr$_{pos}$  & CErr$_{neg}$   & Acc            & Conf  & ECE            & CErr$_{pos}$  & CErr$_{neg}$   \\ \cmidrule(l){2-18} 
                         & \multicolumn{1}{l|}{\multirow{5}{*}{Unlearnable}} & Vanilla   & 90.90          & 98.17 & 7.28          & 1.09          & 90.84          & \textbf{70.29} & 94.29 & 24.05          & 3.95          & 90.14          & 56.02          & 90.45 & 34.43          & 7.05          & 87.26          \\
                         & \multicolumn{1}{l|}{}                             & TS        & 90.90          & 89.66 & \textbf{2.02} & 8.73          & 73.58          & \textbf{70.29} & 78.15 & \textbf{7.91}  & 18.42         & 70.04          & 56.02          & 70.34 & \textbf{14.32} & 25.98         & 65.65          \\
                         & \multicolumn{1}{l|}{}                             & LS        & 91.89          & 88.50 & 6.71          & 10.64         & 78.83          & 69.92          & 84.01 & 14.20          & 14.38         & 80.28          & 55.17          & 81.64 & 26.47          & 15.46         & 78.08          \\
                         & \multicolumn{1}{l|}{}                             & EDA       & \textbf{92.39} & 98.34 & 5.95          & \textbf{0.92} & 89.46          & 66.64          & 93.98 & 27.34          & \textbf{3.82} & 89.57          & \textbf{57.05} & 93.45 & 36.43          & \textbf{4.37} & 90.56          \\
                         & \multicolumn{1}{l|}{}                             & Ensemble  & 91.69          & 98.19 & 6.50          & 1.06          & 89.93          & 69.56          & 93.67 & 24.22          & 4.24          & 88.93          & 55.94          & 90.14 & 34.23          & 7.19          & 86.76          \\ \cmidrule(l){2-18} 
                         & \multicolumn{1}{l|}{\multirow{5}{*}{Learnable}}   & E-MLP     & 90.90          & 95.08 & 9.14          & 4.94          & 95.34          & \textbf{70.29} & 83.57 & 22.22          & 16.18         & 82.99          & 56.02          & 77.12 & 25.42          & 22.49         & 76.63          \\
                         & \multicolumn{1}{l|}{}                             & E-T5 \ours     & 90.90          & 71.97 & 19.27         & 21.20         & 3.72           & \textbf{70.29} & 32.10 & 45.94          & 61.74         & 17.53          & 56.02          & 23.64 & 36.13          & 64.58         & 8.63           \\
                         & \multicolumn{1}{l|}{}                             & I-Vanilla & 88.00          & 71.60 & 17.13         & 19.18         & 3.97           & 64.85          & 26.74 & 46.32          & 65.75         & \textbf{12.86} & 44.43          & 17.51 & 31.05          & 66.92         & \textbf{5.07}  \\
                         & \multicolumn{1}{l|}{}                             & I-Iter \ours    & 90.11          & 71.34 & 18.88         & 21.18         & \textbf{3.24}  & 66.54          & 34.13 & 41.70          & 58.17         & 18.82          & 53.28          & 34.05 & 27.10          & 48.25         & 13.86          \\
                         & \multicolumn{1}{l|}{}                             & I-Simul \ours   & 90.60          & 71.07 & 19.80         & 21.91         & 3.41           & 69.35          & 33.96 & 44.16          & 58.75         & 17.46          & 53.50          & 24.20 & 33.35          & 61.15         & 7.36           \\ \midrule
\multirow{12}{*}{Civil}  & \multicolumn{2}{l|}{Dataset}                                  & \multicolumn{5}{c|}{Civil}                                              & \multicolumn{5}{c|}{Hate Speech}                                         & \multicolumn{5}{c}{Implicit Hate}                                        \\ \cmidrule(l){2-18} 
                         & \multicolumn{2}{l|}{Method}                                   & Acc            & Conf  & ECE           & CErr$_{pos}$  & CErr$_{neg}$   & Acc            & Conf  & ECE            & CErr$_{pos}$  & CErr$_{neg}$   & Acc            & Conf  & ECE            & CErr$_{pos}$  & CErr$_{neg}$   \\ \cmidrule(l){2-18} 
                         & \multicolumn{1}{l|}{\multirow{5}{*}{Unlearnable}} & Vanilla   & 86.94          & 98.15 & 10.09         & \textbf{1.14} & 92.91          & 76.99          & 98.22 & 21.94          & \textbf{1.22} & 96.41          & \textbf{62.88} & 96.37 & 32.02          & 2.88          & 95.00          \\
                         & \multicolumn{1}{l|}{}                             & TS        & 86.94          & 90.94 & \textbf{2.88} & 7.29          & 77.87          & 76.99          & 89.70 & 13.34          & 8.58          & 84.16          & \textbf{62.88} & 85.50 & \textbf{21.15} & 12.72         & 82.28          \\
                         & \multicolumn{1}{l|}{}                             & LS        & \textbf{87.91} & 87.73 & 9.52          & 11.79         & 84.24          & \textbf{78.45} & 88.31 & \textbf{10.86} & 11.48         & 87.54          & 62.58          & 86.79 & 24.21          & 12.82         & 86.13          \\
                         & \multicolumn{1}{l|}{}                             & EDA       & 83.61          & 97.01 & 13.40         & 2.08          & 92.35          & 77.82          & 97.28 & 19.65          & 2.30          & 95.82          & 61.53          & 96.68 & 35.14          & \textbf{2.71} & 95.70          \\
                         & \multicolumn{1}{l|}{}                             & Ensemble  & 86.45          & 97.96 & 11.52         & 1.29          & 93.16          & 76.32          & 97.58 & 21.28          & 1.75          & 95.41          & 62.77          & 96.19 & 33.42          & 3.08          & 94.97          \\ \cmidrule(l){2-18} 
                         & \multicolumn{1}{l|}{\multirow{5}{*}{Learnable}}   & E-MLP     & 86.94          & 91.93 & 12.24         & 8.09          & 92.01          & 76.99          & 88.52 & 19.66          & 11.62         & 88.98          & \textbf{62.88} & 83.08 & 25.45          & 17.15         & 83.47          \\
                         & \multicolumn{1}{l|}{}                             & E-T5 \ours     & 86.94          & 70.97 & 15.99         & 18.62         & 1.68           & 76.99          & 46.28 & 48.83          & 52.25         & 41.37          & \textbf{62.88} & 30.90 & 41.57          & 59.84         & 15.20          \\
                         & \multicolumn{1}{l|}{}                             & I-Vanilla & 77.92          & 69.06 & 8.92          & 11.60         & \textbf{0.83}  & 76.99          & 45.25 & 49.59          & 53.24         & \textbf{40.21} & 58.12          & 29.51 & 38.32          & 58.58         & \textbf{13.00} \\
                         & \multicolumn{1}{l|}{}                             & I-Iter \ours    & 85.40          & 75.36 & 10.31         & 12.18         & 2.48           & 76.15          & 50.43 & 49.62          & 50.02         & 51.84          & 60.59          & 34.15 & 38.04          & 54.50         & 16.69          \\
                         & \multicolumn{1}{l|}{}                             & I-Simul \ours   & 87.25          & 70.69 & 16.65         & 19.22         & 1.71           & 78.24          & 45.86 & 50.64          & 53.36         & 43.03          & 62.56          & 29.60 & 41.56          & 60.57         & 13.17          \\ \bottomrule
\end{tabular}
}
\caption{Results of RoBERTa's calibration performance under standard distribution shifts.}
\label{tab:roberta-shift}
\end{table*}

\paragraph{Model scale.}
We consider the scaling law and experiment with various model sizes.
For T5, we choose models with small, base, large, and 3b sizes. 
For RoBERTa, we choose models with tiny, mini, small, medium, base, and large sizes. 
Our results support the ``scaling improves calibration'' conclusion in some cases.
We observe that ECE decreases when larger capacity brings substantial improvement to PLMs' performance (e.g., T5 on SST-2 and MNLI). 
However, when the performance reaches a plateau value, increasing capacity only boosts PLMs' confidence (e.g., T5 and RoBERTa on Yahoo).
In this case, the ECE increases when the PLM’s scale keeps increasing.




\begin{table*}[thb]

\resizebox{\textwidth}{!}{

\begin{tabular}{@{}l|l|ccccc|ccccc|ccccc@{}}
\toprule
Model Scale               & Dataset   & \multicolumn{5}{c|}{Amazon}                                             & \multicolumn{5}{c|}{SST-5}                                                & \multicolumn{5}{c}{SemEval}                                              \\ \midrule
\multirow{6}{*}{T5-small} & Method    & Acc            & Conf  & ECE           & CErr$_{pos}$   & CErr$_{neg}$  & Acc            & Conf  & ECE            & CErr$_{pos}$   & CErr$_{neg}$   & Acc            & Conf  & ECE            & CErr$_{pos}$   & CErr$_{neg}$  \\ \cmidrule(l){2-17} 
                          & E-MLP     & 87.65          & 86.41 & \textbf{4.78} & \textbf{13.59} & 86.43         & \textbf{65.14} & 80.15 & \textbf{15.23} & \textbf{19.86} & 80.17          & 49.23          & 77.14 & 27.91          & \textbf{22.89} & 77.17         \\
                          & E-T5 \ours     & 87.65          & 67.80 & 19.85         & 23.71          & 7.49          & \textbf{65.14} & 28.16 & 37.29          & 64.06          & 13.63          & 49.23          & 30.45 & 19.40          & 50.65          & 12.12         \\
                          & I-Vanilla & 81.64          & 57.35 & 24.28         & 30.30          & \textbf{2.45} & 55.01          & 3.95  & 51.21          & 93.35          & \textbf{0.66}  & 44.57          & 2.17  & 42.43          & 95.53          & \textbf{0.32} \\
                          & I-Iter \ours    & 87.54          & 68.20 & 19.33         & 22.89          & 5.66          & 64.10          & 28.81 & 36.99          & 62.99          & 14.16          & 48.52          & 32.05 & \textbf{17.49} & 47.86          & 13.13         \\
                          & I-Simul \ours   & \textbf{87.66} & 68.61 & 19.05         & 22.63          & 6.35          & 64.57          & 29.59 & 37.57          & 62.38          & 14.95          & \textbf{50.38}          & 35.00 & 18.89          & 45.87          & 15.58         \\ \midrule
\multirow{6}{*}{T5-base}  & Method    & Acc            & Conf  & ECE           & CErr$_{pos}$   & CErr$_{neg}$  & Acc            & Conf  & ECE            & CErr$_{pos}$   & CErr$_{neg}$   & Acc            & Conf  & ECE            & CErr$_{pos}$   & CErr$_{neg}$  \\ \cmidrule(l){2-17} 
                          & E-MLP     & 91.00          & 90.44 & \textbf{4.35} & \textbf{9.56}  & 90.41         & 69.73          & 85.18 & \textbf{15.45} & \textbf{14.69} & 84.87          & 55.03          & 78.39 & 23.36          & \textbf{21.63} & 78.42         \\
                          & E-T5 \ours     & 91.00          & 71.03 & 19.97         & 22.40          & 4.63          & 69.73          & 31.73 & 38.80          & 61.80          & 16.83          & 55.03          & 29.72 & 26.28          & 56.23          & 12.54         \\
                          & I-Vanilla & 88.25          & 70.91 & 17.34         & 20.16          & \textbf{3.86} & 63.07          & 29.81 & 34.08          & 59.42          & \textbf{11.42} & 48.08          & 25.32 & 23.69          & 55.53          & \textbf{7.59} \\
                          & I-Iter \ours    & \textbf{91.69} & 71.76 & 19.93         & 22.23          & 5.43          & 68.23          & 33.46 & 36.87          & 59.79          & 18.96          & \textbf{56.23} & 35.21 & \textbf{21.42} & 50.98          & 17.48         \\
                          & I-Simul \ours   & 91.38          & 70.92 & 20.47         & 22.80          & 4.30          & \textbf{70.29} & 32.03 & 42.12          & 60.65          & 14.72          & 54.75          & 26.18 & 30.70          & 59.34          & 8.67          \\ \midrule
\multirow{6}{*}{T5-large} & Method    & Acc            & Conf  & ECE           & CErr$_{pos}$   & CErr$_{neg}$  & Acc            & Conf  & ECE            & CErr$_{pos}$   & CErr$_{neg}$   & Acc            & Conf  & ECE            & CErr$_{pos}$   & CErr$_{neg}$  \\ \cmidrule(l){2-17} 
                          & E-MLP     & 91.58          & 91.95 & \textbf{4.70} & \textbf{8.04}  & 91.89         & \textbf{73.85} & 83.52 & \textbf{10.24} & \textbf{16.52} & 83.61          & 56.65          & 78.26 & \textbf{21.61} & \textbf{21.74} & 78.26         \\
                          & E-T5 \ours     & 91.58          & 70.10 & 21.48         & 23.70          & 2.66          & \textbf{73.85} & 29.96 & 47.35          & 64.65          & 14.75          & 56.65          & 28.56 & 29.98          & 57.52          & 10.36         \\
                          & I-Vanilla & 88.88          & 69.42 & 19.46         & 22.12          & \textbf{1.81} & 71.79          & 28.30 & 46.83          & 65.12          & \textbf{11.55} & 49.00          & 24.66 & 25.95          & 56.30          & \textbf{6.37} \\
                          & I-Iter \ours    & 92.96          & 88.26 & 10.48         & 8.74           & 48.71         & 72.45          & 70.35 & 30.29          & 25.22          & 58.71          & \textbf{58.08} & 84.26 & 35.21          & 12.77          & 80.14         \\
                          & I-Simul \ours   & \textbf{93.34} & 74.45 & 19.39         & 20.62          & 5.43          & 73.66          & 36.92 & 45.40          & 57.27          & 20.66          & 56.87          & 40.04 & 28.43          & 44.23          & 19.29         \\ \bottomrule
\end{tabular}

}
\caption{Results of T5's calibration performance with increasing model scales.}
\label{tab:emergent_scale}
\end{table*}

\begin{table*}

\centering
\resizebox{\textwidth}{!}{
\begin{tabular}{@{}ll|cccccc|cccccc@{}}
\toprule
\multicolumn{2}{c|}{ID Dataset}                               & \multicolumn{6}{c|}{SST-2}                                                                                      & \multicolumn{6}{c}{Yahoo}                                                                                      \\ \midrule
\multicolumn{2}{c|}{OOD Dataset}                              & \multicolumn{2}{c|}{SST-2}           & \multicolumn{2}{c|}{Bookcorpus}      & \multicolumn{2}{c|}{Random Words} & \multicolumn{2}{c|}{Yahoo}           & \multicolumn{2}{c|}{Bookcorpus}      & \multicolumn{2}{c}{Random Words} \\ \midrule
\multicolumn{2}{c|}{Method}                                   & Conf  & \multicolumn{1}{c|}{Entropy} & Conf  & \multicolumn{1}{c|}{Entropy} & Conf           & Entropy          & Conf  & \multicolumn{1}{c|}{Entropy} & Conf  & \multicolumn{1}{c|}{Entropy} & Conf           & Entropy         \\ \midrule
\multicolumn{1}{l|}{\multirow{5}{*}{Unlearnable}} & Vanilla   & 98.04 & \multicolumn{1}{c|}{5.01}    & 93.38 & \multicolumn{1}{c|}{15.97}   & 84.46          & 34.95            & 82.76 & \multicolumn{1}{c|}{51.94}   & 47.62 & \multicolumn{1}{c|}{152.43}  & 56.95          & 126.54          \\
\multicolumn{1}{l|}{}                             & TS        & 93.89 & \multicolumn{1}{c|}{18.02}   & 85.07 & \multicolumn{1}{c|}{35.23}   & 72.49          & 54.69            & 75.72 & \multicolumn{1}{c|}{76.29}   & 38.43 & \multicolumn{1}{c|}{177.74}  & 47.70          & 154.00          \\
\multicolumn{1}{l|}{}                             & LS        & 88.64 & \multicolumn{1}{c|}{33.90}   & 83.65 & \multicolumn{1}{c|}{40.46}   & 72.31          & 55.30            & 74.35 & \multicolumn{1}{c|}{93.81}   & 44.29 & \multicolumn{1}{c|}{168.14}  & 54.08          & 145.94          \\
\multicolumn{1}{l|}{}                             & EDA       & 98.27 & \multicolumn{1}{c|}{4.33}    & 93.73 & \multicolumn{1}{c|}{15.45}   & 83.00          & 37.15            & 83.68 & \multicolumn{1}{c|}{46.75}   & 50.59 & \multicolumn{1}{c|}{141.92}  & 69.03          & 92.58           \\
\multicolumn{1}{l|}{}                             & Ensemble  & 97.96 & \multicolumn{1}{c|}{5.20}    & 93.21 & \multicolumn{1}{c|}{16.47}   & 82.75          & 37.87            & 82.41 & \multicolumn{1}{c|}{53.01}   & 48.29 & \multicolumn{1}{c|}{150.39}  & 55.87          & 130.57          \\ \midrule
\multicolumn{1}{l|}{\multirow{5}{*}{Learnable}}   & E-MLP     & 88.62 & \multicolumn{1}{c|}{35.37}   & 86.94 & \multicolumn{1}{c|}{38.69}   & 85.04          & 42.17            & 74.93 & \multicolumn{1}{c|}{-}       & 61.80 & \multicolumn{1}{c|}{-}       & 67.57          & -               \\
\multicolumn{1}{l|}{}                             & E-T5 \ours     & 55.96 & \multicolumn{1}{c|}{62.11}   & 56.35 & \multicolumn{1}{c|}{64.08}   & 64.02          & 60.32            & 60.29 & \multicolumn{1}{c|}{-}       & 13.64 & \multicolumn{1}{c|}{-}       & 22.56          & -               \\
\multicolumn{1}{l|}{}                             & I-Vanilla & 56.31 & \multicolumn{1}{c|}{62.13}   & 57.72 & \multicolumn{1}{c|}{63.99}   & 66.47          & 59.90            & 60.51 & \multicolumn{1}{c|}{-}       & 13.71 & \multicolumn{1}{c|}{-}       & 22.78          & -               \\
\multicolumn{1}{l|}{}                             & I-Iter \ours   & 43.43 & \multicolumn{1}{c|}{57.59}   & 43.24 & \multicolumn{1}{c|}{60.62}   & 56.07          & 61.10            & 61.35 & \multicolumn{1}{c|}{-}       & 20.62 & \multicolumn{1}{c|}{-}       & 39.08          & -               \\
\multicolumn{1}{l|}{}                             & I-Simul \ours  & 63.24 & \multicolumn{1}{c|}{10.50}   & 65.74 & \multicolumn{1}{c|}{2.25}    & 77.68          & 0.01             & 60.52 & \multicolumn{1}{c|}{-}       & 6.44  & \multicolumn{1}{c|}{-}       & 14.67          & -               \\ \bottomrule
\end{tabular}
}

\caption{Results on task-irrelevant inputs with T5. We don't report the entropy results of learnable methods when Yahoo is adopted as ID dataset since the class numbers are different in unlearnable (10 original classes in Yahoo) and learnable methods (2 classes), which will result in unfair comparison.}
\label{tab:entropyt5}
\end{table*}

\paragraph{Pretraining.}
We choose the pre-trained RoBERTa-base and pre-trained T5-base (Pretrained), and compare them with several non-pretrained models, including random initialized RoBERTa-base and T5-base (Random), BiLSTM (LSTM)~\citep{hochreiter1997long}, Term Frequency Inverse Document Frequency (TF-IDF)~\citep{luhn1957statistical}, and Bag-of-word (BoW)~\citep{harris1954distributional}.
We find that pretraining only reduces ECE on relative simpler datasets, like SST-2 and AG-News, but bring negligible benefits on MNLI and Yahoo. 
This finding shares the same ground with scaling experiments. 

\section{Construction of the Calibration Training Dataset}
\label{sec:appendix_construction_of_cal}
In this paper, we consider the classification tasks. 
The construction process can be extended to the natural language generation tasks. 
We have an annotated dataset $\mathbb{D}=\{(x_i,y_i)_{i=1}^{N}\}$ for the standard training on the classification tasks. 
We typically fit a model $\mathcal{F}$ on the training dataset by minimizing the pre-defined loss (e.g., cross-entropy loss). 
We denote the original task as the main task. 
Then for the newly introduced calibration task, we need to generate a calibration training dataset $\mathbb{D}^*$ for training. 
To do so, we first train the model on the main task using the training dataset, and employ the trained model to give predictions on samples from the validation set. 
Then the calibration training dataset $\mathbb{D}^*=\{(x_i,y_i^*,c_i)_{i=1}^{M}\}$ can be generated from the validation set, where $x_i$ is the original sample in the validation set, $y_i^*$ is model's original prediction, and $c_i$ is a binary value that indicates whether the original prediction is correct or not.
Specifically, we perform downsampling to ensure a balanced label distribution.

In this paper, we adopt the same process to generate the calibration training dataset. 
But different methods may adopt specially designed training paradigms to utilize the calibration training data. 
We described the training details in Sec.~\ref{sec:cal_method}. 

%
%
%
%


\section{Additional Results of Calibration Methods}
\label{appendix:second_stage}
\looseness=-1
For exploring the effectiveness of existing calibration methods, we provide results with RoBERTa in Table~\ref{tab:roberta-shift}, Table~\ref{tab:hard2easy-roberta}, and Table~\ref{tab:robertaentropy}
The results with the model scaling effect are in Table~\ref{tab:emergent_scale}.

\begin{table*}[t!]
\resizebox{\textwidth}{!}{
\begin{tabular}{@{}l|ll|ccccc|ccccc|ccccc@{}}
\toprule
\multirow{12}{*}{Dynasent} & \multicolumn{2}{l|}{Dataset}                                  & \multicolumn{5}{c|}{Dynasent}                                           & \multicolumn{5}{c|}{Amazon}                                             & \multicolumn{5}{c}{DSC}                                                  \\ \cmidrule(l){2-18} 
                           & \multicolumn{2}{l|}{Method}                                   & Acc            & Conf  & ECE           & CErr$_{pos}$  & CErr$_{neg}$   & Acc            & Conf  & ECE           & CErr$_{pos}$  & CErr$_{neg}$   & Acc            & Conf  & ECE            & CErr$_{pos}$  & CErr$_{neg}$   \\ \cmidrule(l){2-18} 
                           & \multicolumn{1}{l|}{\multirow{5}{*}{Unlearnable}} & Vanilla   & \textbf{78.61} & 94.56 & 17.10         & \textbf{3.56} & 88.06          & 85.47          & 97.84 & 12.48         & 1.18          & 92.08          & 87.93          & 97.23 & 9.30           & 1.74          & 89.70          \\
                           & \multicolumn{1}{l|}{}                             & TS        & \textbf{78.61} & 77.47 & \textbf{0.95} & 19.47         & 66.96          & 85.47          & 86.61 & \textbf{2.54} & 11.24         & 74.11          & 87.93          & 85.09 & 2.99           & 12.84         & 70.03          \\
                           & \multicolumn{1}{l|}{}                             & LS        & 76.48          & 85.95 & 9.46          & 12.37         & 80.47          & \textbf{85.85} & 89.34 & 7.39          & 9.53          & 82.53          & \textbf{87.15} & 88.19 & 5.46           & 10.71         & 80.75          \\
                           & \multicolumn{1}{l|}{}                             & EDA       & 76.97          & 95.65 & 18.74         & 2.92          & 90.85          & 84.12          & 97.92 & 13.81         & 1.08          & 92.64          & 85.53          & 97.13 & 11.62          & 1.64          & 89.87          \\
                           & \multicolumn{1}{l|}{}                             & Ensemble  & 77.67          & 94.85 & 17.22         & 3.44          & 88.89          & 85.37          & 97.88 & 12.52         & \textbf{1.12} & 92.11          & 86.69          & 97.11 & 10.43          & \textbf{1.76} & 89.77          \\ \cmidrule(l){2-18} 
                           & \multicolumn{1}{l|}{\multirow{5}{*}{Learnable}}   & E-MLP     & \textbf{78.61} & 71.06 & 19.59         & 28.81         & 70.59          & 85.47          & 85.74 & 12.10         & 14.25         & 85.69          & 87.93          & 79.37 & \textbf{14.46} & 20.61         & 79.25          \\
                           & \multicolumn{1}{l|}{}                             & E-T5 \ours     & \textbf{78.61} & 64.94 & 23.76         & 26.76         & \textbf{34.43} & 85.47          & 85.53 & 13.23         & 9.45          & \textbf{56.03} & 87.93          & 81.72 & 14.91          & 13.50         & \textbf{49.71} \\
                           & \multicolumn{1}{l|}{}                             & I-Vanilla & 77.38          & 66.71 & 22.76         & 24.92         & 38.06          & 83.85          & 85.80 & 12.18         & 7.99          & 53.56          & 87.10          & 82.30 & 14.25          & 12.89         & 49.77          \\
                           & \multicolumn{1}{l|}{}                             & I-Iter \ours   & 77.89          & 64.17 & 21.98         & 28.43         & 38.09          & 84.49          & 87.49 & 10.00         & 7.47          & 60.06          & 87.05          & 82.83 & 12.14          & 12.86         & 53.81          \\
                           & \multicolumn{1}{l|}{}                             & I-Simul \ours  & 78.63          & 65.00 & 25.56         & 27.08         & 35.84          & 83.65          & 79.79 & 15.36         & 13.28         & 44.38          & 85.79          & 77.29 & 17.78          & 16.91         & 42.30          \\ \bottomrule
\end{tabular}
}
\caption{Results of RoBERTa's calibration performance under hard-to-easy distribution shifts.}

\label{tab:hard2easy-roberta}
\end{table*} 

\begin{table*}

\resizebox{\textwidth}{!}{

\begin{tabular}{@{}ll|cccccc|cccccc@{}}
\toprule
\multicolumn{2}{c|}{ID Dataset}                                 & \multicolumn{6}{c|}{SST-2}                                                                                      & \multicolumn{6}{c}{Yahoo}                                                                                      \\ \midrule
\multicolumn{2}{c|}{OOD Dataset}                                & \multicolumn{2}{c|}{SST-2}           & \multicolumn{2}{c|}{Bookcorpus}      & \multicolumn{2}{c|}{Random Words} & \multicolumn{2}{c|}{Yahoo}           & \multicolumn{2}{c|}{Bookcorpus}      & \multicolumn{2}{c}{Random Words} \\ \midrule
\multicolumn{2}{c|}{Method}                                     & Conf  & \multicolumn{1}{c|}{Entropy} & Conf  & \multicolumn{1}{c|}{Entropy} & Conf           & Entropy          & Conf  & \multicolumn{1}{c|}{Entropy} & Conf  & \multicolumn{1}{c|}{Entropy} & Conf           & Entropy         \\ \midrule
\multicolumn{1}{l|}{\multirow{5}{*}{Unlearnable}} & Vanilla     & 98.33 & \multicolumn{1}{c|}{4.27}    & 94.85 & \multicolumn{1}{c|}{12.63}   & 96.28          & 9.97             & 90.18 & \multicolumn{1}{c|}{26.96}   & 72.17 & \multicolumn{1}{c|}{77.84}   & 78.49          & 59.14           \\
\multicolumn{1}{l|}{}                             & TS          & 93.43 & \multicolumn{1}{c|}{19.62}   & 86.41 & \multicolumn{1}{c|}{32.66}   & 87.50          & 32.46            & 71.73 & \multicolumn{1}{c|}{90.13}   & 44.01 & \multicolumn{1}{c|}{163.43}  & 50.51          & 148.65          \\
\multicolumn{1}{l|}{}                             & LS          & 87.88 & \multicolumn{1}{c|}{35.74}   & 83.30 & \multicolumn{1}{c|}{42.64}   & 82.88          & 44.11            & 82.08 & \multicolumn{1}{c|}{74.02}   & 67.53 & \multicolumn{1}{c|}{110.10}  & 74.89          & 93.55           \\
\multicolumn{1}{l|}{}                             & EDA         & 98.43 & \multicolumn{1}{c|}{3.67}    & 95.54 & \multicolumn{1}{c|}{10.79}   & 91.55          & 20.06            & 94.24 & \multicolumn{1}{c|}{15.08}   & 83.30 & \multicolumn{1}{c|}{44.77}   & 86.10          & 35.91           \\
\multicolumn{1}{l|}{}                             & Ensemble    & 98.24 & \multicolumn{1}{c|}{4.49}    & 94.65 & \multicolumn{1}{c|}{12.87}   & 93.26          & 15.98            & 91.22 & \multicolumn{1}{c|}{23.92}   & 75.10 & \multicolumn{1}{c|}{69.13}   & 80.31          & 54.06           \\ \midrule
\multicolumn{1}{l|}{\multirow{5}{*}{Learnable}}   & E-MLP       & 94.48 & \multicolumn{1}{c|}{15.99}   & 80.75 & \multicolumn{1}{c|}{36.41}   & 63.81          & 59.36            & 74.15 & \multicolumn{1}{c|}{-}       & 41.87 & \multicolumn{1}{c|}{-}       & 42.31          & -               \\
\multicolumn{1}{l|}{}                             & E-T5 \ours       & 84.79 & \multicolumn{1}{c|}{16.26}   & 63.99 & \multicolumn{1}{c|}{24.34}   & 22.84          & 27.72            & 68.71 & \multicolumn{1}{c|}{-}       & 22.70 & \multicolumn{1}{c|}{-}       & 15.20          & -               \\
\multicolumn{1}{l|}{}                             & I-Vanilla   & 84.83 & \multicolumn{1}{c|}{16.33}   & 65.34 & \multicolumn{1}{c|}{25.09}   & 23.08          & 28.39            & 69.55 & \multicolumn{1}{c|}{-}       & 24.84 & \multicolumn{1}{c|}{-}       & 17.78          & -               \\
\multicolumn{1}{l|}{}                             & I-Iter \ours & 56.89 & \multicolumn{1}{c|}{20.06}   & 62.99 & \multicolumn{1}{c|}{21.10}   & 42.25          & 30.37            & 76.16 & \multicolumn{1}{c|}{-}       & 54.33 & \multicolumn{1}{c|}{-}       & 48.54          & -               \\
\multicolumn{1}{l|}{}                             & I-Simul \ours & 75.24 & \multicolumn{1}{c|}{9.44}    & 46.51 & \multicolumn{1}{c|}{13.88}   & 8.11           & 5.44             & 64.66 & \multicolumn{1}{c|}{-}       & 19.70 & \multicolumn{1}{c|}{-}       & 19.47          & -               \\ \bottomrule
\end{tabular}
}
\caption{Results on task-irrelevant inputs with RoBERTa. We don't report the entropy results of learnable methods when Yahoo is adopted as ID dataset since the class numbers are different in unlearnable (10 original classes in Yahoo) and learnable methods (2 classes), which will result in unfair comparison.}
\label{tab:robertaentropy}
\end{table*}

\section{Further Analysis of Distribution Shifts}
\label{sec:further_dis}
%
In Sec.~\ref{sec:standard}, we show that PLMs are less calibrated under distribution shifts, consistent with previous work~\citep{DBLP:conf/emnlp/DesaiD20, DBLP:conf/nips/MindererDRHZHTL21}.
However, can we safely conclude that distribution shifts degrade PLMs' calibration performance?
We study \textbf{hard-to-easy distribution shifts} (see Appendix~\ref{sec:appendix_eval_setting} for the detailed setting) to further investigate the essence of this problem.
In this setting, models are trained on a difficult ID dataset and infer on easier OOD datasets.
This comes with relatively lower ID and higher OOD performance.
Specifically, we consider the sentiment analysis task and choose Dynasent (Amazon and DSC) as the ID (OOD) datasets. 
The details of the datasets are described in Appendix~\ref{sec:appendix_dataset}.



The results of T5 and RoBERTa are shown in Table~\ref{tab:hard2easyt5} and Table~\ref{tab:hard2easy-roberta} respectively.
We observe completely different results with Sec.~\ref{sec:standard}.
Across all methods, the ECE and CErr$_{pos}$ decrease under the hard-to-easy distribution shifts, contradictory to the previous conclusion that PLMs are less calibrated on OOD samples. 
In hard-to-easy shifts, performance and confidence both increase due to the relative simpleness of the OOD samples. 
The indication is that PLMs' relative calibration performance on ID and OOD samples relies on the dataset difficulty, and the conclusion that PLMs are less calibrated under distribution shifts is one-sided.
This is consistent with our empirical study in Sec.~\ref{sec:empirical_study} that emphasizes the influence of dataset difficulty on PLMs calibration.







To further investigate the influence of dataset difficulty on PLMs' calibration performance, we evaluate \textbf{the calibration on task-irrelevant inputs} (see Appendix~\ref{sec:appendix_eval_setting} for the detailed setting) of PLMs trained on ID datasets with different difficulty (e.g., SST-2 and Yahoo). 
The task-irrelevant inputs include plain texts (e.g., bookcorpus) and random words. 
Since no golden labels are provided, we measure the calibration performance through maximum confidence scores and predictive entropy.



\begin{table*}[t]
\centering

\resizebox{0.75\textwidth}{!}{
\begin{tabular}{@{}l|l|ccccc@{}}
\toprule
Task                                                                                    & Dataset       & \# Classes & Avg.Len     & Train  & Dev   & Test  \\ \midrule
                                                                                        & SST-2         & 2          & 19.23       & 6920   & 1821  & 872   \\
\multirow{2}{*}{\begin{tabular}[c]{@{}l@{}}Sentiment\\ Analysis\end{tabular}}           & Amazon        & 3          & 77.86       & 24000  & 78741 & 91606 \\
                                                                                        & SST-5         & 3          & 18.75       & -      & -     & 1067  \\
                                                                                        & SemEval       & 3          & 19.61       & -      & -     & 6000  \\ \midrule
\multirow{3}{*}{\begin{tabular}[c]{@{}l@{}}Natural\\ Language\\ Inference\end{tabular}} & MNLI          & 3          & 19.36/10.06 & 373067 & 19635 & 9815  \\
                                                                                        & HANS          & 2          & 9.15/5.61   & -      & -     & 30000 \\
                                                                                        & ANLI          & 3          & 54.40/10.34 & -      & -     & 3200  \\ \midrule
\multirow{2}{*}{\begin{tabular}[c]{@{}l@{}}Topic\\ Classification\end{tabular}}         & Yahoo         & 10         & 96.98       & 126000 & 14000 & 60000 \\
                                                                                        & AG            & 4          & 38.5        & 10000  & -     & 7600  \\ \midrule
\multirow{3}{*}{\begin{tabular}[c]{@{}l@{}}Toxic\\ Detection\end{tabular}}              & Civil         & 2          & 52.86       & 48000 & 12000 & 97320 \\
                                                                                        & Hate Speech   & 2          & 21.55       & -      & -     & 478   \\
                                                                                        & Implicit Hate & 2          & 17.34       & -      & -     & 21479 \\ \midrule
\multirow{2}{*}{\begin{tabular}[c]{@{}l@{}}Plain\\ Text\end{tabular}}                   & Book Corpus   & -          & 13.39       & -      & -     & 10000 \\
                                                                                        & Random Words  & -          & 20.28       & -      & -     & 1000  \\ \bottomrule
\end{tabular}
}
\caption{Dataset Statistics.}
\label{tab:statistics}
\end{table*}

\begin{table*}[t!]
\resizebox{\textwidth}{!}{
\begin{tabular}{@{}l|l|l|l@{}}
\toprule
Task                                                                            & Dataset       & Template                                                                                                                                                                                               & Verbalizer                                                                                                                                                     \\ \midrule
                                                                                & SST-2         & It was \{"mask"\} . \{"placeholder": "text\_a"\}                                                                                                                                                       & {[}bad, good{]}                                                                                                                                                \\ \cmidrule(l){2-4} 
\multirow{2}{*}{\begin{tabular}[c]{@{}l@{}}Sentiment\\ Analysis\end{tabular}}   & Amazon        & It was \{"mask"\} . \{"placeholder": "text\_a"\}                                                                                                                                                       & {[}bad, good, neutral{]}                                                                                                                                       \\ \cmidrule(l){2-4} 
                                                                                & SST-5         & It was \{"mask"\} . \{"placeholder": "text\_a"\}                                                                                                                                                       & {[}bad, good, neutral{]}                                                                                                                                       \\ \cmidrule(l){2-4} 
                                                                                & SemEval       & It was \{"mask"\} . \{"placeholder": "text\_a"\}                                                                                                                                                       & {[}bad, good, neutral{]}                                                                                                                                       \\ \midrule
                                                                                & MNLI          & \begin{tabular}[c]{@{}l@{}}Given the two sentences: \\ (1) \{"placeholder": "text\_a"\}. \\ (2) \{"placeholder": "text\_b"\}. \\ Does the first sentence entails the second ? \{"mask"\}.\end{tabular} & {[}No, Yes, Maybe{]}                                                                                                                                           \\ \cmidrule(l){2-4} 
\begin{tabular}[c]{@{}l@{}}Natural\\ Language\\ Inference\end{tabular}          & HANS          & \begin{tabular}[c]{@{}l@{}}Given the two sentences: \\ (1) \{"placeholder": "text\_a"\}. \\ (2) \{"placeholder": "text\_b"\}. \\ Does the first sentence entails the second ? \{"mask"\}.\end{tabular} & {[}No, Yes, Maybe{]}                                                                                                                                           \\ \cmidrule(l){2-4} 
                                                                                & ANLI          & \begin{tabular}[c]{@{}l@{}}Given the two sentences: \\ (1) \{"placeholder": "text\_a"\}. \\ (2) \{"placeholder": "text\_b"\}. \\ Does the first sentence entails the second ? \{"mask"\}.\end{tabular} & {[}No, Yes, Maybe{]}                                                                                                                                           \\ \midrule
\multirow{2}{*}{\begin{tabular}[c]{@{}l@{}}~\\Topic\\ Classification\end{tabular}} & Yahoo         & \begin{tabular}[c]{@{}l@{}}A \{"mask"\} \\ question : \{"placeholder": "text\_a"\} \{"placeholder": "text\_b"\}\end{tabular}                                                                           & \begin{tabular}[c]{@{}l@{}}{[}society, science,\\ health, education,\\ computers, sports,\\ business, entertainment,\\ relationships, politics{]}\end{tabular} \\ \cmidrule(l){2-4} 
                                                                                & AG            & \begin{tabular}[c]{@{}l@{}}A \{"mask"\} news : \{"placeholder": "text\_a"\} \\ \{"placeholder": "text\_b"\}\end{tabular}                                                                               & \begin{tabular}[c]{@{}l@{}}{[}politics, sports,\\ business, technology{]}\end{tabular}                                                                         \\ \midrule
                                                                                & Civil         & It was \{"mask"\} . \{"placeholder": "text\_a"\}                                                                                                                                                       & {[}benign, toxic{]}                                                                                                                                            \\ \cmidrule(l){2-4} 
\begin{tabular}[c]{@{}l@{}}Toxic\\ Detection\end{tabular}                       & Hate Speech   & It was \{"mask"\} . \{"placeholder": "text\_a"\}                                                                                                                                                       & {[}benign, toxic{]}                                                                                                                                            \\
                                                                                & Implicit Hate & It was \{"mask"\} . \{"placeholder": "text\_a"\}                                                                                                                                                       & {[}benign, toxic{]}                                                                                                                                            \\ \bottomrule
\end{tabular}
}

\caption{The manual templates and verbalizers adopted for each dataset.}
\label{tab:template}
\end{table*}

The results of T5 are shown in Table~\ref{tab:entropyt5}, and RoBERTa are shown in Table~\ref{tab:robertaentropy}.
We show that PLMs have unreasonable high confidence in task-irrelevant inputs, especially when trained on SST-2. 
Comparing the results when trained on SST-2 or Yahoo, we find that the ID training dataset has significant influence on PLMs calibration.
Still, this can be attributed to the dataset difficulty.
We also observe the superior performance of learnable calibration methods. 
They produce lower confidence scores on plain text and random tokens compared to unlearnable ones. 

\looseness=-1
In summary, the influence of distribution shifts on PLMs calibration is dependent on the evaluation datasets chosen. 
The original conclusion that calibration performance degrades on OOD samples is based on two premises: 
(1) PLMs are overconfident in their wrong predictions, which is supported by our experiments;
(2)~The OOD datasets are harder so PLMs cannot achieve good performance.
The second premise has not always been satisfied, and we show that the relative dataset difficulty significantly influences PLMs' calibration performance on ID and OOD samples.

\begin{table*}[t!]
\renewcommand{\arraystretch}{1.3}
\resizebox{\textwidth}{!}{
\begin{tabular}{@{}l|l|l|l@{}}
\toprule
Task                                                                          & Dataset        & Template                                                                                                                                                                                                                                     & Verbalizer        \\ \midrule
                                                                              & SST-2          & \begin{tabular}[c]{@{}l@{}}Sentence: \{"placeholder": "text\_a"\} The predicted sentiment is \{"placeholder": "text\_b"\} . \\ Is the prediction True or False ? It's \{"mask"\} .\end{tabular}                                              &                   \\ \cmidrule(lr){2-3}
\multirow{2}{*}{\begin{tabular}[c]{@{}l@{}}~\\Sentiment\\ Analysis\end{tabular}} & Amazon         & \begin{tabular}[c]{@{}l@{}}Sentence: \{"placeholder": "text\_a"\} The predicted sentiment is \{"placeholder": "text\_b"\} . \\ Is the prediction True or False ? It's \{"mask"\} .\end{tabular}                                              &                   \\ \cmidrule(lr){2-3}
                                                                              & SST-5          & \begin{tabular}[c]{@{}l@{}}Sentence: \{"placeholder": "text\_a"\} The predicted sentiment is \{"placeholder": "text\_b"\} . \\ Is the prediction True or False ? It's \{"mask"\} .\end{tabular}                                              &                   \\ \cmidrule(lr){2-3}
                                                                              & SemEval        & \begin{tabular}[c]{@{}l@{}}Sentence: \{"placeholder": "text\_a"\} The predicted sentiment is \{"placeholder": "text\_b"\} . \\ Is the prediction True or False ? It's \{"mask"\} .\end{tabular}                                              &                   \\ \cmidrule(r){1-3}
                                                                              & MNLI           & \begin{tabular}[c]{@{}l@{}}Given the two sentences: \{"placeholder": "text\_a"\} \\ The predicted relationship between the two sentences is \{"placeholder": "text\_b"\} \\ Is the prediction True or False ? It's \{"mask"\} .\end{tabular} &                   \\ \cmidrule(lr){2-3}
\begin{tabular}[c]{@{}l@{}}Natural\\ Language \\ Inference\end{tabular}       & HANS           & \begin{tabular}[c]{@{}l@{}}Given the two sentences: \{"placeholder": "text\_a"\} \\ The predicted relationship between the two sentences is \{"placeholder": "text\_b"\} \\ Is the prediction True or False ? It's \{"mask"\} .\end{tabular} & {[}False, True{]} \\ \cmidrule(lr){2-3}
                                                                              & ANLI           & \begin{tabular}[c]{@{}l@{}}Given the two sentences: \{"placeholder": "text\_a"\} \\ The predicted relationship between the two sentences is \{"placeholder": "text\_b"\} \\ Is the prediction True or False ? It's \{"mask"\} .\end{tabular} &                   \\ \cmidrule(r){1-3}
\begin{tabular}[c]{@{}l@{}}Topic\\ Classification\end{tabular}                & Yahoo          & \begin{tabular}[c]{@{}l@{}}Sentence: \{"placeholder": "text\_a"\} The predicted topic is \{"placeholder": "text\_b"\} \\ Is the prediction True or False ? It's \{"mask"\} .\end{tabular}                                                    &                   \\ \cmidrule(r){1-3}
                                                                              & Civil          & \begin{tabular}[c]{@{}l@{}}Sentence: \{"placeholder": "text\_a"\} The predicted toxicity is \{"placeholder": "text\_b"\} . \\ Is the prediction True or False ? It's \{"mask"\} .\end{tabular}                                               &                   \\ \cmidrule(lr){2-3}
\begin{tabular}[c]{@{}l@{}}Toxic\\ Detection\end{tabular}                     & Hate Speech    & \begin{tabular}[c]{@{}l@{}}Sentence: \{"placeholder": "text\_a"\} The predicted toxicity is \{"placeholder": "text\_b"\} . \\ Is the prediction True or False ? It's \{"mask"\} .\end{tabular}                                               &                   \\ \cmidrule(lr){2-3}
                                                                              & Implicite Hate & \begin{tabular}[c]{@{}l@{}}Sentence: \{"placeholder": "text\_a"\} The predicted toxicity is \{"placeholder": "text\_b"\} . \\ Is the prediction True or False ? It's \{"mask"\} .\end{tabular}                                               &                   \\ \bottomrule
\end{tabular}

}

\caption{The manual templates and verbalizers of the calibration task for each dataset.}
\label{tab:calibration_template}
\end{table*}

\section{Details of Evaluation setting.}
\label{sec:appendix_eval_setting}
\paragraph{Hard-to-easy shift.}
we choose Dynasent as the in-distribution dataset, and choose Amazon and DSC as the out-of-distribution datasets. 
The evaluation metrics are the same as the ones adopted in experiments on standard OOD shifts. 
This evaluation setting is expected to test the conclusion that  PLMs' calibration performance degrades under distribution shifts. 

\paragraph{Calibration on task-irrelevant inputs}
We choose SST-2 and Yahoo as the in-distribution datasets, and choose Bookcorpus and a synthetic dataset as out-of-distribution datasets. 
Each sample in the synthetic dataset is constructed by composing random words. 
Well-calibrated PLMs should give very low confidence and high probability entropy in the task-irrelevant inputs.

\end{document}